\documentclass[lettersize,journal]{IEEEtran}
\usepackage{amsmath,amsfonts}
\usepackage{algorithmic}
\usepackage{algorithm}
\usepackage{array}
\usepackage[caption=false,font=normalsize,labelfont=sf,textfont=sf]{subfig}
\usepackage{textcomp}
\usepackage{stfloats}
\usepackage{url}
\usepackage{verbatim}
\usepackage{graphicx}
\usepackage{cite}
\usepackage{amsmath}
\usepackage{graphicx}
\usepackage{multirow}
\usepackage{makecell}
\begin{document}

\title{Recent Advances of Local Mechanisms in Computer Vision: A Survey and Outlook of Recent Work}

\author{Qiangchang Wang, Yilong Yin
\thanks{Q. Wang, Y. Yin are with Shandong University, China (E-mail: qiangchang.wang@gmail.com, ylyin@sdu.edu.cn).}}

\markboth{Journal of \LaTeX\ Class Files,~Vol.~14, No.~8, August~2021}%
{Shell \MakeLowercase{\textit{et al.}}: A Sample Article Using IEEEtran.cls for IEEE Journals}


\maketitle

\begin{abstract}
Inspired by the fact that human brains can emphasize discriminative parts of the input and suppress irrelevant ones, substantial local mechanisms have been designed to boost the development of computer vision. They can not only focus on target parts to learn discriminative local representations, but also process information selectively to improve the efficiency. In terms of application scenarios and paradigms, local mechanisms have different characteristics. In this survey, we provide a systematic review of local mechanisms for various computer vision tasks and approaches, including fine-grained visual recognition, person re-identification, few-/zero-shot learning, multi-modal learning, self-supervised learning, Vision Transformers, and so on. Categorization of local mechanisms in each field is summarized. Then, advantages and disadvantages for every category are analyzed deeply, leaving room for exploration. Finally, future research directions about local mechanisms have also been discussed that may benefit future works. To the best our knowledge, this is the first survey about local mechanisms on computer vision. We hope that this survey can shed light on future research in the computer vision field. 
\end{abstract}


\section{Introduction}\label{sec_introduction}


\IEEEPARstart{O}{ur} visual processing system can focus on some parts of the input and ignore other irrelevant information, benefiting in analyzing and understanding complex scenes effectively \cite{xu2015show,rensink2000dynamic,corbetta2002control}. Inspired by this, different local mechanisms are proposed to benefit the development of computer vision. Currently, local mechanisms play a crucial role in many computer vision tasks and approaches, such as fine-grained visual recognition, person re-identification, few-/zero-shot learning, multi-modal learning, self-supervised learning, Vision Transformers (ViT) \cite{dosovitskiy2021image}, etc. In these tasks, local mechanisms not only refer to information relating to local areas of target objects, but also including processing information selectively instead of covering all information globally. 

For different approaches and requirements, representative abilities of learned features vary under different environments and appearances. In some scenarios, the representation ability of local features determines the performance. Local mechanisms have several advantages: 1) Although the ViT achieves decent performance on various computer vision tasks, they have some drawbacks: The computation and memory cost is quadratically related with the image resolution and token number; The inductive bias is lacked, making the ViT data hungry. It can be expected that the aforementioned drawbacks can be alleviated by designing some mechanisms to inject the inductive bias into the ViT or process tokens selectively; 2) Due to many factors (e.g., illumination changes and image quality variations), the data distribution under different domains or modalities is different, leading to performance degradation in the cross-domain or cross-modality setting. Since local features describe general detailed information which is transferable between different modalities or domains, they can be used to boost the performance of multimodal learning and self-supervised learning; 3) It is generally required to train deep learning models on large-scale datasets. However, labeling large-scale data is often expensive, time-consuming, and sometimes infeasible. For example, it is difficult to collect a large number of images of rare species. It is worth exploring to recognize novel classes given only a few labeled examples or novel class descriptions. Local features describe fine-grained information in objects, which are easy to transfer between seen classes and unseen classes. Local features can be regarded as a good manner of data augmentation, especially for novel classes with few labeled samples or category descriptions. Therefore, local representations have been widely used in few-/zero-shot learning; 4) Due to various factors (e.g., occlusions, pose, or viewpoint variations), global image representations suffer from dramatic changes, leading to performance drop under these challenging scenarios. On the other hand, local representations contain local details, which are more robust to occlusions, pose, or viewpoint variations. This is the reason why local features are widely used in some applications to boost the performance, like fine-grained visual recognition and person re-identification. 

It should be noticed that it is difficult to say what features are local because localness is a relative concept. As argued in \cite{hinton2021represent}, images consist of hierarchical semantic concepts with five levels, including the lowest level (pixel elements), sub-part level (object shapes and textures), part level, object level, and scene level. Extracting which level of semantic representations depends on the solving problem and required information. In the past decade, local mechanisms have played an increasingly important role in computer vision. Progresses can be coarsely divided into three categories. The first category is about hard part division. It divides the global information into parts horizontally or vertically in a fixed way, such as cropping parts around fixed locations or keypoints on the original image or feature maps. In the second category, various methods adopt soft part division to automatically emphasize important parts and suppress irrelevant ones. Typical methods include attention modules, semantic segmentation, clustering, and so on. The last category is the Transformer era, showing great potentials of capturing local details. It is clear that local mechanisms are powerful and general modules to benefit the development of computer vision.

\begin{figure}
\centering
\includegraphics[width=0.42\textwidth]{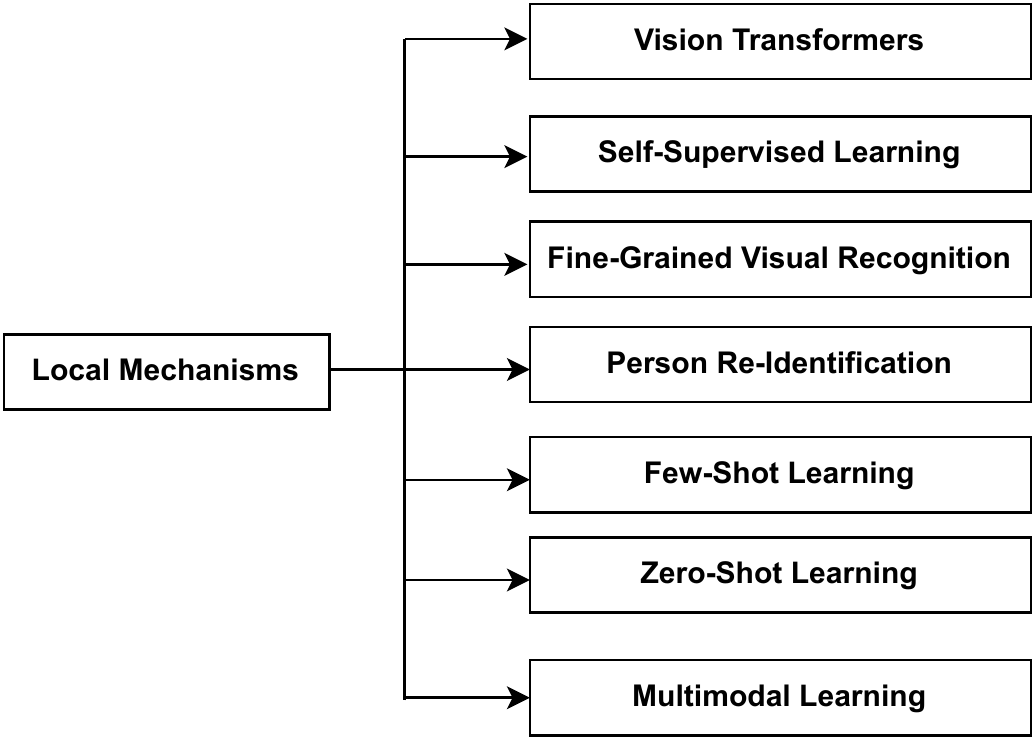}
\caption{The reviewed deep learning methods and computer vision applications with local mechanisms.}
\label{fig_overview}
\end{figure}

With the progress of deep learning, the number of related papers have been growing dramatically. Therefore, creating a comprehensive and thorough survey is challenging. As shown in Fig. \ref{fig_overview}, this survey provides an overview on local representations in various computer vision tasks and how local mechanisms have been adopted in different approaches. The remarkable papers about the implementations and effects of local mechanisms in the following areas are reviewed: including Vision Transformers, self-supervised learning, fine-grained visual recognition, person re-identification, few-/zero-shot learning, and multimodal learning. Selection criteria of are listed here: To better analyze the state-of-the-art approaches, papers from top journals and conferences about local mechanisms in the last several years are discussed. Besides, for the completeness, some early outstanding works may also be covered. Although we attempt to comprehensively discuss every related paper at the time of submission, some outstanding papers may be missed in this survey because of the dramatic increase in computer vision papers and our limited knowledge and perceptions. We sincerely apologize to these authors who made great efforts into this field but fail to be mentioned here. 

The contributions of this survey can be summarized as follows:
\begin{enumerate}

\item To the best of our knowledge, this is the first attempt to review the recent progress about local mechanisms in several important computer vision fields and approaches. 

\item Many important specific local designing mechanisms in several computer vision fields are summarized. This makes the implementations of local representation learning process easy to understand. The advantages and disadvantages in each computer vision field are also discussed to boost the development of local mechanisms.

\item Potential research directions of local mechanisms in the future are discussed.

\end{enumerate}

The remainder of this paper is organized as follows.  Section \ref{sec_vit} lists local mechanisms in Vision Transformers due to their remarkable performance in different computer vision fields. Section \ref{sec_ssl} shows the progress about the local mechanisms in self-supervised learning. Section \ref{sec_fgvr} discusses various local feature learning methods about fine-grained visual recognition. Section \ref{sec_reid} provides a review about local mechanisms in person re-identification. Sections \ref{sec_fsl} and \ref{sec_zsl} review the progress of local mechanisms in few-shot learning and zero-shot learning, respectively. Section \ref{sec_mml} discusses the updates of local mechanisms in multimodal learning. Section \ref{sec_outlook} discusses future directions of local mechanisms in computer vision. Finally, section \ref{sec_conclusion} presents conclusions of local mechanisms.

\section{Related Surveys}

Currently, to the best of our knowledge, this survey is the first attempt to review various local mechanisms comprehensively in computer vision. Several papers summarize the development of attention modules \cite{chaudhari2021attentive,guo2022attention} and Vision Transformers \cite{han2022survey,khan2022transformers}, while our paper reviews local mechanisms in computer vision more generally, not just attention modules and Vision Transformers. In this survey, we provide a classification which categories various local mechanisms according to their field of applications. Doing so allows us to concentrate on the analysis of local mechanisms in each specific area which have different characteristics. It can be understood how local mechanisms can be used to solve different issues in computer vision. 

Although there are related surveys in each field, the review perspective is different from ours where they summarize the development from various perspectives in the field. In comparison, our review perspective is limited to local mechanisms which is extended into different fields for comprehensive and deep analysis. 




\section{Vision Transformers}\label{sec_vit}

Inspired by the success of traditional Transformers in natural language processing \cite{vaswani2017attention}, many researchers explore their potentials on computer vision. Recently, Vision Transformers (ViT) \cite{dosovitskiy2021image} have shown favorable performance in various vision tasks by leveraging the strong power of the self-attention mechanism in learning long-range correlations among image patches. 

\begin{figure}[ht!]
\centering
\includegraphics[width=0.46\textwidth]{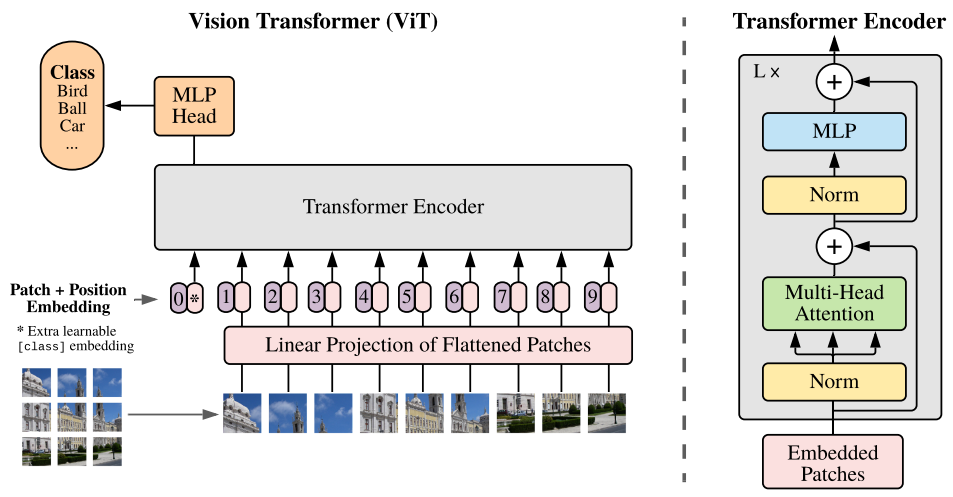}
\caption{The general framework of Vision Transformer (ViT). The image is from \cite{dosovitskiy2021image}. }
\label{fig_vit_framework}
\end{figure}

An overview of the model is shown in Fig. \ref{fig_vit_framework}. Specifically, firstly, an input image $x \in R^{H\times W \times C}$ is reshaped into a sequence of flattened 2D patches $x_p \in R^{N \times P^2\times C}$ where $H$, $W$ is the size of the input image, $C$ is the number of channels, $P$ is the size of each image patch, $N=HW/P^2$ represents the number of image patches in the input sequence of the Transformer. Secondly, each patch is linearly projected into $D$ dimensions with the following formula:
\begin{equation}
    z_0=[x_{class}; x_p^1E; x_p^2E; ...; x_p^NE]+E_{pos},
\end{equation}
where $x_{class}$ is a classification token, $E\in R^{P^2\times C\times D}$ is the linear projection, $E_{pos}\in R^{(N+1)\times D}$ denotes the position embeddings to keep the position information. Thirdly, patch embeddings are further processed by $L$ Transformer encoders which consist of multi-head self-attention (MSA) and MLP blocks. Layernorm (LN) \cite{ba2016layer} is applied before every block as follows:
\begin{flalign}
    \begin{split}
    z_l^*=MSA(LN(z_{l-1}))+z_{l-1},\\
    z_l = MLP(LN(z_l^*))+ z_l^*,
    \end{split}
\end{flalign}
where $l=1...L$. Finally, the classification token in the last layer is used for classification:
\begin{equation}
    y=LN(z_L^{0}).
\end{equation}

However, there exist several drawbacks: 1) In general, the performance of the ViT is closely related with the token number and image resolution \cite{dosovitskiy2021image}, however, the computation and memory cost of the ViT is also quadratically related. This makes the ViT difficult to learn fine-grained details. Consequently, performance of the ViT on dense tasks is limited, like semantic segmentation and object detection; 2) Lacking inductive bias, ViTs have a sub-optimal performance compared with similarly sized CNNs when training on small-sized data. 

\begin{figure}[ht!]
\centering
\includegraphics[width=0.35\textwidth]{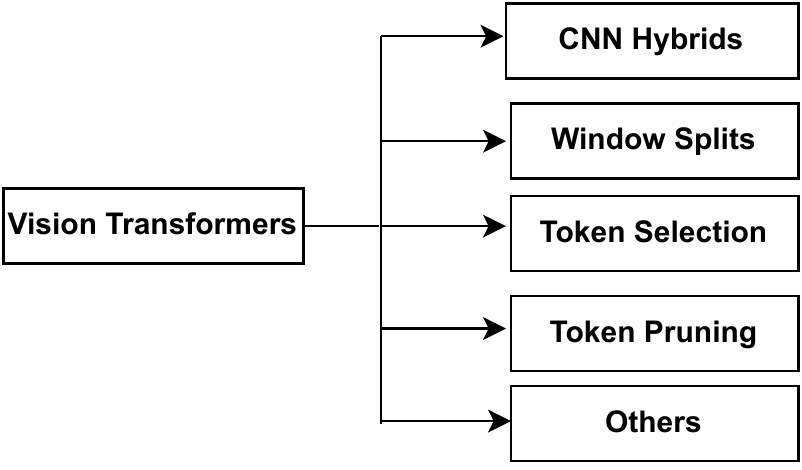}
\caption{The family of the ViT with local mechanisms.}
\label{fig_vit_family}
\end{figure}

Fortunately, images have more spatial redundancy than languages \cite{sun2015photoplethysmography}, like task-unrelated regions. Therefore, many different kinds of approaches are proposed, as shown in Fig. \ref{fig_vit_family}. Firstly, CNN hybrids can combine CNNs with the ViT, incorporating the inductive bias from the CNN to accommodate the ViT. Secondly, window splits generate different windows where different levels of information is learned within and across windows, dramatically decreasing the computational cost. Thirdly, tremendous efforts are made to automatically select discriminative tokens(i.e., token selection). Fourthly, many token pruning approaches are also proposed to adaptively excavate redundant tokens. Fifthly, other mechanisms (e.g., deformable attentions, mixup operations) are also designed to learn useful local features from the ViT.

\subsection{CNN Hybrids}

Before the ViT, CNNs are the mainstream approach in the deep learning era which can aggregate useful information by a shifted window. Specifically, it has two advantages: 1) It can extract local features in a neighborhood. This is complementary with global features learned by the ViT. Besides, it has a relatively low computational cost; 2) Low-level features which constitute fundamental structures can be captured by the CNNs, boosting the representational ability of the ViT. Therefore, the CNN and ViT are combined to achieve a better performance.

Although Swin Transformers \cite{liu2021swin} have obtained decent performance on various vision tasks, its shifted windows have uneven window sizes, leading to implementation difficulties in deep learning frameworks. A novel spatial attention is carefully designed in Twins \cite{chu2021twins} where a locally-grouped self-attention learns detailed and short-range features and a global sub-sampled attention extracts global and long-range information with separable depth-wise convolutions. 

In order to improve the performance of ViTs under small-sized datasets, Convolutional vision Transformer (CvT) \cite{wu2021cvt} and Convolution-enhanced image Transformer (CeiT) \cite{yuan2021incorporating}. Specifically, CvT incorporates convolutions into ViT, which consists of a hierarchy of Transformers and a convolutional Transformer block. Consequently, desirable properties of both Transformers (self-attention, global context, and good generalization) and CNNs (shift, scale, and distortion invariance) are maintained. Similarly, CeiT leverages merits of CNNs (low-level features, locality enhancement) and Transformers (long-range dependencies) where an image-to-token module generates patches of low-level features, a locally-enhanced feed forward enhances the correlation learning among neighboring tokens, and a layer-wise class token attention incorporates multi-level representations. 

There exist several drawbacks in Transformers: 1) Low-level features are failed to be captured which are basics of fundamental structures; 2) The memory and computation in self-attention scale quadratically with spatial or embedding dimensions; 3) Each head in multi-head self-attention only constitutes a subset of dimensions, degrading the performance; 4) The input tokens and position encoding are at a fixed scale. ResT \cite{zhang2021rest} (named after ResNet \cite{he2016deep}) proposes a memory-efficient multi-head self-attention to model interactions among different heads, improving their diversity. Besides, the positional encoding is redesigned for arbitrarily sized images.

In order to fill the gap between locality of CNNs and global relations of ViTs, Local Vision Transformer (Local ViT) \cite{han2021connection} performs the attention over small local windows where the local attention is rephrased as a channel-wise spatially-locally connected layer with dynamic weight connections.

The performance usually drops dramatically when either scaling Swin Transformer \cite{liu2021swin} or PVT \cite{wang2021pyramid} to a mobile-sized model. Lite Vision Transformer (LVT) \cite{yang2022lite} proposes two advanced self-attention modules for mobile deployments. Specifically, convolutional self-attention combines local self-attention with a convoltional kernel to learn low-level features. Recursive self-attentions extract the multi-scale context and use a recursive mechanism to boost the representation ability for high-level features.

ViT usually has an inferior performance under high-resolution low-computation scenarios compared with CNNs. The bottleneck is the softmax attention module which has a quadratic complexity with the input resolution. Although existing methods limit the softmax attention within local windows or reduce the key/value dimensions, the global relations are missed. EfficientViT is proposed in \cite{cai2023efficientvit} to preserve global and local feature abilities with linear computational complexity. In particular, the softmax attention is replaced with linear attentions. Besides, depthwise convolution is used to enhance the ability to extract discriminative local features.

The ViTs are not as efficient as CNNs. An efficient ViT architecture, named Doubly-Fused ViT (DFvT) \cite{gaodoubly} where convolutions are fused with the ViT pipeline. Specifically, a context module which consists of a fused sub-sampled operator and subsequent self-attention can capture global features efficiently. Besides, a spatial module preserves fine-grained spatial representations. Moreover, a dual attention enhancement module is proposed to selectively combine low-level and high-level features.

\subsection{Window Splits}

The ViT tends to have a high computational cost and latency. Moreover, it tends to neglect fine-grained details. Many approaches address this issue by splitting images into small-sized windows, then learning detailed information within windows and building long-range relations across windows.

It is observed that the ViT obtains inferior results compared with CNNs, given a midsize dataset (e.g., ImageNet). This is because discriminative local structures are failed to be captured among neighboring pixels in the hard split of image patches. In order to address this issue, a tokens-to-token vision Transformer (T2T-ViT) is proposed in \cite{yuan2021tokens} to progressively aggregate adjacent tokens into one token. The ViT splits an input image into a sequence of image patches, but neglects the local structure information inside the patch. A Transformer iN Transformer (TNT) is proposed in \cite{han2021transformer}. Except the outer Transformer block which learns global relations among patches, an inner Transformer block is proposed to model local structure information on pixels within each patch.

Image tokens have a fixed size in ViT, which mainly have two limitations: First, this is sub-optimal because visual elements have varying sizes for some vision tasks, like object detection; Second, this increases the computational complexity of self-attention for high-resolution images, which grows quadratically as the resolution of the input image. In \cite{liu2021swin}, a general-purpose Transformer, called Swin Transformer, is proposed to solve different tasks. It restricts self-attention computation within non-overlapped sub-windows and calculates cross-window interactions with a shifted window. This allows networks to learn features at various scales and has a linear complexity with respect to the image size.

To learn multi-scale features and cope with computational and memory-intensive issues, a regional-to-local attention (RegionViT) is proposed in \cite{chen2022regionvit}. In which, regional and local tokens are generated from an image with varying patch sizes. Then, the regional self-attention is applied on all regional tokens to extract global information and local self-attention is employed between one regional token and its associated local tokens to learn local features. 

To reduce the high latency and large memory consumption for high-resolution images, a new shift-invariant local attention layer, named query and attend (QnA), is proposed in \cite{arar2022learned}. Specifically, the self-attention is conducted within non-interleaving windows, building a linear complexity to the image size. However, the cross-window relation is neglected, degrading the performance. The QnA can aggregate information in an overlapped way by introducing learned queries.

The ViT has a limited ability to capture multi-scale features, suffering from a sub-optimal performance for objects at different scales. To address this issue, shunted self-attention (SSA) \cite{ren2021shunted} learns attentions at various scales by injecting heterogeneous receptive field sizes into tokens. Particularly, it selectively aggregates tokens to denote large objects and preserves some tokens to keep fine-grained features. Consequently, relations about multi-granularity features are effectively captured, and the token number and the computation cost are reduced.    

In the ViT, an image is divided into patches with a fixed size. However, various patches contribute differently in human-centric vision tasks, like the human object and background. To overcome this issue, Token Clustering Transformer (TCFormer) \cite{zeng2022not} is proposed to generate tokens with flexible locations, shapes, and sizes by progressive clustering and merging tokens. Further, a multi-stage token aggregation is proposed to fuse tokens in different stages, preserving various image details.

To improve efficiency of Transformers, some works restrict from the global attention to local/window attentions. However, the receptive field size is enlarged slowly. Cross-Shaped Window (CSWin) \cite{dong2022cswin} is proposed. Specifically, a cross-shape window divides the multi-head self-attention into two groups and conducts the self-attention in the horizontal and vertical strips simultaneously. Locally-enhanced positional encoding is proposed to add local positional information.

Either coarse global or local self-attention is usually adopted to reduce the computational cost, while neglecting local relations or global modeling. To this end, Orthogonal Transformer (OT) \cite{huang2022orthogonal} is proposed. Concretely, orthogonal self-attention can model global relations and explore local dependency. Besides, positional MLP is proposed to incorporate positional information for arbitrary resolution.

\subsection{Token Selection}
Due to the complexity of image structures, image tokens contain different information which contribute differently. An adaptive mechanism is highly expected to select discriminative tokens and discard useless ones.

It is argued in \cite{wang2021not} that the token number should be adaptively adjusted for every image. Some easy images only need coarse image patches, while some difficult images need fine-grained ones. Motivated by this fact, a dynamic Transformer (DVT) is proposed to set a decent number of tokens for each image. Specifically, multiple Transformers are cascaded with an increasing number of tokens, which are sequentially activated based on the prediction confidence. Besides, to reduce redundant computations across cascaded Transformers, feature reuse and relationship reuse are designed. 

Instead of utilizing predefined patch splitting methods, a method called Tokenlearner  \cite{ryoo2021tokenlearner} explores important tokens adaptively. Specifically, important regions are located automatically by multiple spatial attention maps, which are further tokenized for subsequent learning.

In previous methods, local information is neglected in image classification which would improve the generalization ability of models. In LV-ViT \cite{jiang2021all}, all image patch tokens are leveraged which contain rich features on local image patches. Specifically, the token labeling strategy is proposed to assign each patch token to local details provided by machine annotators, embedding rich local features for image classification and some downstream tasks, like semantic segmentation.

It is observed in EViT \cite{liang2021evit} that different image tokens play different roles in ViTs and some even contribute negatively. A token reorganization method is proposed to identify and fuse image tokens. More specifically, the token importance is measured by attention value between this token and the classification token. Then, important image tokens are preserved by conducting a Top-K operation. Meanwhile, less discriminative tokens are fused into one but not discarded. This is because some large objects may occupy a large proportion of the images, making is unreasonable to preserve a fixed number of image tokens.

Images have large areas of low-level texture and uninformative background, bringing high computation inefficiency. A self-motivated slow-fast token evolution (Evo-ViT) is developed in \cite{xu2022evovit} which consists of the structure preserving token selection and a slow-fast updating method, addressing the limitations of modern token pruning methods. 

A differentiable Adaptive Token Sampling (ATS) is proposed in \cite{fayyaz2022adaptive} to score and adaptively sample important tokens. Consequently, the number of selected tokens varies based on the image content.

\begin{table*}
\centering
\caption{Representative ViT models with local mechanisms.}
\begin{tabular}{|c|c|c|c|c||c|c|c|c|c|}
\hline
\multicolumn{1}{|c|}{\multirow{1}{*}{Category}} & \multirow{1}{*}{Methods} & \multirow{1}{*}{Venue} & \multirow{1}{*}{Highlights} \\
\hline
\multirow{9}{*}{\rotatebox[origin=c]{90}{\makecell{CNN \\Hybrids}}}& \makecell{Twins \cite{chu2021twins}}& ICCV21 &\makecell{A locally-grouped self-attention, a global sub-sampled attention}\\
\cline{2-4}
& \makecell{CVT \cite{wu2021cvt}}& ICCV21 &\makecell{Hierarchy of Transformers, convolutional Transformer blocks}\\
\cline{2-4}
& \makecell{CeiT \cite{yuan2021incorporating}}& NeurIPS21 &\makecell{Image-to-token, locally-enhanced feed forwardlayers, layer-wise class token attentions}\\
\cline{2-4}
& \makecell{ResT \cite{zhang2021rest}}& NeurIPS21 &\makecell{Memory-efficient multi-head self-attention, redesigned positional encoding}\\
\cline{2-4}
& \makecell{Local ViT \cite{han2021connection}}& ICLR22 &\makecell{A channel-wise spatially-locally connected layer, dynamic weight connections}\\
\cline{2-4}
& \makecell{LVT \cite{yang2022lite}}& CVPR22 &\makecell{Convolutional self-attention, recursive self-attention}\\
\cline{2-4}
& \makecell{EfficientViT \cite{cai2023efficientvit}}& CVPR23 &\makecell{Linear attention, depth-wise convolution}\\
\cline{2-4}
& \makecell{DFvT \cite{gaodoubly}}& ECCV22 &\makecell{A context module, a spatial module, a dual attention enhancement}\\
\hline
\multirow{11}{*}{\rotatebox[origin=c]{90}{\makecell{Window \\Splits}}}&\makecell{Swin Transformer \cite{liu2021swin}}& ICCV21 &\makecell{Cross-window interactions, intra-window interactions, shifted window}\\
\cline{2-4}
& \makecell{T2T-ViT \cite{yuan2021tokens}}&  ICCV21 &\makecell{Progressive aggregation of adjacent tokens}\\
\cline{2-4}
& \makecell{TNT \cite{han2021transformer}}& NeurIPS21 &\makecell{Transformer iN Transformer, outer Transformer block, inner Transformer block}\\
\cline{2-4}
& \makecell{Local ViT \cite{han2021connection}}& ICLR22 &\makecell{Channel-wise spatially-locally connected layer, dynamic weight connections}\\
\cline{2-4}
& \makecell{RegionViT \cite{chen2022regionvit}}& ICLR22 &\makecell{Regional self-attention, local self-attention}\\
\cline{2-4}
& \makecell{QnA \cite{arar2022learned}}& CVPR22 &\makecell{Cross-window relation, within-window aggregation, learned queries}\\
\cline{2-4}
& \makecell{SSA \cite{ren2021shunted}}& CVPR22 &\makecell{Multi-scale features, selective token aggregation}\\
\cline{2-4}
& \makecell{CSWin \cite{dong2022cswin}}& CVPR22 &\makecell{Cross-shape window, locally-enhanced positional encoding}\\
\cline{2-4}
& \makecell{TCFormer \cite{zeng2022not}}& CVPR22 &\makecell{Progressive token clustering, multi-stage token aggregation}\\
\cline{2-4}
& \makecell{OT \cite{huang2022orthogonal}}& NeurIPS22 &\makecell{Orthogonal self-attention, positional MLP}\\
\hline
\multirow{6}{*}{\rotatebox[origin=c]{90}{\makecell{Token \\Selection}}}& \makecell{DVT \cite{wang2021not}}& NeurIPS21 &\makecell{Cascaded Transformer, feature reuse and relationship reuse}\\
\cline{2-4}
& \makecell{Tokenlearner \cite{ryoo2021tokenlearner}}& NeurIPS21 &\makecell{Adaptive localization of important regions}\\
\cline{2-4}
& \makecell{LV-ViT \cite{jiang2021all}}& NeurIPS21 &\makecell{Token labeling strategy}\\
\cline{2-4}
& \makecell{EViT \cite{liang2021evit}}& NeurIPS21 &\makecell{Token reorganization, important tokens preserving, uninformative tokens fusion}\\
\cline{2-4}
& \makecell{Evo-ViT \cite{xu2022evovit}}& AAAI22 &\makecell{Structure preserving token selection, slow-fast updating method}\\
\cline{2-4}
& \makecell{ATS \cite{fayyaz2022adaptive}}& ECCV22 &\makecell{Adaptive token sampling}\\
\hline 
\multirow{4}{*}{\rotatebox[origin=c]{90}{\makecell{Token \\Pruning}}}& \makecell{SViTE \cite{chen2021chasing}}& NeurIPS21 &\makecell{Dynamically train sparse sub-networks, structured sparsity, learnable token selection}\\
\cline{2-4}
& \makecell{Dynamic ViT \cite{rao2021dynamicvit}}& NeurIPS21 &\makecell{Dynamic token sparsification}\\
\cline{2-4}
& \makecell{A-ViT \cite{yin2021adavit}}& CVPR22 &\makecell{Adaptive token reduction, distributional prior regularization}\\
\cline{2-4}
& \makecell{\cite{tang2022patch}}& CVPR22 &\makecell{Patch
slimming}\\
\hline
\multirow{9}{*}{\rotatebox[origin=c]{90}{\makecell{Others}}}& \makecell{ObjectFormer \cite{wang2022objectformer}}& CVPR22 &\makecell{High frequency features, object prototypes}\\
\cline{2-4}
& \makecell{DAT \cite{xia2022vision}}& CVPR22 &\makecell{Deformable self-attention}\\
\cline{2-4}
& \makecell{RVT \cite{mao2022towards}}& CVPR22 &\makecell{Robust ViT, position-aware attention scaling, patch-wise augmentation}\\
\cline{2-4}
& \makecell{MetaFormer \cite{yu2022metaformer}}& CVPR22 &\makecell{Spatial token pooling}\\
\cline{2-4}
& \makecell{TransMix \cite{chen2022transmix} }& CVPR22 &\makecell{Mixup Transformer}\\
\cline{2-4}
& \makecell{TokenMix \cite{liu2022tokenmix}}& ECCV22 &\makecell{Token-level mixup, content-based activation maps}\\
\cline{2-4}
& \makecell{TokenMixup \cite{choi2022tokenmixup} }& NeurIPS22 &\makecell{Saliency-aware augmentation, curriculum learning, vertical TokenMixup}\\
\cline{2-4}
& \makecell{CF-ViT \cite{chen2023general}}& AAAI23 &\makecell{Coarse inference stage, fine inference stage}\\
\hline
\end{tabular}
\label{tab_vit}
\end{table*}

\subsection{Token Pruning}

The ViT is usually computationally expensive with a large model size. It is observed that some tokens have marginal influences on the performance. Therefore, some token pruning mechanisms are designed to adaptively prune redundant tokens, accelerating the ViT.

Despite its explosive popularity, the ViT generally has a tremendous models size and a high training budget. Prior post-training pruning can reduce the model size, but the training cost remains high. Sparse vision Transformer exploration (SViTE) \cite{chen2021chasing} explores sparsity in the ViT in an end-to-end way. Specifically, it has the following three merits: 1) It dynamically train sparse subnetworks instead of the full models with a fixed parameter budget; 2) Structured sparsity is explored by a first-order importance approximation to guide the prune-and-grow of self-attention head inside ViTs; 3) A novel learnable token selection method determines the most important patch embeddings.

It is observed in \cite{rao2021dynamicvit} that ViTs only leverage a subset of most important tokens. Motivated by this, a dynamic token sparsification network (dynamic ViT) is proposed to prune redundant tokens progressively and dynamically in an input-dependent way. Specifically, a module is designed to predict the importance for each token, which is applied hierarchically to prune redundant tokens. To train the module in an end-to-end way, an attention mask differentiably prunes a token by blocking its interactions with other tokens.

Deploying ViTs on edge devices is challenging because they are computationally expensive. A-ViT \cite{yin2021adavit} can adaptivley adjust the amount of token computation without modifying the network or inference cost. Besides, a distributional prior regularization is introduced to guide token halting towards a specific distribution, stabilizing the training. 

To reduce the high computational cost of the ViT, a novel patch
slimming approach \cite{tang2022patch} can remove useless patches in a top-down way. Specifically, important patches are identified in the last layer which are used to select important patches and discard useless ones in previous layers.

\subsection{Others}

Other modules target at learning discriminative local features in the ViT, such as deformable attentions, data augmentations.

It is expected that image manipulation detection should not only check whether some pixels are out of distribution, but also examine whether objects are consistent with each other. Further, visual artifacts are not perceptible in the RGB domain. In ObjectFormer \cite{wang2022objectformer}, high frequency features are extracted and combined with RGB features to provide multimodal patch embeddings. Further, a set of object prototypes are learned to model object-level consistency among different regions.

Swin Transformer \cite{liu2021swin} or PVT \cite{wang2021pyramid} can reduce the computation complexity in ViT. However, their mechanisms are data-agnostic, possibly missing important regions and preserving less informative ones. A naive implementation of adopting deformable attentions in Transformers usually has a high memory/computation complexity. Deformable Attention Transformer (DAT) \cite{xia2022vision} proposes a deformable self-attention module which learns sparse attentions and geometric transformations in a data-dependent way, capturing important regions.

Few existing works investigate the robustness of the ViT. Robust Vision
Transformer (RVT) \cite{mao2022towards} systematically analyzes the components of ViT and revises components to build robust ViT. Further, the position-aware attention scaling and patch-wise augmentation are proposed to augment the model.

It is hypothesized that the general architecture of Transformers, instead of the carefully designed token mixer approach, is more important to the performance. To verify this, MetaFormer is designed \cite{yu2022metaformer}. Specifically, the spatial pooling is used to conduct token mixing. Although it is simple, the spatial token pooling can achieve decent performance on various vision tasks.

As one of the most successful data augmentation methods, mixup can augment the input by combining two random instances and reassigning the ground truth \cite{zhang2018mixup}. It can enable the ViTs to address the overfitting issue, however, it is possible that no objects exist due to the random augmentation but there is still label response. To this end, TransMix \cite{chen2022transmix} can assign labels under the guidance of attention maps in the ViT. TokenMix \cite{liu2022tokenmix} can mix two images at the token level by dividing the mixing region into different separate splits. Besides, the target labels are generated with content-based activation maps of the two input images. Recent mixup approaches focus on mixing with saliency. However, these saliency-aware mixup methods are computationally heavy, which is especially serious for parameter-heavy Transformer models. To address this problem, an attention-guided token-level data augmentation, called TokenMixup, is proposed in \cite{choi2022tokenmixup}. Concretely, the saliency-aware data augmentation is used to achieve 15$\times$ speed-up. Besides, curriculum learning is adopted to enable adaptive augmentation based on sample difficulty. Furthermore, vertical TokenMixup is introduced to perform mixup within a single sample and allow multi-scale feature augmentation.

The spatial dimension of the input image has considerable redundancy, resulting in heavy computational cost. A coarse-to-fine vision transformer (CF-ViT) is proposed in \cite{chen2023general} which implements inferences in a two-stage manner. In particular, in the coarse stage, an input image is split into a small-length patch sequences. If needed, the discriminative patches are further split into fine-grained granularity at the fine stage.

\subsection{Discussions}

Representative Vision Transformers with local mechanisms are summarized in Table \ref{tab_vit}. There are a number of different local mechanisms in the ViT. 1) CNN hybrids: they incorporate the convolution operation into the ViT, providing complementary low-level and local features with limited extra computational cost. 2) Window splits: Original images are split into different windows, followed by learning detailed information within windows and building long-range relations across windows. 3) Token selection: Different image tokens contribute differently. To address this issue, token selection is proposed to adaptively select important tokens and discard useless ones. 4) Token pruning: It can accelerate the ViT by pruning redundant tokens. 5) Other modules, including deformable attentions and data augmentation, can also learn important part features.

\section{Self-Supervised Learning}\label{sec_ssl}

Deep learning has demonstrated an excellent performance on various tasks, especially  supervised learning on image classification \cite{deng2009imagenet,he2016deep}, object detection \cite{ren2015faster,lin2017focal,lin2017feature}, semantic segmentation \cite{long2015fully,he2017mask}, and face recognition \cite{schroff2015facenet,deng2019arcface,wang2020hierarchical}. However, supervised learning relies heavily on expensive manual annotations and suffers from generalization errors, spurious correlations, and adversarial attacks \cite{liu2021self}. As a promising alternative, since self-supervised learning (SSL) has the capability of learning representations from unlabeled data, it has attracted widespread attentions, which has been proved to be as effective as supervised learning and even outperforms it in some scenarios \cite{caron2020unsupervised,chen2020improved}.

\begin{figure}[ht!]
\centering
\includegraphics[width=0.49\textwidth]{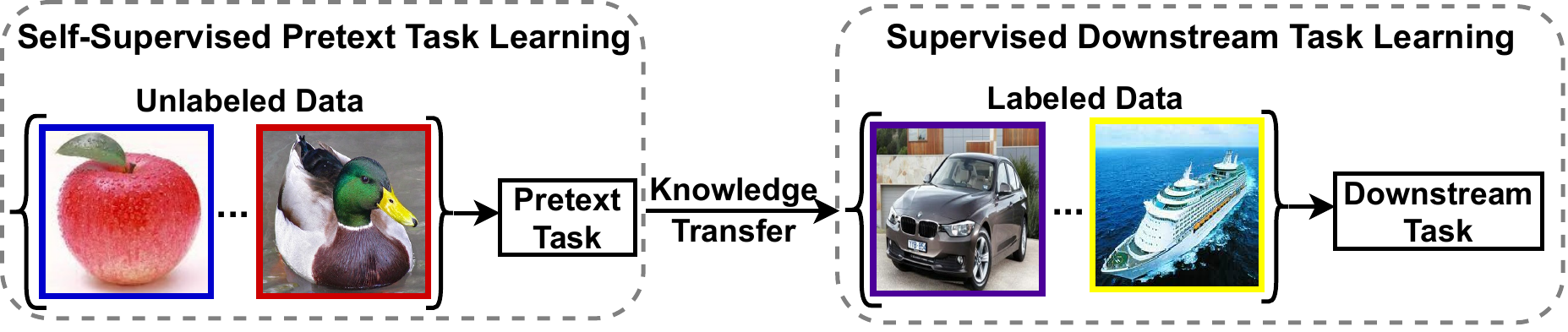}
\caption{The general pipeline of the self-supervised learning (SSL).}
\label{fig_ssl_pipeline}
\end{figure}

Fig. \ref{fig_ssl_pipeline} demonstrates the general pipeline of the SSL which consists of self-supervised pretext task learning and supervised downstream task learning. In particular, in the first stage, since there are no human-annotated labels, a pretext task is usually defined to generate pseudo labels based on human prior knowledge, making unlabeled data supervise themselves. Consequently, self-supervised features can be exploited. Then, in the second stage, the knowledge from the self-supervised model is transferred to downstream tasks by fine-tuning. The performance on downstream tasks is used to evaluate the quality of SSL. The learned representations via self-supervision tend to be more generalizable to improve the performance of various downstream tasks where the supervision fine-tuning adapts self-supervised features into specific downstream tasks.

\begin{figure}[t]
\centering
\includegraphics[width=0.48\textwidth]{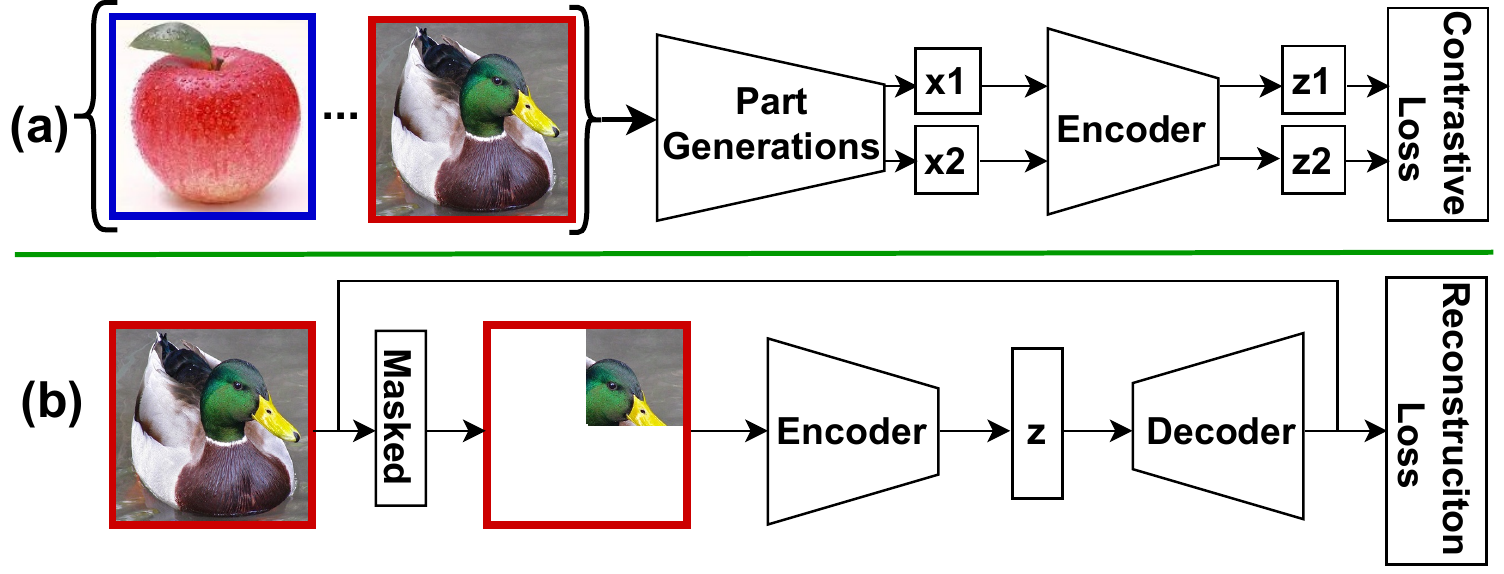}
\caption{Two typical pretext tasks in the SSL discussed in this survey where (a) is about contrastive learning and (b) refers to masked prediction.}
\label{fig_ssl}
\end{figure}

The mainstream pretext tasks can be categorized into generative and discriminative methods which explore inherent representations from unlabeled data. The discriminative methods targets at minimizing the distance of views augmented from the same image and maximizing views from two different images. Among discriminative methods, one representative methods is contrastive learning. In contrastive learning, as shown in Fig. \ref{fig_ssl} (a), part generations are used to generate a pair of part data $(x1, x2)$ which form positive pairs if they come from the same image, and constitute negative pairs if they are generated from different images. Then, their corresponding representations $(z1, z2)$ are generated, based on which contrastive loss is conducted to pull positive pairs closely and push away negative pairs. Apart from the discriminative methods, generative learning usually maps the input into latent representations with an encoder and generates the input from the latent representations with a decoder \cite{kingma2013auto, pathak2016context}. As it is receiving increasingly popular, masked prediction is discussed among generative methods in this survey, which refers to predicting the unmasked input from the masked input. In particular, as shown in Fig. \ref{fig_ssl} (b), some parts in an image are first masked. Then, an encoder learns an explicit representation $z$, based on which a decoder is used to reconstruct the unmasked input image. Details are illustrated in the following sections.

\begin{figure}[t]
\centering
\includegraphics[width=0.47\textwidth]{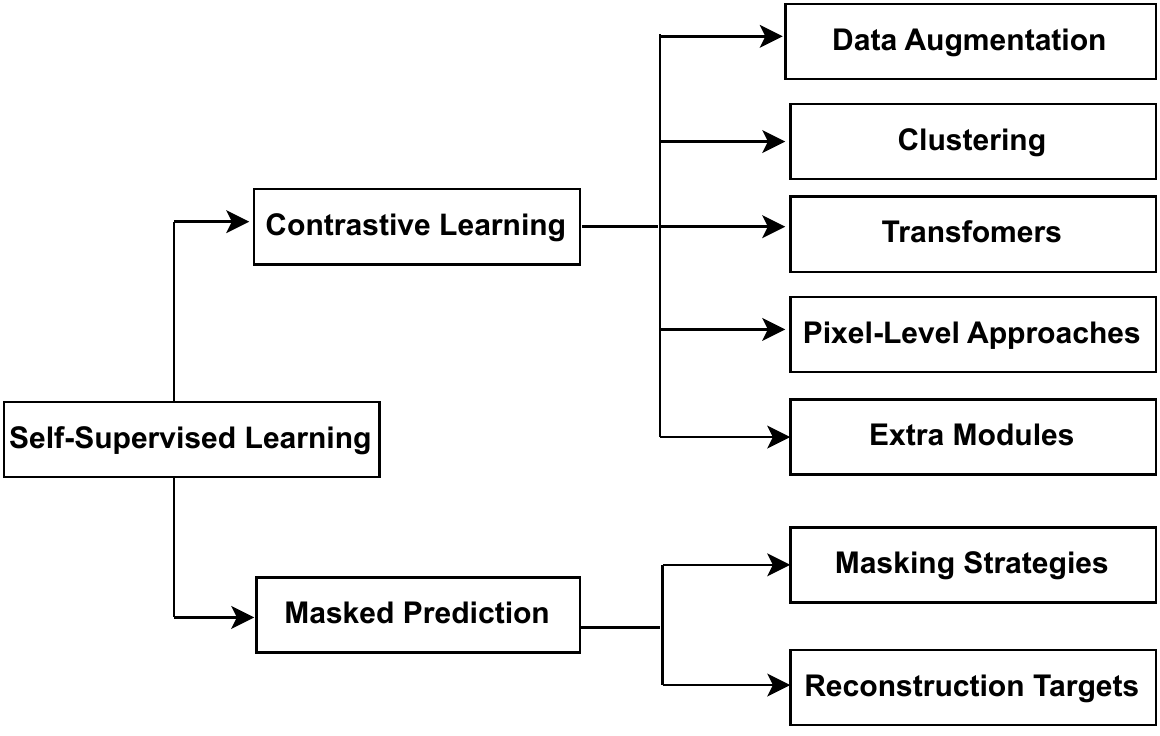}
\caption{Family of SSL methods with local mechanisms, including contrastive learning and masked prediction.}
\label{fig_ssl_family}
\end{figure}

\subsection{Contrastive Learning}

Contrastive learning is an important technique in the SSL. However, many contrastive learning approaches usually adopt global representations which can incorporate irrelevant background or spatial alignments among different augmentation views. Besides, the learned global discriminative features may lack spatial sensitivity, degrading their performance on dowstream fine-tuned tasks, especially for dense downstream tasks, like segmentation and detection. To address these issues, many local constrastive objectives and frameworks are explored, including data augmentation, clustering, part generations, pixel-level approaches, and Transformers, as shown in Fig. \ref{fig_ssl_family}.

\subsubsection{Data Augmentation}
This group of methods relies on the assumption that the input image has a canonical view, based on which some transformations are applied to change the original input. Augmented views are viewed as positive pairs and mutual information between them is maximized. The idea is to predict whether the pair of inputs belong to the same or different classes. In this setting, it is important to introduce variations among inputs of the same class. To achieve this target, many data augmentation methods are used, including random cropping, color jittering, patch reorganization, and Gaussian blurring. 

Current contrastive methods aim at learning invariant image-level representations under semantic-preserving augmentations, however, they tend to ignore spatial consistency of local representations, resulting in inferior performances on object detection or semantic segmentation that contain multiple objects. Spatially Consistent Representation Learning (SCRL) \cite{roh2021spatially} can generate consistent spatial representations of the same cropped region under different augmentation operations by minimizing the distance between spatially paired regions.

It is observed that random small-sized crops have large variance than large crops which may encode different content. However, existing works encourage random crops from the same image to have similar representations. A new strategy is proposed in \cite{zhang2022leverage}, called LoGo, that builds on local and global crops to obtain view invariances. It encourages similarity between global crops from the same image, as well as a global and local crop. Besides, it enforces local crops to be dissimilar to achieve diversity of local representations.

In deepfake detection, it is highly expected to learn a generalizable representation that is sensitive to various forgery types. In \cite{chen2022self}, augmented forgeries with various forgery settings (i.e., regions, blending types, and ratios) are generated to enrich forgery. Meanwhile, the model is pushed to predict forgery settings, strengthening the forgery sensitivity. Besides, an adversarial training generates the most difficult forgeries. 

In \cite{ci2022fast}, it is noticed that traditional contrastive-based self-supervised learning suffers from long training time to obtain decent performance. To improve the efficiency, a novel framework, Fast-MoCo, is proposed to use combinatorial patches to construct multiple positive patches from two augmented pairs. Consequently, the training time is significantly decreased with similar performance.

In previous methods, cropped regions are usually used to construct positive pairs, however, the remained regions are less investigated. In \cite{xuregioncl}, it is the first attempt to show the importance of both regions. This is achieved by designing a simple pretext task, named Region Contrastive Learning (RegionCL). Specifically, a region with the same size is cropped for each image. The cropped regions are swapped between different images to compose new images with the remaining regions. Then, positive pairs are constructed from the same original image and negative pairs are constructed from different images.


Small crops usually ignore diverse backgrounds and degrade their variances. A mosaic representation learning framework (MosRep) is proposed in \cite{wang2023mosaic} considers both diverse backgrounds and local-to-global correspondence. It composes small crops randomly selected from different images into a mosaic view, introducing diverse background information. Besides, the mosaic view is jittered to discard spatial locations.

\textbf{Pros and cons}. Within this family, methods differ in what transformation operations are used. It is necessary to design task-specific data augmentation methods, meeting the requirements of different tasks. Therefore, learned representations are not general to solve different tasks. It is important to design specific augmentation operations for each task based on prior knowledge, which is labor-intensive and time-consuming.

\subsubsection{Clustering}

This family of methods group the training data into a number of clusters with small intra-cluster and high inter-cluster variations. Clustering mainly consists of two steps: 1) Assigning data points into clusters based on the distance between the cluster and the samples; 2) Using the cluster assignments as pseudolabels to optimize the SSL model.   


Local aggregation (LA) \cite{zhuang2019local} is proposed to group similar data to move together in the embedding space and separate dissimilar instances by optimizing a local soft-clustering metric.

It is desirable to learn self-supervised dense features, making it possible to learn on large-scale generic images. Besides, unsupervised segmentation can be regarded as an efficient data labeling. However, it is difficult to guide the model to learn representations for dense tasks that correspond to different object classes. Existing methods rely on object priors, such as saliency and contour detectors to introduce object definitions into pretext tasks in a semi-supervised way, which are not generalizable. Learn object parts (Leopart) is introduced in \cite{ziegler2022self} which has a dense image patch clustering pretext task to learn semantic spatial tokens in an unsupervised manner.


Previous work usually relies on predefined specialized priors or pretext tasks to learn visual representations. To address this issue, contrastive learning from data-driven semantic slots (SlotCon) is proposed \cite{wen2022selfsupervised} for joint semantic clustering and feature learning. Semantic clustering assigns pixels to a set of learnable prototypes, adapting to each sample by attentive pooling and forming new slots. Then, contrastive learning is conducted on slots to boost the discriminativeness of features and benefit the clustering of semantically similar pixels.

\textbf{Pros and cons}. There are two unsupervised learning paradigms. In particular, it generates parts with a clustering-based pretext task, based on which form positive pairs and negative pairs to conduct contrastive learning. However, due to the complex unsupervised process and lacking of a proper supervision signal, it is difficult to analyze invariances introduced by clustering and contrastive learning approaches. Consequently, the learned representations may not be discriminative, leading to sub-optimal results on some specific tasks.

\subsubsection{Transformers}

Vision Transformers have gained much attention as a decent alternative to CNNs with better results over CNNs in various tasks. Therefore, there are some attempts to employ SSL into ViTs. Although they achieve good performance, the full advantages of the ViTs are not fully explored. For example, the ViT can process patch-level representations and build relations among different parts. Several works investigate how to combine patch-level representations from the ViT with SSL.

Self-supervised pretraining is investigated on ViTs in knowledge distillation with no labels (DINO) \cite{caron2021emerging}. There are two observations: 1) Self-supervised ViT contains semantic segmentation of an image; 2) Learned features have excellent performances with k-NN classifiers. In DINO, two global views and several local views are cropped where all crops serve as the input of student networks and global crops are passed though the teacher, encouraging local-to-global correspondences. It should be observed that cropped parts may contain useless information. Besides, semantic crops may be split into different crops, resulting semantic inconsistent local representations.

Previous methods fail to consider local features in downstream dense prediction tasks. Masked Self-supervised Transformer (MST) \cite{li2021mst} can extract the local context of an image without losing holistic semantic features. Specifically, a masked token strategy filters out some tokens of local patches while preserving the important structure in the SSL. Masked tokens are recovered by a global encoder with remained tokens, keeping important spatial information for downstream dense tasks. However, cropped parts may have semantically dissimilar information. It is sub-optimal to conduct contrastive learning on semantically dissimilar parts. 

The attention of CNN network is under-explored to limit the performance. A CNN Attention REvitalization (CARE) framework is proposed \cite{ge2021revitalizing} to train CNN encoders under the guidance of Transformers. The attention is learned with Transformers to benefit the CNN in the SSL. 

To adopt contrastive learning on semantic discriminative crops, SelfPatch \cite{yun2022patch} regards semantically similar neighboring patches as positives. Specifically, a fixed number of adjacent patches are selected as positives using cosine similarity, then an attention-based aggregation module summarizes patches to remove noisy patches. Lastly, the distance between each patch and the corresponding summarized ones is minimized.

Data mixing is effective to boost the performance of various models, however, it remains unexplored in the self-supervised ViT. Simple Data
Mixing Prior (SDMP) \cite{ren2022simple} is the first attempt in this field. It not only captures relationships between source images and mixed ones, but also builds connections between mixed samples to boost the self-supervised feature learning. 

Except the aforementioned areas, self-supervised ViT is also applied into video action recognition and person re-identification. Self-supervised Video Transformer (SVT) \cite{ranasinghe2022self} first samples local views with different spatio-temporal windows and global views with varying temporal steps. Then, it learns cross-view correspondence between local and global views, and motion correspondence between global views. Besides, slow-fast training is supported to handle inputs generated with different frame rates by dynamic positional encoding. Current SSL methods for person re-identification directly apply image classification-based approaches without any adaptions. This could lead to sub-optimal results because of dataset domain gaps and missing local details. Part-Aware Self-Supervised pre-training (PASS) \cite{zhu2022part} is a ReID-specific Transformer method to learn fine-grained features. Images are firstly divided into multiple local regions. Local views are randomly cropped from each region are assigned with a specific learnable token which corresponds to a specific fine-grained information of this area.

\textbf{Pros and cons}. Motivated by the success of the ViT, the aforementioned methods attempt to apply contrastive learning into the ViT. However, the advantages of the ViT are not fully explored. It is worth generating high-quality local positive pairs and negative pairs, utilizing the powerful capability of the ViT to boost the development.

\subsubsection{Pixel-Level Approaches}

Most existing works focus on image-level or part-level representations which are sub-optimal for dense prediction tasks, such as image segmentation, object detection. Therefore, pixel-level approaches are designed to conduct contrastive learning on pixel levels to benefit dense tasks.

Dense Contrastive Learning (DenseCL) \cite{wang2021dense} can optimize a pairwise contrastive similarity loss at the pixel level on dense vectors between two views of input images.

There exists a gap between pre-trained SSL model and downstream dense prediction tasks. Dense Semantic Contrast (DSC) \cite{li2021dense} can model semantic category decision boundary at the pixel level for semantic segmentation. Besides, a dense cross-image semantic contrastive framework learns low-, middle-, and high-level representations.

Learning parts with supervised learning needs human annotations, which are labor-expensive, slow, and infeasible to collect for various real-world objects. On the other hand, unsupervised part learning is receiving attentions of the community. An unsupervised approach is proposed in \cite{choudhury2021unsupervised} to discover object parts. First, contrastive learning is constructed at pixel-level to guide models to decompose images into parts. Second, pixel-level reconstructions provide complementary cues to find meaningful parts. Third, new evaluation protocols are introduced to assess the automated part discovery.

Object detection needs to locate objects in an image and recognize their corresponding semantic category. For localization and recognition, it is expected to perform at the pixel-level and region-level contrastive learning. In \cite{bai2022point}, a pixel-level region contrast is proposed to balance localization and recognition. It is implemented by sampling point pairs from individual regions, learning both inter-image and intra-image distinctions.

\textbf{Pros and cons}. These approaches are specifically designed to learn pixel-level self-supervised representations for dense tasks, which have sub-optimal results on recognition tasks. If there is a good balance between pixel-level, part-level, and object-level contrastive learning objectives, different granularity information can be learned for coarse-grained and fine-grained tasks.

\subsubsection{Extra Modules}

Extra modules can discover useful object parts via saliency, attentions, segmentation, or response maps in CNNs, based on which part-level contrastive learning is conducted.

Part-based representations are robust to occlusions, pose, or viewpoint variations. Self-Supervised Co-Part Segmentation (SCOPS) \cite{hung2019scops} is a self-supervised method for part segmentation. This is achieved by devising several losses on part response maps: 1) Geometric concentraion loss enables parts to concentrate geometrically and form connected parts; 2) Equivariance loss makes models robust to appearance and pose variations; 3) Semantic consistency loss encourages semantic consistency across different objects.  

Learning dense semantic representations without supervision remains unexplored on large-scale datasets. Without relying on proxy tasks or end-to-end clustering, MaskContrast \cite{van2021unsupervised} adopts an unsupervised saliency estimator to explore object mask proposals. This mid-level visual prior can incorporate semantically meaningful representations for unsupervised semantic segmentation.

Current instance discrimination methods only optimizes global features, however, spatial consistency cannot be met, which is problematic for object detection. Region Similarity representation learning (ReSim) \cite{xiao2021region} can learn spatially consistent features across different convolutional layers. Specifically, it performs global and region-level contrastive learning by maximizing similarity across views and sub-images, respectively.

Recent most works in semantic segmentation exploit intra-image correlations, however, semantic relations prevail in the whole dataset. A novel region-aware contrastive learning (RegionContrast) \cite{hu2021region} uses contrastive learning for the semantic segmentation problem. A segmentation head is used on feature maps to generate parts. Firstly, a supervised contrastive learning explores semantic relations. Secondly, region-level centers are constructed to store the information of the whole dataset, based on which contrastive learning is performed.

To adaptively aggregate spatial features, learning where to learn (LEWEL) is proposed in \cite{huang2022learning}. Specifically, a novel reinterpretation scheme generates a set of alignment maps, adaptively aggregating spatial information for SSL. Besides, the coupled projection head encourages aligned and global embeddings to be reciprocal to each other.

Recent works on dense SSL can be divided into pixel-based and geometric-based where the former can cause incorrect correspondences and the latter may have spatial semantic inconsistency. Set Similarity (SetSim)  \cite{wang2022exploring} proposes set-wise similarity across two views. Specifically, attentional features of views are used to build the set which can remove noisy background and preserve the coherence of the same image, based on which the contrastive learning is employed. Besides, the cross-view nearest neighbors of sets are explored to boost the structure neighbor information.

Most recent SSL methods focus on learning either global features with invariance properties or local representations. VICRegL \cite{bardes2022vicregl} is proposed to learn global and local features simultaneously. Concretely, the VICReg criterion is employed on pairs of global feature vectors and pairs of local feature vectors from the pooling layer simultaneously. 

\textbf{Pros and cons}. It is complex and heavily sensitive to establish the correspondence among patches or pixels. Some methods, like segmentation and saliency detection, implicitly assumes that these regions tend to be discriminative regions. This assumption, however, is not always true for different scenarios. Besides, the performance heavily depends on the extra modules to generate discriminative parts. Consequently, the performance is limited if the extra modules perform poorly. Moreover, extra modules require intense computations. This is especially burdensome for training laborious SSL models, as they generally retain high training time.

\begin{table*}[ht!]
\centering
\caption{Representative SSL models with local mechanisms.}
\begin{tabular}{|c|c|c|c|c|}
\hline
\multicolumn{2}{|c|}{\multirow{1}{*}{Category}} & \multirow{1}{*}{Methods} & \multirow{1}{*}{Venue} & \multirow{1}{*}{Highlights} \\
\hline
\multirow{32}{*}{\rotatebox[origin=c]{90}{\makecell{Contrastive Learning}}}&\multirow{7}{*}{\rotatebox[origin=c]{90}{\makecell{Data \\Augmentation}}}&SCRL \cite{roh2021spatially}& \makecell{CVPR21} &\makecell{Spatially consistent local feature learning}\\
\cline{3-5}
&& \makecell{LoGo \cite{zhang2022leverage} }&CVPR22  &\makecell{Dissimilarity among local crops, \\similarity between global crops from the same image and a global and local crop}\\
\cline{3-5}
&& \cite{chen2022self}& \makecell{CVPR22} &\makecell{A synthesizer and adversarial training }\\
\cline{3-5}
&& Fast-MoCo \cite{ci2022fast}& \makecell{ECCV22} &\makecell{Combinatorial patches}\\
\cline{3-5}
&& RegionCL \cite{xuregioncl}& \makecell{ECCV22} &\makecell{Region swapping between images}\\
\cline{3-5}
&& MosRep \cite{wang2023mosaic}& \makecell{ICLR23} &\makecell{Mosaic view augmentation and jittering }\\
\cline{2-5}
 &\multirow{3}{*}{\rotatebox[origin=c]{90}{\makecell{Clustering}}}& \makecell{LA \cite{zhuang2019local}}&ICCV19 &\makecell{A local soft-clustering metric}\\
\cline{3-5}
&& \makecell{Leopart \cite{ziegler2022self}}&CVPR22 &\makecell{Patch clustering pretext task}\\
\cline{3-5}
&& \makecell{SlotCon \cite{wen2022selfsupervised}}&NeurIPS22 &\makecell{Semantic clustering, \\data-dependent slots}\\
\cline{2-5}
&\multirow{7}{*}{\rotatebox[origin=c]{90}{\makecell{Transformers}}}&DINO \cite{caron2021emerging}& \makecell{ICCV21} &\makecell{Semantic segmentation features, good k-NN performance, global and local crops }\\ 
\cline{3-5}
&&\makecell{MST \cite{li2021mst}}&NeurIPS21 &\makecell{A masked token strategy, a global encoder}\\
\cline{3-5}
&&\makecell{CARE \cite{ge2021revitalizing}}&NeurIPS21 &\makecell{Transformer and CNN streams}\\
\cline{3-5}
&&SelfPatch \cite{yun2022patch}& \makecell{CVPR22} &\makecell{Semantically similar neighbors}\\
\cline{3-5}
&& \makecell{SDMP \cite{ren2022simple}}&CVPR22 &\makecell{Self-supervised ViT, relationships between mixed data}\\
\cline{3-5}
&& \makecell{SVT \cite{ranasinghe2022self}}&CVPR22 &\makecell{Cross-view and motion correspondence, slow-fast training, video Transformer}\\
\cline{3-5}
&& \makecell{PASS \cite{zhu2022part}}&ECCV22 &\makecell{Learnable fined-grained tokens}\\
\cline{2-5}
&\multirow{5}{*}{\rotatebox[origin=c]{90}{\makecell{Pixel-Level \\Methods}}}& DenseCL \cite{wang2021dense}& \makecell{CVPR21} &\makecell{Pixel-level pairwise contrastive loss}\\
\cline{3-5}
&& DSC \cite{li2021dense}& \makecell{MM21} &\makecell{Semantic category decision boundary, dense cross-image semantic contrastive}\\
\cline{3-5}
&& \cite{choudhury2021unsupervised}& \makecell{NeurIPS21} &\makecell{Contrastive learning-based part decomposition, \\pixel-level reconstructions, automated part discovery assessment}\\
\cline{3-5}
&& \cite{bai2022point}& \makecell{CVPR22} &\makecell{Pixel-level region contrast }\\
\cline{2-5}
&\multirow{7}{*}{\rotatebox[origin=c]{90}{\makecell{Extra \\Modules}}}& \makecell{SCOPS \cite{hung2019scops}}&CVPR19 &\makecell{Geometric concentraion loss, equivariance loss, semantic consistency loss}\\
\cline{3-5}
&& MaskContrast \cite{van2021unsupervised}& \makecell{ICCV21} &\makecell{Unsupervised saliency estimator}\\
\cline{3-5}
&&ReSim \cite{xiao2021region}& \makecell{ICCV21} &\makecell{Global and region-level contrastive learning}\\
\cline{3-5}
&&RegionContrast \cite{hu2021region}& \makecell{ICCV21} &\makecell{Supervised and region-level contrastive learning}\\
\cline{3-5}
&& \makecell{LEWEL \cite{huang2022learning}}&CVPR22 &\makecell{Reinterpretation scheme, coupled projection}\\
\cline{3-5}
&& \makecell{SetSim \cite{wang2022exploring}}&CVPR22 &\makecell{Set-wise contrastive learning, cross-view nearest neighbors of sets}\\
\cline{3-5}
&& \makecell{VICRegL \cite{bardes2022vicregl}}&NeurIPS22 &\makecell{VICReg criterion on global and local features}\\
\hline
\multirow{19}{*}{\rotatebox[origin=c]{90}{\makecell{Masked Prediction}}}& \multirow{10}{*}{\rotatebox[origin=c]{90}{\makecell{Masking Strategies}}} &SimMIM \cite{xie2022simmim} & \makecell{CVPR22} &\makecell{Random masking, raw pixel regression}\\
\cline{3-5}
& &AttMask \cite{kakogeorgiou2022hide} & \makecell{ECCV22} &\makecell{Attention-guided masking}\\
\cline{3-5}
& &SemMAE \cite{li2022semmae} & \makecell{NeurIPS22} &\makecell{Self-supervised semantic part learning, semantic-guided masking}\\
\cline{3-5}
& &ADIOS \cite{shi2022adversarial} & \makecell{ICML22} &\makecell{Masking function, image encoder, an adversarial objective}\\
\cline{3-5}
& &PACMAC \cite{prabhu2022adapting} & \makecell{NeurIPS22} &\makecell{Attention-conditioned masking}\\
\cline{3-5}
& &VideoMAE \cite{tong2022videomae} & \makecell{NeurIPS22} &\makecell{Video tube masking, a high masking ratio}\\
\cline{3-5}
& &MCMAE \cite{gao2022mcmae} & \makecell{NeurIPS22} &\makecell{Multi-scale hybrid convolution-transformer, MAE}\\
\cline{3-5}
& &CIM \cite{fang2023corrupted} & \makecell{ICLR23} &\makecell{An auxiliary generator, an enhancer network}\\
\cline{3-5}
& &MixedAE \cite{chen2023mixed} & \makecell{CVPR23} &\makecell{Recognizing homologous patches, object-aware self-supervised pre-training}\\
\cline{3-5}
& &TBM \cite{li2023token} & \makecell{CVPR23} &\makecell{Token boosting module}\\
\cline{2-5}
&\multirow{7}{*}{\rotatebox[origin=c]{90}{\makecell{Reconstruction \\Targets}}}&MaskCo \cite{zhao2021self}& \makecell{ICCV21} &\makecell{Contrastive mask prediction, mask prediction head}\\ 
\cline{3-5}
& &BEIT \cite{bao2022beit}& \makecell{ICLR22} &\makecell{Masked image modeling}\\
\cline{3-5}
& &MAE \cite{he2022masked}& \makecell{CVPR22} &\makecell{Asymmetric encoder-decoder}\\
\cline{3-5}
& &MaskFeat  \cite{wei2022masked} & \makecell{CVPR22} &\makecell{Space-time mask and predict}\\
\cline{3-5}
& &MFM \cite{xie2023masked} & \makecell{ICLR23} &\makecell{Masked frequency modeling}\\
\cline{3-5}
& &CIM \cite{li2023correlational} & \makecell{CVPR23} &\makecell{Correlation modeling}\\
\cline{3-5}
& &LocalMIM \cite{wang2023masked} & \makecell{CVPR23} &\makecell{Local multi-scale reconstruction}\\
\hline
\end{tabular}
\label{tab_ssl}. 
\end{table*}

\subsection{Masked Prediction}
Since it is receiving increasing attentions, we only focus on masked prediction in the generative learning. Following the assumption that context can provide useful information to infer the missing data, masked prediction firstly masks portion of an image, then reconstructs the masked part based on the visible one. There are two key steps in a typical masked prediction pipeline: masking strategy and reconstruction target where the former is how to mask and the latter is what to predict.

\subsubsection{Masking Strategies} Different from current works which require complex designs, a simple framework, named SimMIM, is proposed in \cite{xie2022simmim}. Specifically, random masking is applied on the input image with a moderately large masked patch size. Besides, predicting RGB values of raw pixels by direct regression is used. 

Masked language modeling is different from masked image modeling due to the redundancy of image tokens. Consequently, random masking tends to mask redundant tokens. To address this limitation, attention-guided masking (AttMask) \cite{kakogeorgiou2022hide} is proposed to decide which tokens to mask. In particular, a teacher transformer encoder generates an attention map, guiding masking for the student. Similarly, semantic decomposition of an image makes masked autoencoding different between vision and language. To obtain the visual analogue of words, part-based image representations are investigated in SemMAE \cite{li2022semmae}. Specifically, self-supervised semantic part learning is employed to generate semantic parts. Besides, semantic-guided masking can learn promising image representations by incorporating semantic information.

It is what is actually masked, not how much is masked, which is important for self-supervised representation learning. To this end, Adversarial Inference-Occlusion Self-supervision (ADIOS) \cite{shi2022adversarial} is proposed which simultaneously trains a masking function and an image encoder with an adversarial objective by identifying and filtering out regions of related pixels within an image. 

Instead of masking the input image randomly, the attention mechanism in the ViT is leveraged in Probing Attention-Conditioned Masking Consistency (PACMAC) \cite{prabhu2022adapting} where attention-conditioned masking is performed to generate a set of disjoint masks corresponding to highly responded regions in the input image. Then, the predictive consistency is probed across a set of the generated partial target inputs.

It is generally required to pre-train video Transformers on large-scale datasets to obtain favorable performance on small-sized datasets. To this end, video masked autoencoders (VideoMAE) \cite{tong2022videomae} are proposed to conduct video tube masking with a high masking ratio, making this task challenging. Consequently, this can encourage models to learn representative features and address the information leakage issue.

To explore whether the performance can be improved with the multi-scale backbone and local and global operations, MCMAE is proposed in \cite{gao2022mcmae} by designing multi-scale hybrid convolution-transformer architecture via the mask auto-encoding scheme. 

Instead of applying artificial tokens to corrupt a portion of non-overlapped patches in MIM, Corrupted Image Modeling (CIM) \cite{fang2023corrupted} uses a trainable BEiT \cite{bao2021beit} as an auxiliary generator to corrupt the input where some patches are replaced with alternatives from the BEiT output. An enhancer network is used to either recover all image pixels, or distinguish the corrupted patches from the uncorrupted ones.

The effective data augmentation strategies for the MAE remains to be further explored, which are different from the those used in constrastive learning. Mixed Autoencoder (MixedAE) \cite{chen2023mixed} studies the mixing augmentation for MAE. To address the increase of mutual information, homologous recognition serves as an auxiliary supervision, not only finding homologous patches for each patch, but also performing object-aware self-supervised pre-training.

Real-world data may be corrupted and unreliable. It is challenging to pre-train ViT with masked auto-encoder on corrupted and unreliable data. To address this issue, Token Boosting Module (TBM) is introduced in \cite{li2023token}, allowing ViT to learn robust and generalizable features.

\textbf{Pros and cons}. In masked prediction, due to the spatial feature redundancy \cite{he2022masked} in computer vision, it is important to define a proper masking strategy, including how much and where to mask. Masking too many areas makes it difficult to infer the missed regions, while masking too few areas makes it less likely to learn a representative model. It is also important to determine where to mask. Masking too much redundant information cannot benefit the learning of powerful models, while masking too much discriminative information increases the difficulty of reconstruction.

\subsubsection{Reconstruction Targets} As for the reconstruction target, beyond predicting raw pixels, several other alternatives are proposed, e.g. traditional features \cite{wei2022masked}, frequencies \cite{xie2023masked}, correlation maps \cite{li2023correlational}, or tokens \cite{bao2022beit}. 

SSL methods are mainly based on a pretext task where the semantic consistency (SC) is assumed to hold across views. The SC can be satisfied for the object-centric bias, like the ImageNet dataset. However, when extending the SSL to datasets, two cropped views may contain different objects belonging to different categories, failing to satisfy the SC. Mask Contrast (MaskCo) \cite{zhao2021self} introduces contrastive mask prediction as a novel pretext task for the SSL. In particular, contrastive mask prediction is used where the masked and unmasked features of the same region are compared. Besides, a mask prediction head is used to bridge the gap between masked and unmasked features.

Bidirectional Encoder representation from Image Transformers (BEIT) \cite{bao2022beit} propose a masked image modeling to pre-train the ViT. Specifically, the input image is first tokenized into visual tokens. Then, some image patches are masked randomly before inputting to the ViT. Last, the original image is reconstructed based on masked patches.

Masked AutoEncoders (MAE) \cite{he2022masked} mask random patches from images and reconstruct missing pixels. Specifically, an asymmetric encoder-decoder network is proposed where an encoder operates on a small part of visible patches (e.g., 25\%) and a decoder reconstructs the original image from latent representations and masked tokens. Masking a high ratio of patches reduces training time and improves accuracy.

Masked Feature prediction (MaskFeat)  \cite{wei2022masked} adopts a simple mask-and-predict idea for self-supervised pre-training of video models. It firstly randomly replaces the space-time video cubes with a mask token and regresses the representation of the masked regions which uses histograms of oriented gradients \cite{dalal2005histograms} as the target features.

Instead of randomly masking in the spatial domain with heavy spatial redundancy, Masked Frequency Modeling (MFM) \cite{xie2023masked} shifts attention to the frequency domain which is more ideal without using extra data, extra model, or mask tokens. It suggests both high-frequency components and low-frequency elements are useful in learning discriminative features. 

Correlational Image Modeling (CIM) \cite{li2023correlational} randomly crops image parts (examples) at various scales, shapes, rotations, and transformations from an input image (context) and predicts correlation maps via a cross-attention block between the examplars and the context. 

Existing models reconstruct only at the top layer of encoder where the key lower layers are not explicitly guided and non-trivial inter-patch interactions are not well explored. LocalMIM \cite{wang2023masked} proposes local multi-scale reconstruction where the lower and upper layers reconstruct fine-scale and coarse-scale supervisions, respectively. 

\textbf{Pros and cons}. As for the reconstruction target, defining reasonable ground-truth target and measuring the differences between the prediction and ground-truth target is crucial. According to specific characteristics in each task, human prior knowledge can be used to assist designing meaningful targets. Instead of recovering all masked information, it is necessary to ignore redundant information in reconstruction targets, reducing computational overhead.

\subsection{Discussions}

The reviewed local SSL methods are demonstrated in Table \ref{tab_ssl}. The pretext task of contrastive learning is the mainstream SSL approaches in CNNs. Despite the great success with CNNs, contrastive learning tends to suffer from performance degradation on global views in the ViT due to the lack of inductive bias, which requires a stronger supervision for better pre-training performance. Therefore, many ViT based approaches focus on generating part-levels pseudo labels. Recent research attempts to reproduce the success of masked language modeling \cite{kenton2019bert} in the computer vision field of self-supervised learning (i.e., masked image modeling in the ViT). 

Besides, a comparative study should be conducted to explore the inherent properties of contrastive learning and masked prediction. Then, a possible future research direction can exploit the benefits from both methods in local perspective. Another direction is to enhance the individual properties of contrastive learning and masked prediction, respectively.

\section{Fine-Grained Visual Recognition}\label{sec_fgvr}

\begin{figure*}[ht!]
\centering
\includegraphics[width=.08\textwidth,height=.08\textwidth]{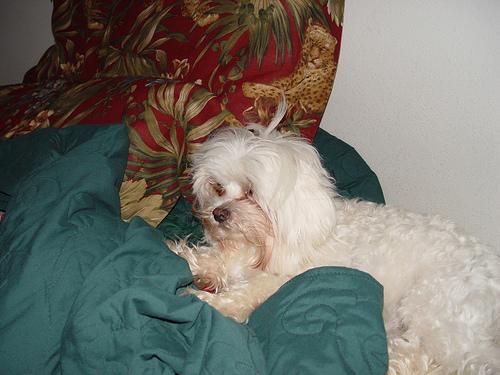}
\includegraphics[width=.08\textwidth,height=.08\textwidth]{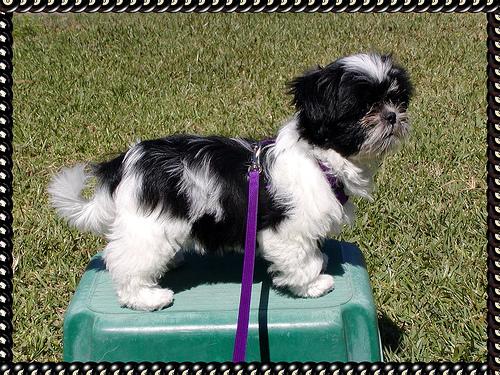}
\includegraphics[width=.08\textwidth,height=.08\textwidth]{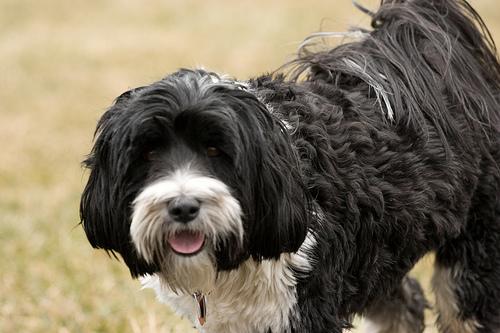}
\includegraphics[width=.08\textwidth,height=.08\textwidth]{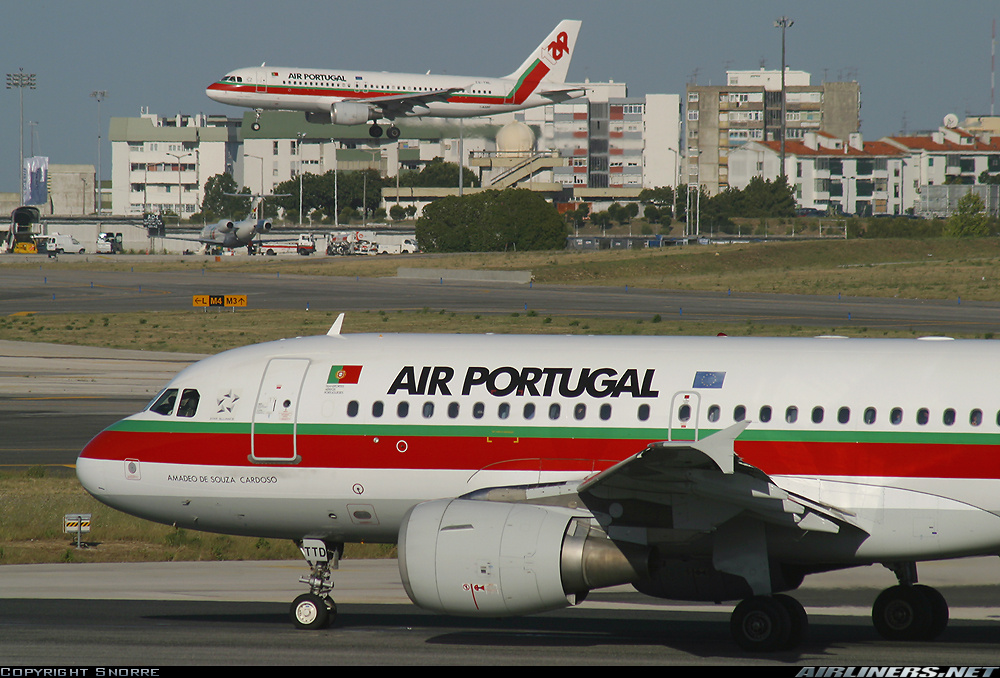}
\includegraphics[width=.08\textwidth,height=.08\textwidth]{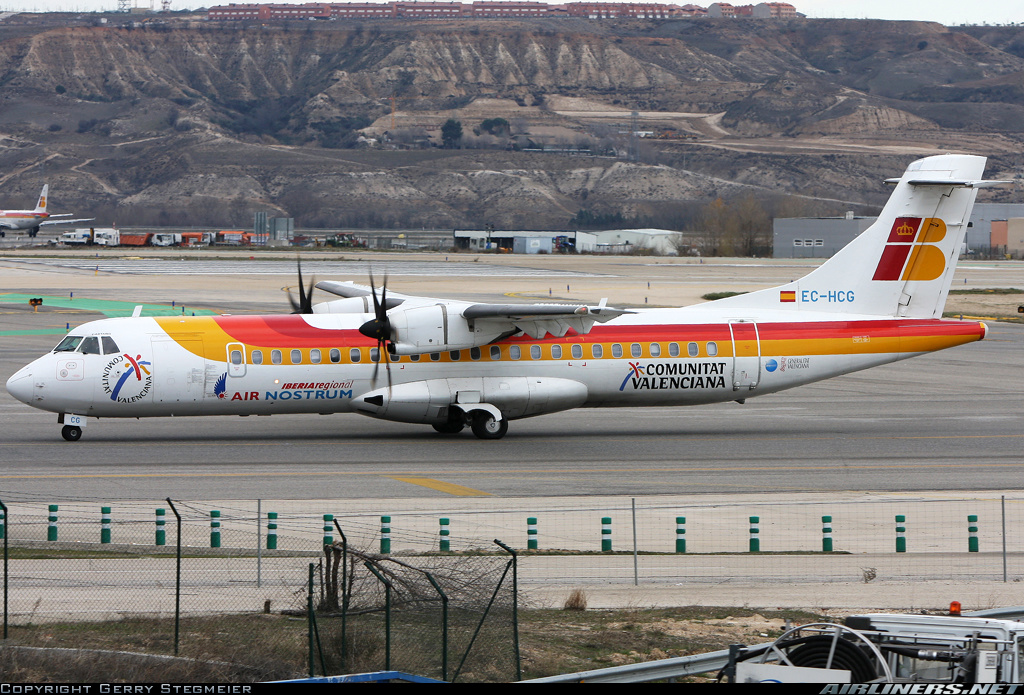}
\includegraphics[width=.08\textwidth,height=.08\textwidth]{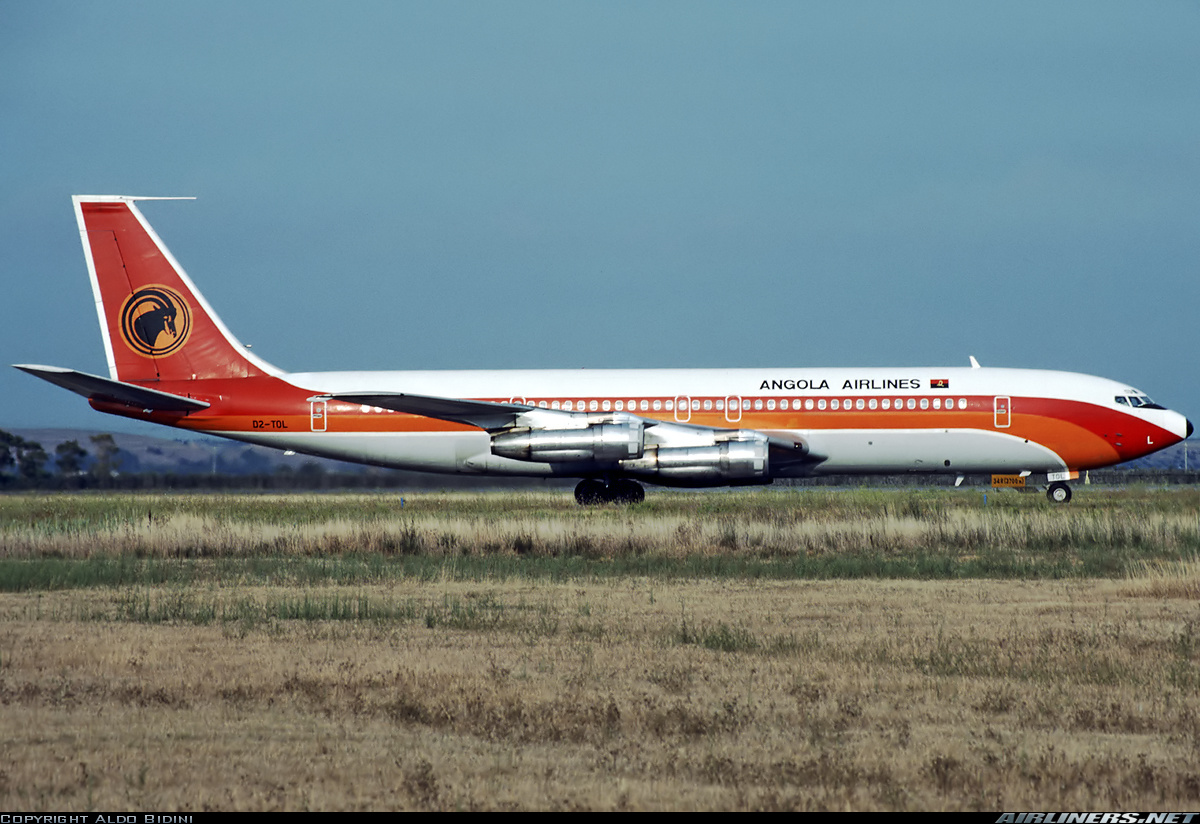}
\includegraphics[width=.08\textwidth,height=.08\textwidth]{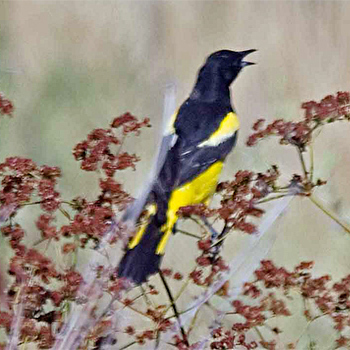}
\includegraphics[width=.08\textwidth,height=.08\textwidth]{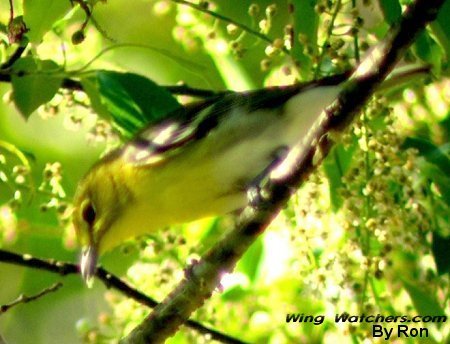}
\includegraphics[width=.08\textwidth,height=.08\textwidth]{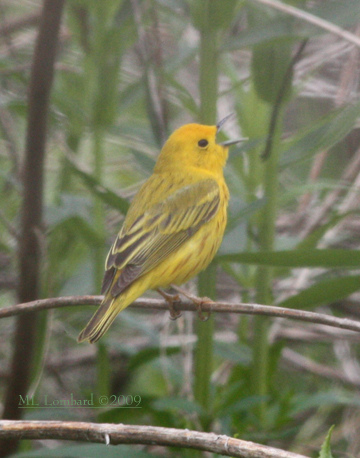}

\includegraphics[width=.08\textwidth,height=.08\textwidth]{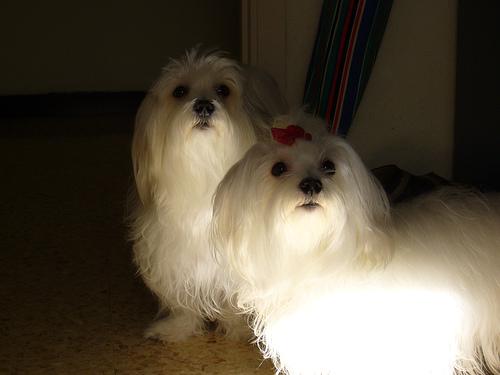}
\includegraphics[width=.08\textwidth,height=.08\textwidth]{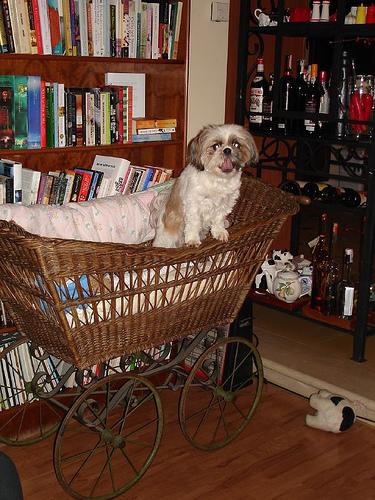}
\includegraphics[width=.08\textwidth,height=.08\textwidth]{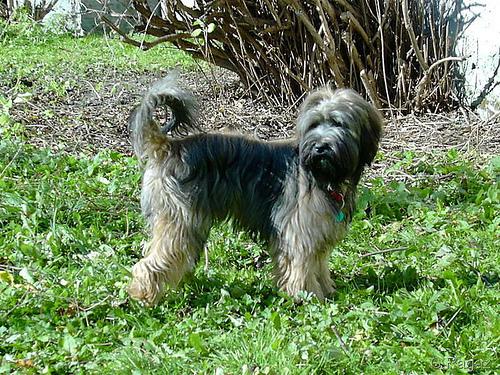}
\includegraphics[width=.08\textwidth,height=.08\textwidth]{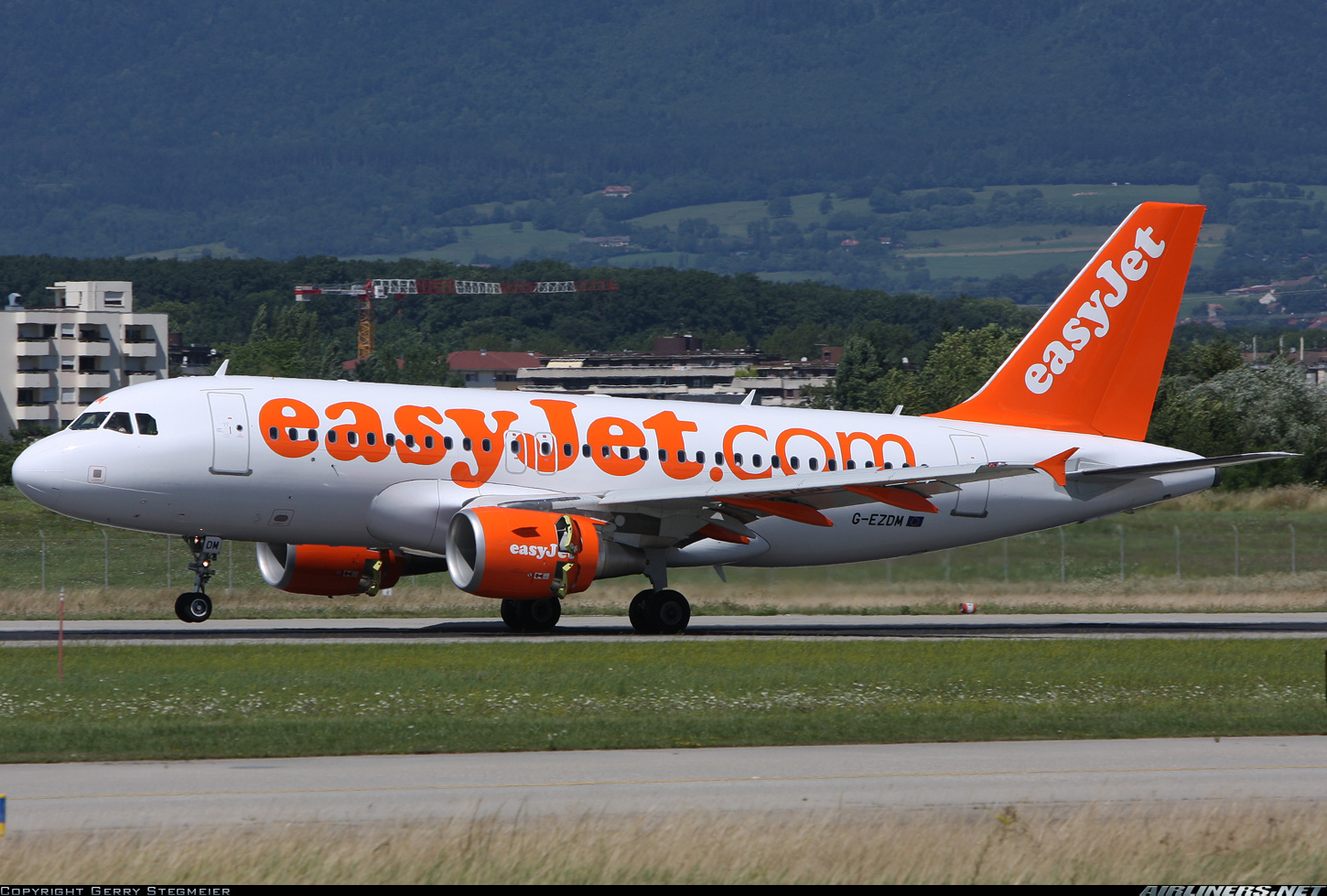}
\includegraphics[width=.08\textwidth,height=.08\textwidth]{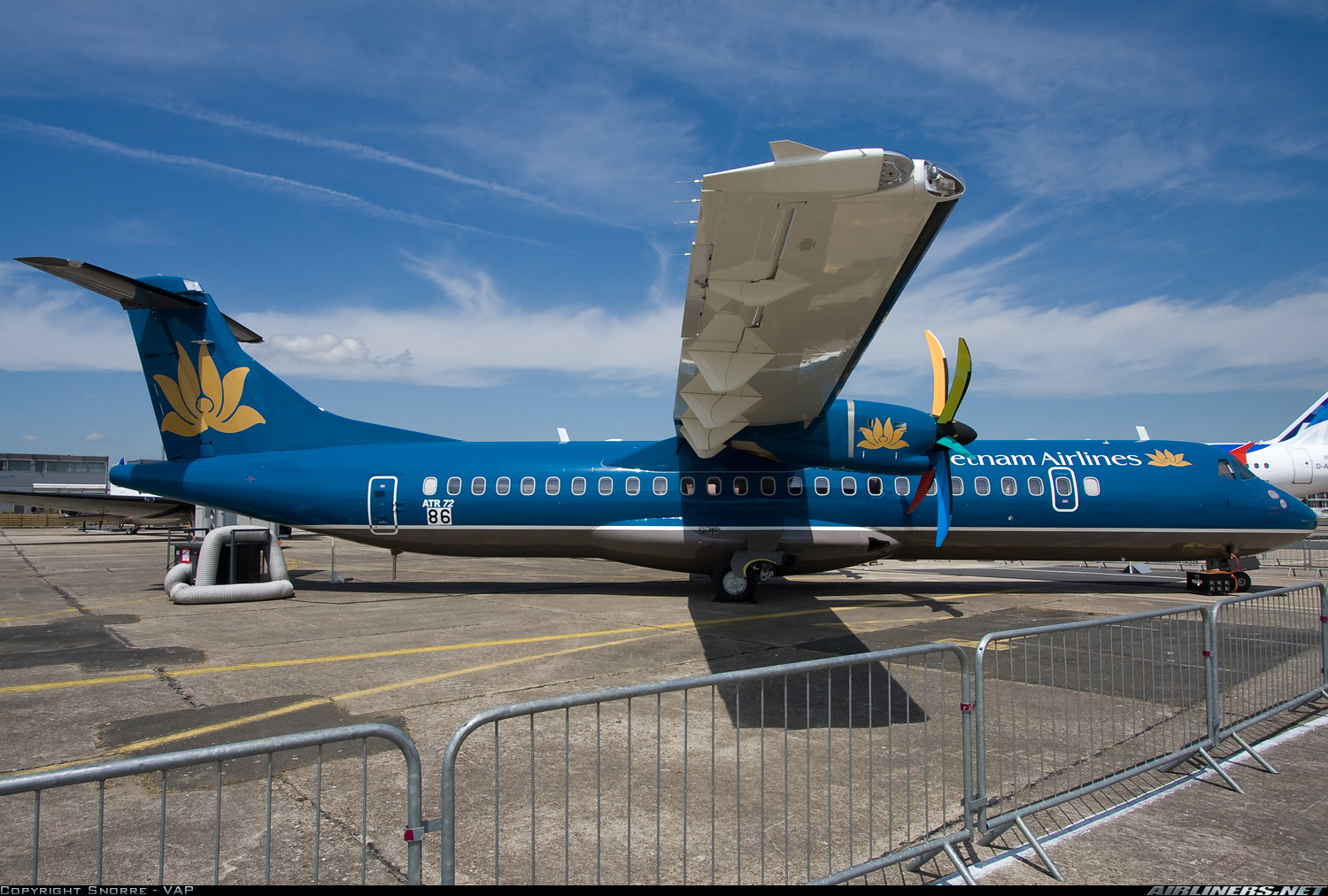}
\includegraphics[width=.08\textwidth,height=.08\textwidth]{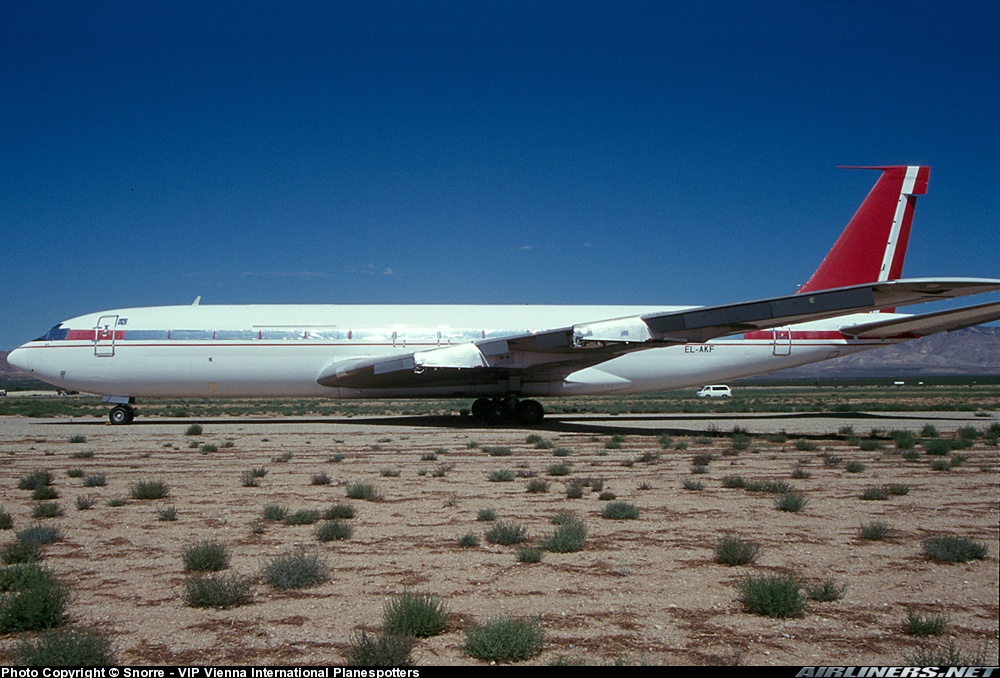}
\includegraphics[width=.08\textwidth,height=.08\textwidth]{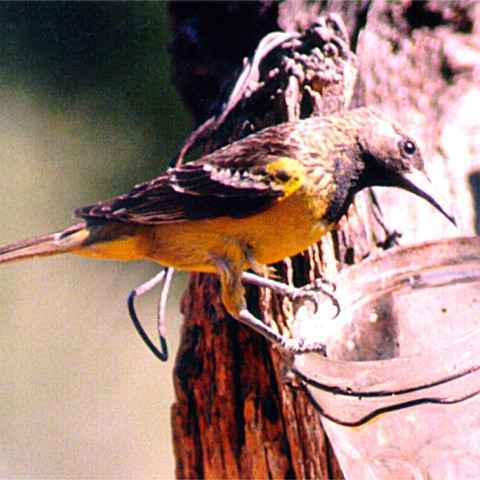}
\includegraphics[width=.08\textwidth,height=.08\textwidth]{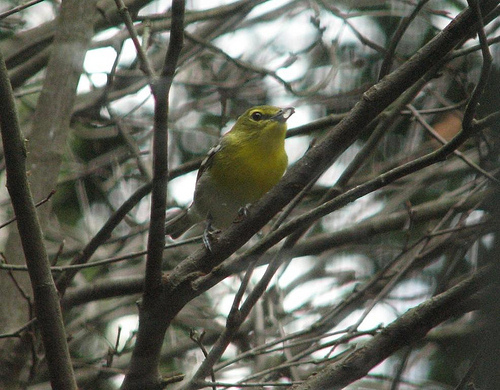}
\includegraphics[width=.08\textwidth,height=.08\textwidth]{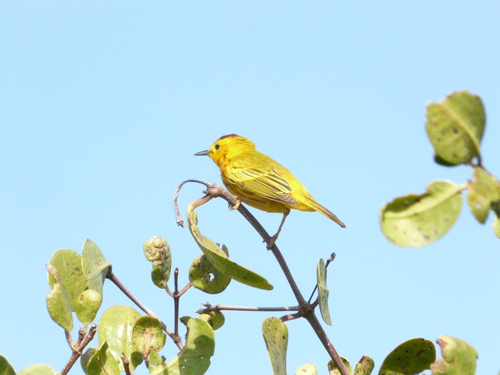}

\includegraphics[width=.08\textwidth,height=.08\textwidth]{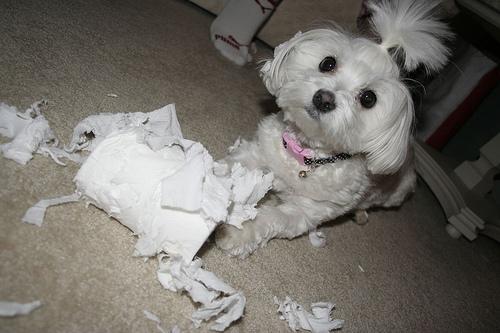}
\includegraphics[width=.08\textwidth,height=.08\textwidth]{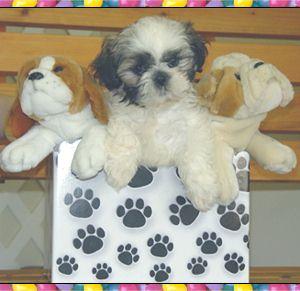}
\includegraphics[width=.08\textwidth,height=.08\textwidth]{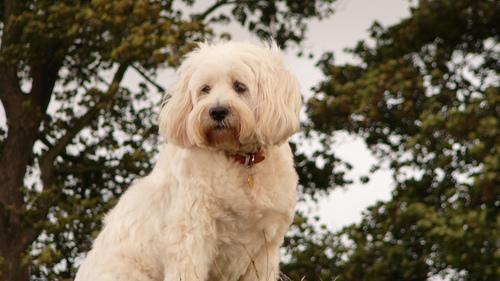}
\includegraphics[width=.08\textwidth,height=.08\textwidth]{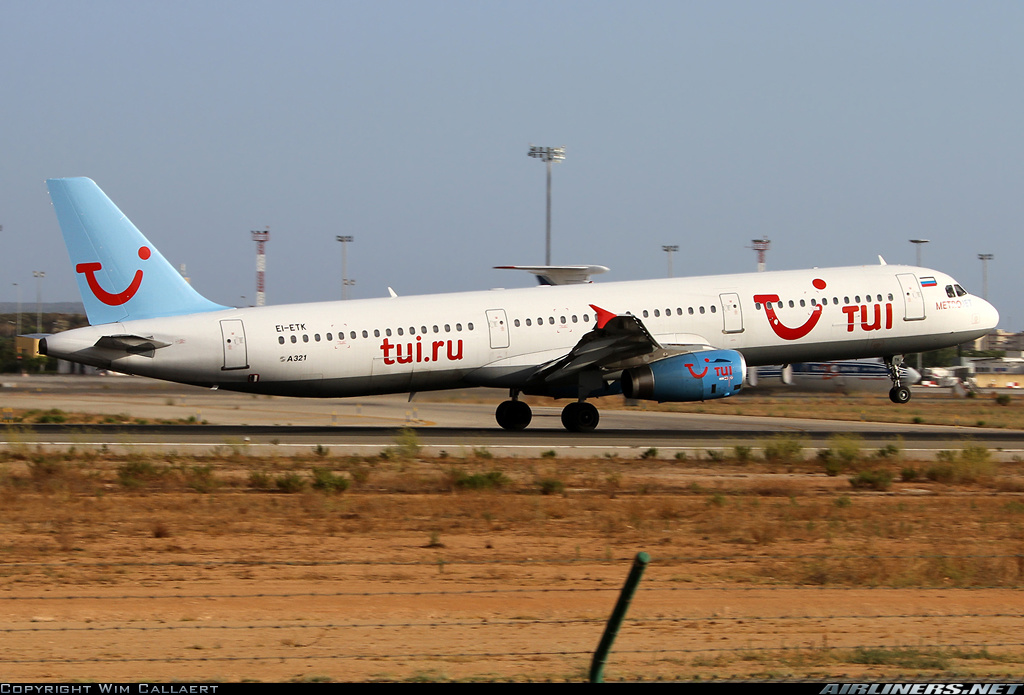}
\includegraphics[width=.08\textwidth,height=.08\textwidth]{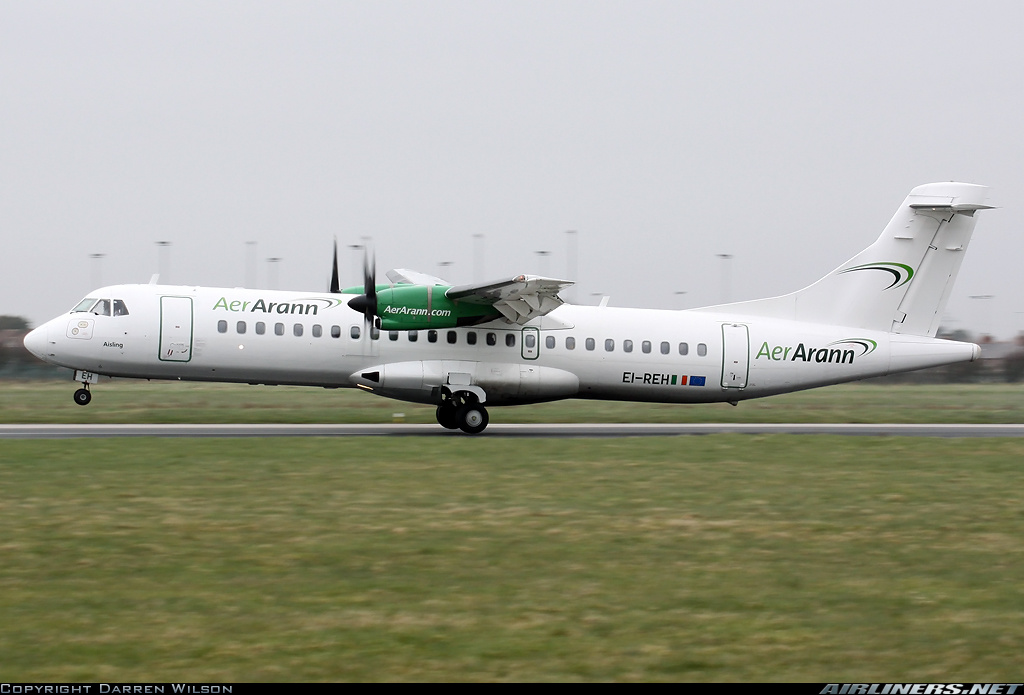}
\includegraphics[width=.08\textwidth,height=.08\textwidth]{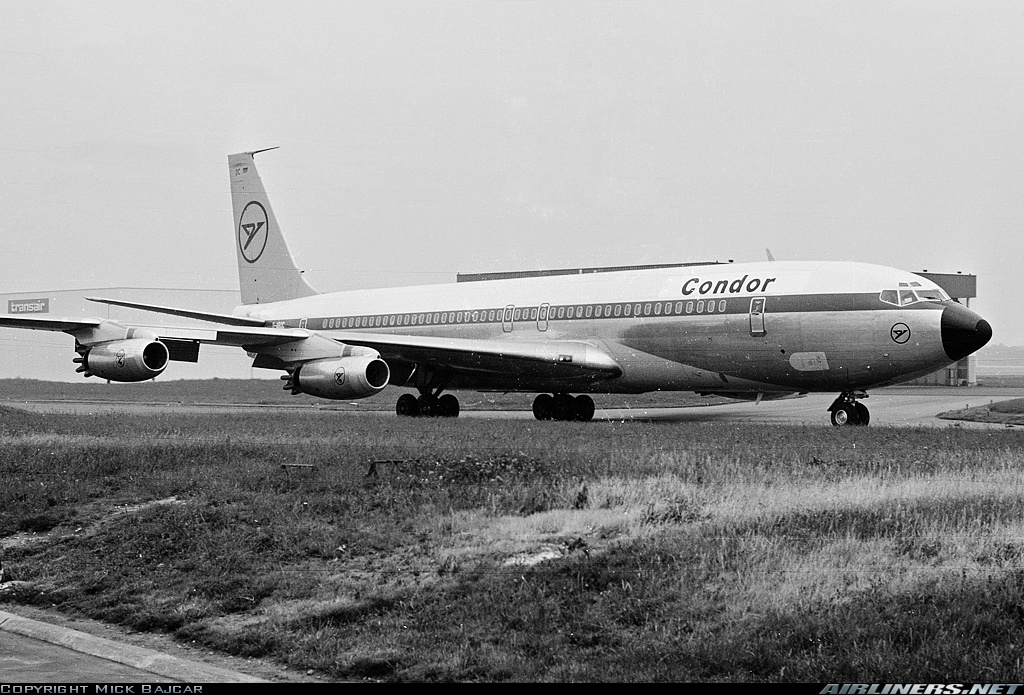}
\includegraphics[width=.08\textwidth,height=.08\textwidth]{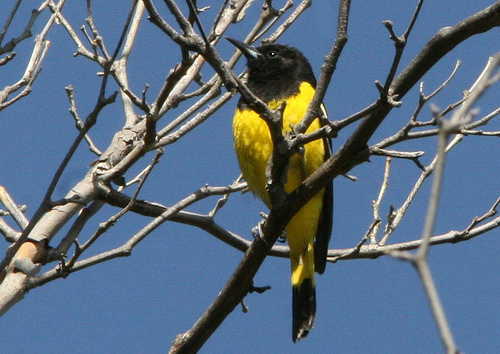}
\includegraphics[width=.08\textwidth,height=.08\textwidth]{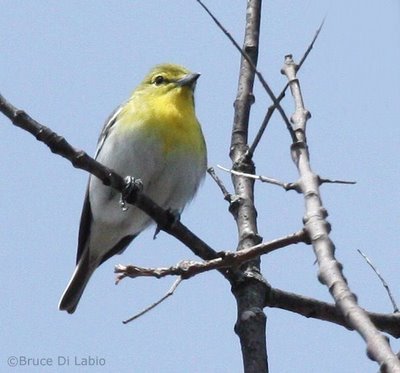}
\includegraphics[width=.08\textwidth,height=.08\textwidth]{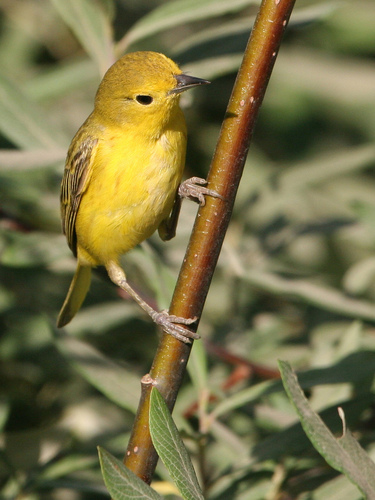}

\caption{Some example fine-grained images from Standford Dogs \cite{khosla2011novel} (Columns 1-3), FGVC-Aircraft \cite{maji2013fine} (Columns 4-6), and CUB-200 \cite{wah2011caltech} (Columns 7-9). Images in each column belong to the same category while images from different columns belong to different categories.}
\label{fig_fgvr_example}
\end{figure*}

\begin{figure}[ht!]
\centering
\includegraphics[width=0.49\textwidth]{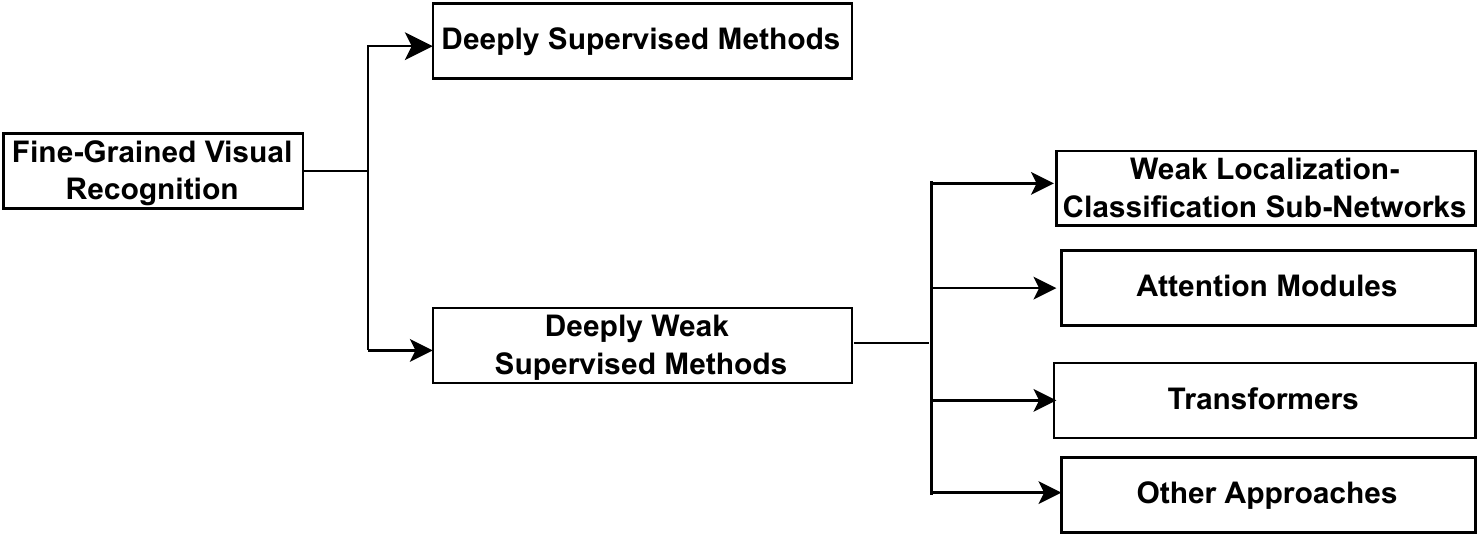}
\caption{The family of fine-grained visual recognition (FGVR) with local mechanisms.}
\label{fig_fgvr_family}
\end{figure}

Fine-grained visual recognition (FGVR) targets at recognizing the fined-grained class under a basic category, such as various species of animals \cite{wah2011caltech,khosla2011novel,van2018inaturalist,van2015building}, models of cars \cite{krause20133d} and aircrafts \cite{maji2013fine}, and so on. Some example images are shown in Fig. \ref{fig_fgvr_example}. We can observe that there are large intra-class variations and small inter-class variations. The FGVR has been widely used in different real-world applications, like intelligent retail systems.

Due to the inherently large intra-class variations and small inter-class differences in sub-classes, the FGVR task is a challenging problem. As pointed out in works to date, the global geometry and appearances of fine-grained classes tend to be similar, while the subtle differences on some key parts can be different. Therefore, it is important to extract local features for FGVR. Broadly, as shown in Fig. \ref{fig_fgvr_family}, existing approaches can be grouped into the two following categories: 1) Deeply supervised methods; 2) Deeply weak supervised methods. Compared to the latter which only requires image-level labels, the former needs additional information, like bounding boxes or part annotations. More detailed reviews are described in the following. 

\subsection{Deeply Supervised Methods}

Part-level bounding boxes or segmentation masks can help locate key image parts. Early work locates discriminative image parts with the assistance of manual annotations. It is generally presumed to have annotations in test time in previous work.

Some methods utilize object detection methods to localize useful parts. Without relying on object bounding box at test time, part-based R-CNNs (Part R-CNN) are proposed in \cite{zhang2014part} where CNNs are leveraged to learn whole-object and part detectors with bottom-up region proposals and enforce geometric constraints between parts. One challenge in FGVR is that discriminative parts may appear at various locations, and have rotation and scaling issues. Previous solutions perform localization, alignment, and classification independently, which results in sub-optimal performance. In Deep LAC\cite{lin2015deep}, localization, alignment, and classification are jointly optimized towards better results. In SPDA-CNN \cite{zhang2016spda}, a new model is proposed which incorporates semantic part localization and abstraction. The detection sub-network has a new top-down proposal method to generate small part proposals, based on which the recognition sub-network extracts part-based features and combines them for recognition. 

Some approaches use segmentation models to generate part masks. Compared with detection methods, segmentation approaches can generate more precise part localization. A part-based CNN (PS-CNN) is proposed in \cite{huang2016part}. In particular, a fully convolutional network is employed to localize object parts, followed by a two-stream network which extracts object-level and part-level features.

Part annotations are required at least in training, however, it is usually laborious and time-consuming to annotate object parts, limiting the application scenarios of these approaches.

\subsection{Deeply Weak Supervised Methods}
Recently, weakly-supervised approaches have received increasing attentions, in which fined-grained parts are located under the supervision of image-level labels only. This can alleviate the heavy burden of annotations. Exiting work can be grouped into several types: 1) Weak localization-classification sub-networks; 2) Attention modules; 3) Transformers.

\subsubsection{Weak Localization-Classification Sub-Networks}

Detection or segmentation techniques can be used to locate important image regions corresponding to fine-grained object parts(e.g., bird tails, dog ears).

In \cite{zhang2016weakly}, multi-scale part proposals are generated from object proposals, based on which discriminative part proposals are selected with part prototype clusters to generate discriminative part-level features. In TSC \cite{he2017weakly}, objects are first localized with saliency detection \cite{simonyan2014deep} and co-segmentation \cite{krause2015fine}. Then, discriminative parts are selected with two spatial constraints: box constraints and part constraints, which can enforce the selected parts inside the object region and ensure the localization of the most distinguished parts, respectively. When only guided by image-level labels, it is noticed that CNNs are likely to focus on the most discriminative parts, while neglecting other parts which can play an important role in the FGVR task. In \cite{ge2019weakly}, a weakly supervised complementary parts model is introduced which can capture complete object representations. In particular, object instances are first located by the object detection and instance segmentation via Mask R-CNN \cite{he2017mask} and CRF-based \cite{sutton2012introduction} segmentation. Then, the best parts are searched for each object to preserve as much diversity as possible. Lastly, a bi-directional long short-term memory is used to encode complementary information from different object parts. Most previous works localize discriminative regions independently, while neglecting contextual information and correlations between various parts. A graph-propagation based correlation learning (GCL) is proposed in \cite{wang2020graph} to fully explore and exploit discriminative potentials of correlations for FGVR.

\textbf{Pros and cons}. These approaches require a specially localization module to propose potential regions which are then forwarded through the classification network. However, this has two drawbacks: Firstly, the localization module has a different optimization goal compared with classification. Specifically, the localization network aims at locating parts which are shared across different object classes, encouraging representations to be similar. On the other hand, the classification network targets at extracting features which are different among different classes. The conflict makes it difficult to train these networks to learn representative features; Secondly, the overall time complexity is high. Besides, although some mechanisms are designed, some proposed parts tend to cover large areas of objects which are less discriminative to discern fined-grained classes. Moreover, these methods generate discrimiantive object parts with meaningful definitions, which may be hard to achieve, like flowers with repeating parts \cite{nilsback2008automated}.

\subsubsection{Attention Modules}

Attention modules can be used to automatically locate discriminative object parts.

In \cite{sermanet2014attention}, the attention mechanism is pioneered in the FGVR which is achieved by the RNN model. A bottom-up process is used in previous works to locate foreground object or object parts, learning discriminative features. However, it has high recall but very low precision about potential regions, which limits its effectiveness in some scenarios, like small objects with large background in an image. In \cite{xiao2015application}, three types of attention modules are proposed. Specially, the bottom-up attention proposes potential patches, the object-level top-down attention emphasizes relevant patches related to a object class, and the part-level top-down attention localizes discriminative parts. 

There are two drawbacks in current methods. First, human-defined parts or regions located by unsupervised approaches may be sub-optimal. Second, it is difficult to learn subtle differences to distinguish similar fine-grained classes. It is found that region localization and local feature learning are mutually correlated which can benefit each other. Therefore, a novel recurrent attention convolutional neural network (RA-CNN) is proposed in \cite{fu2017look} to solve these two limitations. An attention proposal sub-network starts from whole images, and iteratively generates part attentions from coarse to fine by using previous predictions as a reference under the guidance of an intra-scale classification loss and an inter-scale ranking loss. 

Despite decent performances, these approaches suffer from several limitations. First, additional steps, like the part localization and feature extraction of the attended regions, result in a high computational complexity. Second, sophisticated architecture designs make the training process complex. Third, correlations among different attention modules are neglected. Consequently, they tend to locate similar regions, missing some discriminative parts that are important in differentiating similar fine-grained classes. In \cite{sun2018multi}, multi-attention multi-class constraints (MAMC) are proposed to pull features from same-attention same-class closer, while pushing different attentions or classes away. Besides, inspired by SENet \cite{hu2018squeeze}, one-squeeze multi-excitation module is proposed to directly locate multiple parts with limited computational cost. 
In \cite{zheng2017learning}, a multi-attention convolutional neural network (MA-CNN) is proposed to jointly learn part proposals and fine-grained features where a channel grouping operation is conducted to generate multiple parts by clustering. In \cite{hu2019see}, a weakly supervised data augmentation network (WS-DAN) is proposed to locate discriminative parts only from the image level. Specifically, attention cropping randomly crops one attention part, emphasizing the local representations. Attention dropping randomly drops one attention region, pushing models to locate comprehensive image parts. 

In these works, the number of attentions is pre-defined, which is inflexible. In \cite{zheng2019looking}, a trilinear attention sample network (TASN) is proposed to extract discriminative representations from hundreds of part proposals and efficiently distill fine-grained features into a single CNN. Specifically, a trilinear module generates attention maps by learning inter-channel correlations. Then, an attention-based sampler emphasizes attended parts with high resolution, followed by a feature distiller to distill part representations into object-level representations. In \cite{ji2020attention}, an attention convolutional binary neural tree (ACNet) is presented, which incorporates convolutional operations along edges of the tree structure for the coarse-to-fine hierarchical feature learning. 

To deal with different hierarchy classifications, the granularity-wise attention is used in humans to emphasize different object regions. To leverage this mechanism, Cross Hierarchical Region Feature (CHRF) is proposed in \cite{liu2022focus}. In particular, a region feature mining module is introduced to learn multi-grained regions. Besides, a cross-hierarchical orthogonal fusion module is designed to combine original features with orthogonal elements from adjacent hierarchies, boosting part representations. 


\textbf{Pros and cons}. Even though attention modules achieve high performance on various datasets, they may be overfitted to the most discriminative parts, while ignoring other potentially important ones. It is expected to design proper mechanisms, guiding networks to explore comprehensive parts. This can make models robust to various challenging scenarios, like occlusions. Besides, these methods generally ignore relations among different image parts, causing performance degradation for images with a large portion of irrelevant background. Moreover, it is difficult to balance between information semantics and discriminative details in CNNs for the FGVR task. Extracting semantic information can well represent different fine-grained classes. This can be achieved by stacking convolution and pooling operations, however, the receptive field is expanded. Consequently, large image areas are emphasized, which degrade crucial fine-grained features for the FGVR.

\begin{table*}
\centering
\caption{Representative works on FGVR with Local features.}
\begin{tabular}{|c|c|c|c|c|}
\hline
\multicolumn{2}{|c|}{\multirow{1}{*}{Category}} & \multirow{1}{*}{Methods} & \multirow{1}{*}{Venue} & \multirow{1}{*}{Highlights}  \\
\hline
\multirow{4}{*}{\rotatebox[origin=c]{90}{\makecell{Deeply \\Supervised \\Methods}}}&\multirow{4}{*}[-0.5em]{\rotatebox[origin=c]{90}{-}}&\makecell{Part R-CNN \cite{zhang2014part}} & ECCV14 &\makecell{Bottom-up region proposals,\\ enforce geometric constraints between parts} \\
\cline{3-5}
&&\makecell{Deep LAC\cite{lin2015deep}} & CVPR15 &\makecell{Joint localization, alignment, and classification}\\
\cline{3-5}
&&\multirow{1}{*}{\makecell{SPDA-CNN \cite{zhang2016spda}}} & \multirow{1}{*}{CVPR16} & \multirow{1}{*}{Top-down proposals} \\
\cline{3-5}
&&\makecell{PS-CNN\cite{huang2016part}} & CVPR16 & \makecell{Fully convolutional networks, \\a two-stream network for object and part features}\\
\hline
\multirow{6}{*}[-10em]{\rotatebox[origin=c]{90}{\makecell{Deeply Weak Supervised Methods}}}&\multirow{4}{*}[-0.2em]{\rotatebox[origin=c]{90}{\makecell{Weak\\Localization-\\Classification\\ Sub-Networks}}}& \cite{zhang2016weakly}& TIP16 &\makecell{Multi-scale part proposals, part prototype clusters} \\
\cline{3-5}
&& \makecell{TSC \cite{he2017weakly}}& AAAI17 &\makecell{Saliency detection and co-segmentation, \\box and parts constraints} \\
\cline{3-5}
&&\cite{ge2019weakly}& CVPR19 &\makecell{Object detection and instance segmentation, \\bi-LSTM fuses part features} \\
\cline{3-5}
&&\makecell{GCL \cite{wang2020graph}}& AAAI20 &\makecell{Graph-propagation based correlation learning} \\
\cline{2-5}
&\multirow{5}{*}[-1em]{\rotatebox[origin=c]{90}{{\makecell{Attention Modules}}}}& \makecell{Two-level attention\cite{xiao2015application}} & CVPR15 &\makecell{Bottom-up attention, object-level and part-level top-down attentions}\\
\cline{3-5}
&& \makecell{RA-CNN \cite{fu2017look}} & CVPR17 &\makecell{Generate part attention from coarse to fine, inter-scale ranking loss} \\
\cline{3-5}
&& \multirow{1}{*}{\makecell{MAMC\cite{sun2018multi}}} & \multirow{1}{*}{ECCV18} &\makecell{Multi-attention multi-class constraint, one-squeeze multi-excitation} \\
\cline{3-5}
&& \makecell{MA-CNN \cite{zheng2017learning}} & CVPR17 &\makecell{Jointly learn part proposals and fine-grained features, channel grouping}\\
\cline{3-5}
&& \makecell{WS-DAN \cite{hu2019see}} & Arxiv19 &\makecell{Attention cropping, attention dropping}\\
\cline{3-5}
&& \makecell{TASN \cite{zheng2019looking}} & CVPR19 &\makecell{Trilinear attention module, attention sampler, knowledge distillation}\\
\cline{3-5}
&& \makecell{ACNet \cite{ji2020attention}} & CVPR20 &\makecell{Binary neural tree, coarse-to-fine hierarchical feature learning} \\
\cline{3-5}
&& \makecell{CHRF \cite{liu2022focus}} & ECCV22 &\makecell{Region feature mining, cross-hierarchical orthogonal fusion} \\
\cline{2-5}
&\multirow{5}{*}[-0.5em]{\rotatebox[origin=c]{90}{{Transformers}}}&  \makecell{PART\cite{zhao2021part}}& TIP21 &\makecell{Automatic part discovery, construct local and contextual correlations}\\
\cline{3-5}
&&  \makecell{TransFG \cite{he2021transfg}}& AAAI22 &\makecell{Part selection}\\
\cline{3-5}
&&  \makecell{FFVT \cite{wang2021feature}}& BMVC21 &\makecell{Hierarchical information, token selection} \\
\cline{3-5}
&&  \makecell{RAMS-Trans\cite{hu2021rams}}& MM21 &\makecell{Learn multi-scale discriminative parts}\\
\cline{3-5}
&&  \makecell{AFTrans \cite{zhang2022free}}& ICASSP22 &\makecell{Adaptively select important local regions}\\
\cline{3-5}
&& \makecell{DCAL \cite{zhu2022dual}}& CVPR22 &\makecell{Global-local cross-attention,  pairwise cross-attention } \\
\cline{2-5}
&\multirow{6}{*}[-0em]{\rotatebox[origin=c]{90}{{\makecell{Other \\Approaches}}}}&  \makecell{STN \cite{jaderberg2015spatial}} & NeurIPS15&\makecell{Spatial transformations} \\
\cline{3-5}
&&  \makecell{DFL-CNN\cite{wang2018learning}}& CVPR18 &\makecell{Bank of convolutional filters, asymmetric multi-stream CNN}\\
\cline{3-5}
&& \makecell{S3N \cite{ding2019selective}}& ICCV19 &\makecell{Exploit class peak responses, a dynamic number of regions} \\
\cline{3-5}
&& \cite{huang2020interpretable}& CVPR20 &\makecell{Interpretable model, object parts occurrence prior} \\
\cline{3-5}
&& \makecell{DCL \cite{chen2019destruction}}& CVPR19 &\makecell{Destruction and construction learning, adversarial learning} \\
\cline{3-5}
&& \makecell{PMG \cite{du2020fine}}& ECCV20 &\makecell{Jigsaw puzzle generator, multi-granularity, progressive learning} \\
\hline
\end{tabular}
\label{tab_fgvr}.
\end{table*}

\subsubsection{Transformers}

Transformers have a strong ability for capturing both global and local features, which have been widely explored recently.

In \cite{zhao2021part}, a joint CNN and Transformer model is proposed, called PArt-guided Relational Transformers (PART) which consist of an automatic part discovery module to extract discriminative part features. Nevertheless, automatically located regions are unstable, leading to overfitting on local regions. So, a Transformer model is adapted to construct local and contextual correlations. 

Different from the previous work which uses some Transformer layers, some recent work explores the potentials of ViT backbones. In TransFG \cite{he2021transfg}, it is the first time to apply the pure ViT into this field. Specifically, a part selection module is proposed to select tokens that correspond to important image patches as the input of Transformers. Two shortcomings exit in the TransFG model which limit further improvements in FGVR. First, the class token only focuses on the global receptive field and fails to captures multi-granularity features. Second, attention weights in different Transformer layers have varying importance. To address these two issues, a novel adaptive attention multi-scale fusion Transformer (AFTrans) is proposed in \cite{zhang2022free}. Specifically, the selection attention collection module utilizes attention weights in ViT to adaptively select important local regions, which is applied in a multi-scale manner.

However, since the final classification token is used for classification which pays more attention to global information and neglects the importance of local features, the performance is degraded. A novel pure Transformer-based framework, called feature fusion vision Transformer (FFVT), is proposed in \cite{wang2021feature} where important token from different Transformer layers are aggregated to take advantage of hierarchical information. Meanwhile, a novel token selection method is proposed to select representative tokens at each layer. 

A recurrent attention multi-scale Transformer (RAMS-Trans) is proposed in \cite{hu2021rams} where the self-attention weights in Transformers are used recursively to learn discriminative parts in a multi-scale way. 

To capture subtle visual differences between different sub-classes in ViTs, a dual cross-attention learning (DCAL) is proposed in \cite{zhu2022dual}. First, a global-local cross-attention models the interactions between global images and local parts, emphasizing the spatial-wise discriminative information. Second, a pairwise cross-attention builds relations between image pairs, which is a regularization to discover informative parts and alleviate the over-fitting issue.


\textbf{Pros and cons}. Generally, holistic structure information of the object should be emphasized, which is beneficial for localizing object foreground parts. The relation between patch tokens and the whole image structure is important for identifying discriminative parts, e.g., localizing gray legs of a bird among twigs. Besides, how to effectively combine local features in Transformers needs to be further explored.

\subsubsection{Other Approaches}

Except for the aforementioned approaches, some recent methods locate useful parts with different mechanisms, like spatial transformer networks, CNN outputs, an interpretation module, and destruction and construction. They can learn discriminative local features, achieving good results. 

The locations of discriminative parts can be automatically learned. Spatial transformer networks (STN) are explored in \cite{jaderberg2015spatial} to perform spatial transformations in an data-driven manner without any additional supervision. In particular, the STN is equipped with multiple transformers. Each spatial transformer corresponds to a part detector which can locate discriminative parts adaptively.

Intermediate CNN responses can be regarded as localized descriptors which could be linked to semantic parts \cite{zeiler2014visualizing}. A set of approaches explore CNN responses in this field. To address the conflict between localization and classification networks, middle level discriminative patches are explicitly generated in DFL-CNN \cite{wang2018learning}  by learning a bank of convolutional filters that extract discriminative patches. This is achieved via an asymmetric multi-stream network structure which uses both patch-level information and global appearance and a non-random layer initialization that activates filters on discriminative patches. 

Previous approaches crop parts and then learn fine-grained features, but are limited by the predefined number of parts and neglect surrounding context. The selective sparse sampling (S3N) is proposed in \cite{ding2019selective} which can extract diverse and fine-grained features. Informative regions are located by exploiting class peak responses. Besides, fine-detailed visual information is selectively aggregated with a dynamic number of regions conditioned on the image content and surrounding context. In \cite{huang2020interpretable}, an interpretable model is proposed. Specifically, a novel prior of the occurrence of object parts, which is combined with region-based part discovery and attributes, demonstrating a good performance. 

Many methods construct fine-grained specific tasks for learning unified discriminative features. In \cite{chen2019destruction}, a destruction and construction learning (DCL) framework is proposed where the destruction encourages networks to focus on discriminative parts for distinguishing different fine-grained classes and the construction learning models the semantic correlations among different parts to restore the spatial layout. This framework only has one-stage training process. Besides, it does not have additional cost in the inference stage. In progressive multi-granularity (PMG) \cite{du2020fine}, a jigsaw puzzle generator is introduced to guide networks to learn at different level of granularity and effectively combine features at these levels.

\subsection{Discussions}

A brief comparison between different models are shown in Table \ref{tab_fgvr}. Both deeply supervised methods and deeply weak supervised methods require supervision signals to learn discriminative features. However, labeling fine-grained datasets is challenging which requires expert knowledge. On the other hand, self-supervised learning can learn features from unlabled data. However, recent research \cite{cole2022does} suggests that  
contrastive learning tends to have the coarse-grained bias, being less effective to capture fine-grained details for the FGVR problem. Therefore, there remain questions related to the utility of self-supervised learning in the FGVR, which motivate future direction of the field. For example, how to design proper local mechanisms to guide the model to focus on fine-grained details for the FGVR problem?

\section{Person Re-Identification}\label{sec_reid}

\begin{figure}[ht!]
\centering
\includegraphics[width=0.36\textwidth]{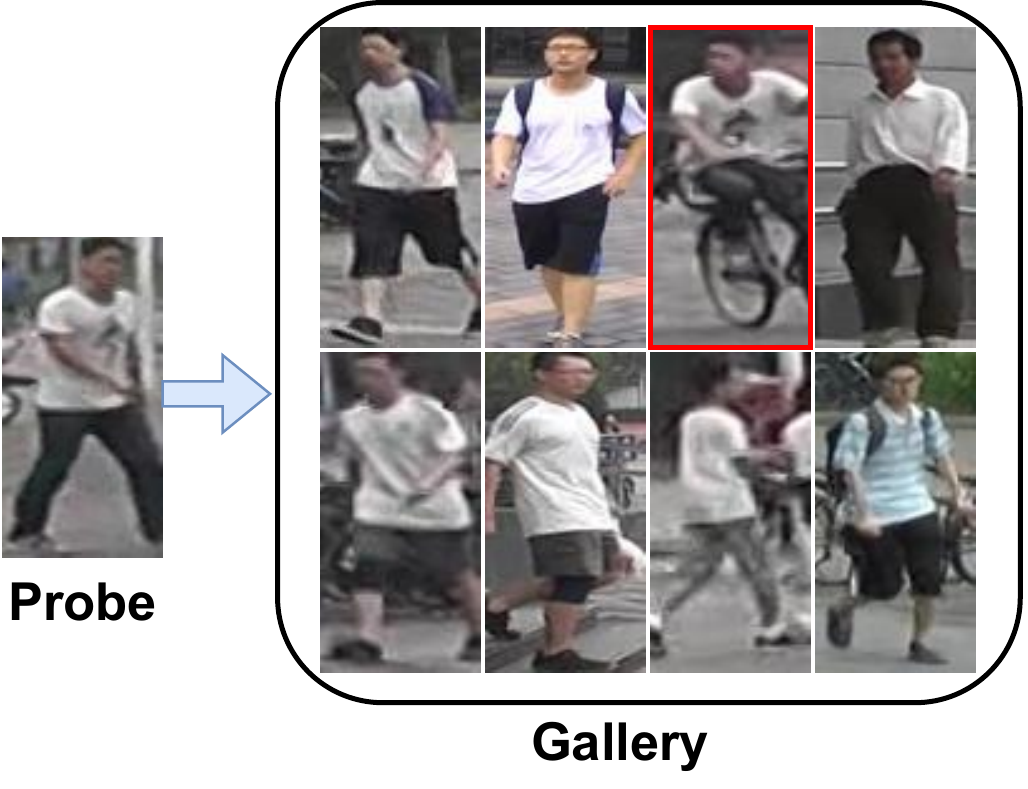}
\caption{The general pipeline of person re-identification (ReID).}
\label{fig_reid_pipeline}
\end{figure}

Generally, given a probe person, person re-identification (ReID) aims at re-identifying the same person across different cameras from different viewpoints in the gallery set, as shown in Fig. \ref{fig_reid_pipeline}. The ReID is attracting attentions from both academia and industry because of its important role in various surveillance applications, such as the criminal retrieval. 

\begin{figure}[ht!]
\centering
\includegraphics[width=0.41\textwidth]{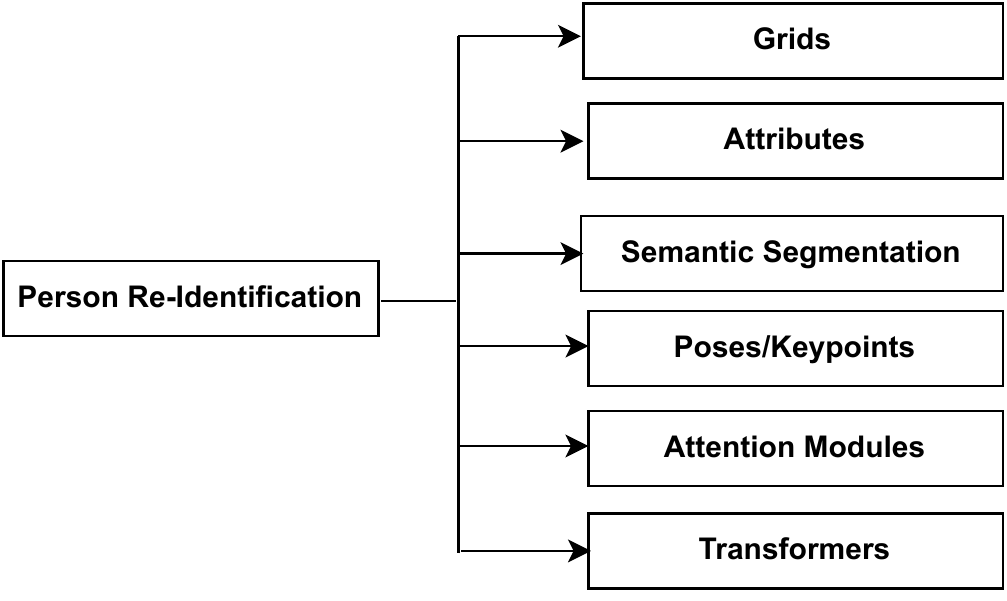}
\caption{The family of ReID with local mechanisms.}
\label{fig_reid_family}
\end{figure}

ReID is a very challenging task due to various factors, such as human body pose variations, camera viewpoints, illumination changes, and cluttered background. Early approaches directly learn information from the whole image. However, background clutters are also extracted, deteriorating the representation. The global misalignment problem also makes the ReID problem challenging. Recently, as demonstrated in Fig. \ref{fig_reid_family}, various approaches are proposed to learn identity features from body parts, which could be roughly summarized into the following categories: 1) Grids. 2) Attributes. 3) Semantic segmentation. 4) Poses/keypoints. 5) Attention modules. 6) Transformers. Methods in each category have their own advantages and drawbacks, which are analyzed subsequently.

\subsection{Grids} 
For grid-based approaches, they split the input image or feature maps into small patches or stripes, and then extract local features from patches or stripes. 

Without relying on additional guidances like human pose estimation, the content consistency within each part is taken into consideration in \cite{sun2018beyond}. Specifically, a part-based convolutional baseline (PCB) is proposed where a uniform partition strategy is used for learning part-level features which are further assembled to convolutional features. Besides, a refined part pooling (RPP) is introduced to reassign the outlier in each part to the parts they are closet to, achieving the within-part consistency. In \cite{wang2018learning_reid}, a multiple granularity network (MGN) is proposed to learn both global and local feature representations in an end-to-end way where the images are uniformly partitioned into several stripes and the number of stripes in different local branches varies, capturing local features with different granularities.

Unsupervised ReID targets at learning discriminative representations from unlabeled data via pseudo-labels. However, the generated labels are noisy and deteriorate the performance. Several methods are proposed to tackle this issue, while neglecting the fine-grained local context. Therefore, a novel Part-based Pseudo Label Refinement (PPLR) is proposed in \cite{cho2022part} which can explore the complementary relationship between global and local representations to reduce the label noise. Particularly, feature maps are uniformly partitioned into different parts. Then, a cross agreement score is introduced to leverage the complementary information between global and part features, based on which agreement-aware label smoothing and part-guided label refinement are proposed to further refine noisy labels.

\textbf{Pros and cons}. For grid based methods, they can extract comprehensive features on rigid body regions. However, these methods implicitly assume that every person appear at a similar pose, which is hard to meet in some real-world scenarios. It is inevitable that these methods would suffer from performance drop under body misalignments caused by occlusion and pose variations. Many complex alignment techniques are designed to address this issue. However, these approaches may involve extra computations and careful tuning. Besides, parts with same semantics may be distributed into several different splits, deteriorating the representations.


\subsection{Attributes}

Most existing work for video-based ReID fails to consider different frame weights caused by viewpoint or pose variations. In \cite{zhao2019attribute}, an attribute-driven approach is introduced to disentangle features into groups of sub-features based on semantic attributes, followed by a sub-feature re-weighting module which maximizes the complementarity between various discriminative regions from different frames using the confidence of attribute recognition. Meanwhile, a transfer learning method is proposed to automatically annotate attribute labels, alleviating the labor-expensive process of manually annotating attributes.

Attributes can narrow the appearance gap across different modalities, but have not been explored in cross-modality ReID. A novel Progressive Attribute Embedding Network (PAENet) is proposed in \cite{zheng2022progressive} to address the above issue. A progressive attribute embedding module combines fine-grained semantic attribute information with global structure information. Besides, an attribute-based auxiliary learning strategy guides networks to extract modality-invariant and identity-specific local representations with attribute and identity losses.

\textbf{Pros and cons}. For attributes-based methods, they can locate semantic areas under the guidance of semantic signals. However, they only focus on limited parts with fixed semantics, while ignoring some important parts. This degrades the performance if some semantics are unavailable in target objects caused by occlusions. Besides, attributes are expensive to collect.

\subsection{Semantic Segmentation}

Several approaches apply semantic segmentation in the ReID task where semantic areas are first segmented and cropped, then features are extracted from these areas. 

There is no correspondence of local regions for images across different poses. To address this issue, a pose-sensitive embedding model is proposed in \cite{sarfraz2018pose} where both the person's coarse pose (i.e., captured views via the camera) and the fine body pose (i.e., joint locations) are incorporated to learn discriminative representations. 

The large pose variations and complex view changes make it difficult to learn and match features from different person images. To address these issues, a pose-driven deep convolutional (PDC) model is proposed in \cite{su2017pose} to adopt human part cues to alleviate the pose variations and learn robust features from both the global image and multiple local parts. Meanwhile, a pose driven weighting sub-network is proposed to learn adaptive weights for global features and local parts. 

The detection of bounding boxes for extracting local patches is difficult for low resolution images. Meanwhile, bounding boxes are coarse which may contain background. Since semantic segmentation can precisely locate body parts even under large pose changes, a human semantic parsing model (SPReID) is used in \cite{kalayeh2018human} to effectively exploit local cues for the ReID task. 

A dual part-aligned representation module (P\textsuperscript{2}-Net) is proposed in \cite{guo2019beyond} in which both human and non-human parts are exploited to capture missed contextual cues. Existing work mainly captures discriminative information about predefined human parts or attributes, while neglecting some objects or parts which are critical in the ReID. Specifically, a human parsing model is applied to learn binary human part masks for human part alignments and a self-attention module is used to capture non-human part masks for non-human part alignments. However, the off-the-shelf models may output inaccurate or even incorrect semantic estimation, limiting the robustness of these models. An identity-guided human semantic parsing approach (ISP) is proposed in \cite{zhu2020identity} to locate both human body parts and personal belongings at pixel-level. More precisely, pixels of all images are first grouped into background or foreground. Then, foreground pixels are further grouped into human parts. Next, cluster assignments are utilized as pseudo-labels of human parts to learn part features.

It should be noticed that semantic segmentation methods usually have a high running time. Besides, these methods can only segment objects with predefined categories, which significantly limits the generalization ability on unseen classes.

\textbf{Pros and cons}. For semantic segmentation based approaches, they learn part-aligned features at specific locations, which can ease the misalignment issue. However, they have several limitations: Firstly, most of them need to fine-tune on a specific dataset, adapting to new scenarios. Since they are not trained in an end-to-end manner, they have a high time complexity and training difficulty of two-stage feature learning; Secondly, generated image parts ignore contextual information, which tend to be overfitted to specific visual patterns; Thirdly, relations among different keypoints are not fully explored, making it challenging to recognize some matches; Fourthly, these approaches only focus on specific semantic parts, but fail to represent comprehensive discriminative parts. For example, the identifiable personal belongings (e.g., bag) cannot be recognized by existing pre-trained models.

\subsection{Poses/Keypoints}

Some methods first conduct pose estimations or detect body keypoints, then crop body parts. However, current pose estimation methods are far from perfect, leading to imprecise localization of body parts. It can be expected that body parts misalignment can occur between bounding boxes. Meanwhile, the bounding boxes cover objects instead of fined-grained parts. A network is proposed in \cite{suh2018part} to learn part-aligned features. Specifically, a two-stream network is proposed to learn appearance and body part feature maps separately. Then, a bilinear pooling layer calculates the bilinear mapping of the appearance and part features at each location, followed by a spatial average operation to generate part-aligned features.

\textbf{Pros and cons}. For pose/keypoint based approaches, they can similar semantics with location information, however, they generate unreliable localization for images with pose or occlusion variations, leading to the misalignment issue.

\begin{table*}
\centering
\caption{Representative ReID models with local mechanisms.}
\begin{tabular}{|c|c|c|c|c||c|c|c|c|c|}
\hline
\multicolumn{1}{|c|}{\multirow{1}{*}{Category}} & \multirow{1}{*}{Methods} & \multirow{1}{*}{Venue} & \multirow{1}{*}{Highlights} \\
\hline
\multirow{4}{*}{\rotatebox[origin=c]{0}{\makecell{Grids}}}& \makecell{\cite{sun2018beyond}}& ECCV18 &\makecell{A part-based convolutional baseline, a refined part pooling}\\
\cline{2-4}
& \makecell{MGN \cite{wang2018learning_reid}}& MM18 &\makecell{Local features with different granularities}\\
\cline{2-4}
& \makecell{ PPLR \cite{cho2022part}}& CVPR22 &\makecell{Cross agreement score, agreement-aware label smoothing, part-guided label refinement}\\
\hline
\multirow{3}{*}{\rotatebox[origin=c]{0}{\makecell{Attributes}}}& \makecell{\cite{zhao2019attribute}}& CVPR19 &\makecell{Attribute-driven, feature disentanglement with attributes,\\ attribute-feature re-weighting, transfer learning}\\
\cline{2-4}
& \makecell{PAENet \cite{zheng2022progressive}}& MM22 &\makecell{Progressive attribute embedding, attribute-based auxiliary learning}\\
\hline
\multirow{6}{*}{\rotatebox[origin=c]{0}{\makecell{Semantic\\Segmentation}}}& \makecell{\cite{sarfraz2018pose}}& CVPR18 &\makecell{Both coarse pose and fine body pose}\\
\cline{2-4}
& \makecell{ PDC\cite{su2017pose}}& ICCV17 &\makecell{Human part cues, a pose driven weighting sub-network}\\
\cline{2-4}
& \makecell{ SPReID\cite{kalayeh2018human}}& CVPR18 &\makecell{Human semantic parsing}\\
\cline{2-4}
& \makecell{ P\textsuperscript{2}-Net \cite{guo2019beyond}}& ICCV19 &\makecell{Human parsing aligns human parts, self-attention aligns non-human parts}\\
\cline{2-4}
& \makecell{ ISP \cite{zhu2020identity} }& ECCV20 &\makecell{Identity-guided human semantic parsing}\\
\hline
\multirow{1}{*}{\rotatebox[origin=c]{0}{\makecell{Poses/Keypoints}}}& \makecell{\cite{suh2018part}}& ECCV18 &\makecell{Part-aligned features, two-stream network, spatial average}\\
\hline
\multirow{10}{*}{\rotatebox[origin=c]{0}{\makecell{Attention\\Modules}}}& \makecell{CAN\cite{liu2017end}}& TIP17 &\makecell{Triplet recurrent neural network}\\
\cline{2-4}
& \makecell{MGCAM \cite{song2018mask}}&CVPR18  &\makecell{A binary body mask, a region-level triplet loss}\\
\cline{2-4}
& \makecell{HA-CNN \cite{li2018harmonious}}& CVPR18 &\makecell{Multi-granularity hard region-level and soft pixel-level attentions, a cross-attention interaction}\\
\cline{2-4}
& \makecell{ABD-Net \cite{chen2019abd}}& ICCV19  &\makecell{Channel and spatial attention, diversity constraints}\\
\cline{2-4}
& \makecell{AACN \cite{xu2018attention}}& CVPR18 &\makecell{A pose-guided part attention, an attention-aware feature composition}\\
\cline{2-4}
& \makecell{MHN \cite{chen2019mixed}}& ICCV19 &\makecell{Applying multiple high-order attentions at different orders}\\
\cline{2-4}
& \makecell{RGA \cite{zhang2020relation}}& CVPR20 &\makecell{Global scope relations}\\
\cline{2-4}
& \makecell{PartMix \cite{kim2023partmix}}& CVPR23 &\makecell{Part mixture across different modalities, contrastive learning, entropy-based part pair mining}\\
\hline
\multirow{7}{*}{\rotatebox[origin=c]{0}{\makecell{Transformers}}}& \makecell{PAT \cite{li2021diverse}}& CVPR21 &\makecell{Pixel context based Transformer encoder, part prototype based Transformer}\\
\cline{2-4}
& \makecell{TransReID \cite{he2021transreid}}& ICCV21 &\makecell{Side information embedding, a jigsaw patch module}\\
\cline{2-4}
& \makecell{COAT \cite{yu2022cascade}}&CVPR22  &\makecell{A cascaded multi-scale Transformer, an occluded attention}\\
\cline{2-4}
& \makecell{DRL-Net \cite{jia2022learning}}& TMM22 &\makecell{Feature disentanglement, a decorrelation constraint}\\
\cline{2-4}
& \makecell{DFLN-ViT \cite{zhao2022spatial}}& MM21 &\makecell{Long-range relations on locations and channels}\\
\cline{2-4}
& \makecell{Pirt \cite{ma2021pose}}& TIP22 &\makecell{Local relations, cross relationships}\\
\cline{2-4}
& \makecell{SPOT \cite{chen2022structure}}& ICLR21 &\makecell{Structure-related appearance features, contextual and positional interactions}\\
\hline
\end{tabular}
\label{tab_reid} 
\end{table*}

\subsection{Attention Modules}

Recently, attention modules are employed to capture discriminative human part features automatically. This has limited additional computational cost without relying on knowledge from other sources (e.g., attributes, poses, keypoints, semantic segmentation).

Local discriminative features are learned once for all in previous methods. Consequently, these approaches suffer from performance drops under factors, such as occlusions or illumination variations, because some parts may become invisible. To alleviate this issue, an end-to-end comparative attention network (CAN) is proposed in \cite{liu2017end} which can adaptively compare discriminative part pairs of person images. It automatically localizes multiple local regions with a triplet recurrent neural network which makes positive pairs closer and negative pairs far away from each other.

Motivated by the fact that removing the background regions in the person image is beneficial for ReID, a mask-guided contrastive attention model (MGCAM) is presented in \cite{song2018mask} in which a binary body mask generates body-aware and background-aware attention maps. Besides, a region-level triplet loss is proposed to pull features from the full image and body region close, and push features from background away. 

Previous highly complex models are less effective when facing a small-scale labelled dataset for training. Besides, misalignments and background clutters also degrade the performance. To address these two issues, a harmonious attention convolutional neural network (HA-CNN) is proposed in \cite{li2018harmonious} to simultaneously learn multi-granularity hard region-level and soft pixel-level attentions for some unknown misalignment each bounding box. Besides, a cross-attention interaction boosts the compatibility between the local and global features.

Many attention-based methods apply spatial clues to extract part-based representations, which mainly focus on foregrounds. In \cite{chen2019abd}, an attentive but diverse network (ABD-Net) is proposed to alleviate issues of misalignment and background clutter, and reduce the correlation between features. More precisely, both channel and spatial attention modules are utilized to encourage networks to focus on discriminative local parts. Besides, the additional diversity constraint reduces highly correlated and redundant features. 

It is argued that pose information is not fully explored for person ReID. A novel framework called attention-aware compositional network (AACN) is proposed in \cite{xu2018attention} to cope with misalignment and occlusion issues which consists of a pose-guided part attention and an attention-aware feature composition. The former filters out background noises and accurately locates desired image parts. The latter predicts visibility scores to deal with the part occlusion problem.

It is observed in \cite{chen2019mixed} that the widely used spatial and channel attentions are either coarse or first-order, which have a limited representational ability to model complex interactions between image parts. A mixed high-order attention network (MHN) is proposed to alleviate this issue. This is implemented by applying multiple high-order attentions at different orders to learn diverse high-order statistics. 

It is claimed that existing methods fail to explore information from global structure patterns which are helpful for inferring semantic attention. A relation-aware global attention (RGA) is proposed in \cite{zhang2020relation} to leverage the appearance feature and its global scope relations to predict the feature importance. Specifically, for a feature pattern, its relations with all feature patterns are stacked as a vector which represents global structure features, mining discriminative patterns.

Mixture-based data augmentation methods for visible-infrared ReID remains unexplored. To this end, PartMix \cite{kim2023partmix} mixes part descriptors across different modalities to sythesize augmented samples on part maps. In particular, positive samples and negative samples within the same and across different subjects are synthesized, followed by constrative learning which regularizes the network. Besides, a entropy-based mining strategy is used to select reliable positive and negative samples for constrative learning.

\textbf{Pros and cons}. For attention-based approaches, they can automatically locate useful object parts automatically. However, learning such attention modules could be overfitted to specific parts because the modules tend to focus on the most discriminative parts to recognize the identity, while ignoring other potentially important parts, especially under some challenging scenarios.

\subsection{Transformers}

Due to the excellent capability of Transformers in learning local features and building relations among different parts, they have also been widely explored in this field. 

There are two issues for occluded ReID observed in \cite{li2021diverse}: background has diverse characteristics, making it difficult to extract robust features for the query person; It is difficult to automatically locate non-occluded parts only using the identity labels. To cope with the aforementioned issues, a part-aware Transformer (PAT) is proposed to discover diverse parts via a pixel context based Transformer encoder and a part prototype based Transformer decoder. In CNN-based networks, different special designs are proposed to include some useful clues such as cameras and viewpoints. 

Unlike the aforementioned work, the popular ViT models are greatly explored in this field. A unified framework (TransReID) is proposed in \cite{he2021transreid} where a pure Transformer framework is applied for the first time for the ReID task. In details, the side information embedding is utilized to encode different types of side information for the ReID, like cameras or viewpoints. Meanwhile, a jigsaw patch module is designed to learn robust representations by shuffling the patches. 

A general person search system usually simultaneously solve the person detection and ReID problem. However, there are three issues to address: 1) Different learning targets for person detection and ReID problem; 2) Scale and pose variations; 3) Occlusions and background clutters. A cascade occluded attention Transformer (COAT) is proposed in \cite{yu2022cascade} where a cascaded multi-scale Transformer helps to address scale and pose/viewpoint changes progressively and an occluded attention module learns embeddings on discriminative parts of each person at each scale, alleviating the occlusion issue. 

Many approaches are proposed to address the misalignment issue with extra complicated modules. In \cite{jia2022learning}, a disentangled representation learning network (DRL-Net) is proposed for occluded ReID without the requirement of alignment. It performs disentanglement on representations of undefined semantic components in occluded images with semantic preference object queries in the Transformer architecture. Besides, a decorrelation constraint is designed in the Transformer decoder which is applied on queries to learn different semantic information.

Many existing methods ignore the importance of correlations between varying channels and locations. A discriminative feature learning network based on a visual Transformer (DFLN-ViT) is proposed in \cite{zhao2022spatial} for visible-infrared ReID where the Transformer is used to build long-range relations among different locations and channels, respectively. 

For occluded ReID, semantic information of different parts changes and relations between these parts are neglected. To solve these problems, a pose-guided intra- and inter-part relational Transformer (Pirt) is proposed in \cite{ma2021pose} where the former builds local relations with mask-guided features and the latter learns cross relationships between different parts. 

To boost the robustness against background distraction and misalignments for shared modality learning, a structure-aware positional Transformer (SPOT) is proposed in \cite{chen2022structure} to extract structure-related appearance features, addressing the complex background noise. Besides, it models contextual and positional interactions, improving robustness against pose and occlusion variations.

\textbf{Pros and cons}. For Transformer models, they can learn relations among different body parts. Although comprehensive image patches are extracted on original images, hard-split image patches also have the misalignment issue. Besides, semantically similar parts may be distributed into different splits. It is necessary to learn semantic-aware local representations.

\subsection{Discussions}
The reviewed representative ReID models with local mechanisms are summarized in Table \ref{tab_reid}. 

Most existing Re-ID methods use a centralized learning paradigm, which collects all training data from different scenarios. However, they fail to consider that personal and private data may not be allowed to be collected into a centralized setting, limiting the application of centralized learning in the real world. Federated learning can train machine learning models on private data distribute over multiple devices \cite{mcmahan2017communication}. In the Re-ID task, how to propose a novel federated learning paradigm to extract local representations on each device and learn a global model across different devices remains further exploration.

\section{Few-Shot Learning}\label{sec_fsl}

\begin{figure}[ht!]
\centering
\includegraphics[width=0.46\textwidth]{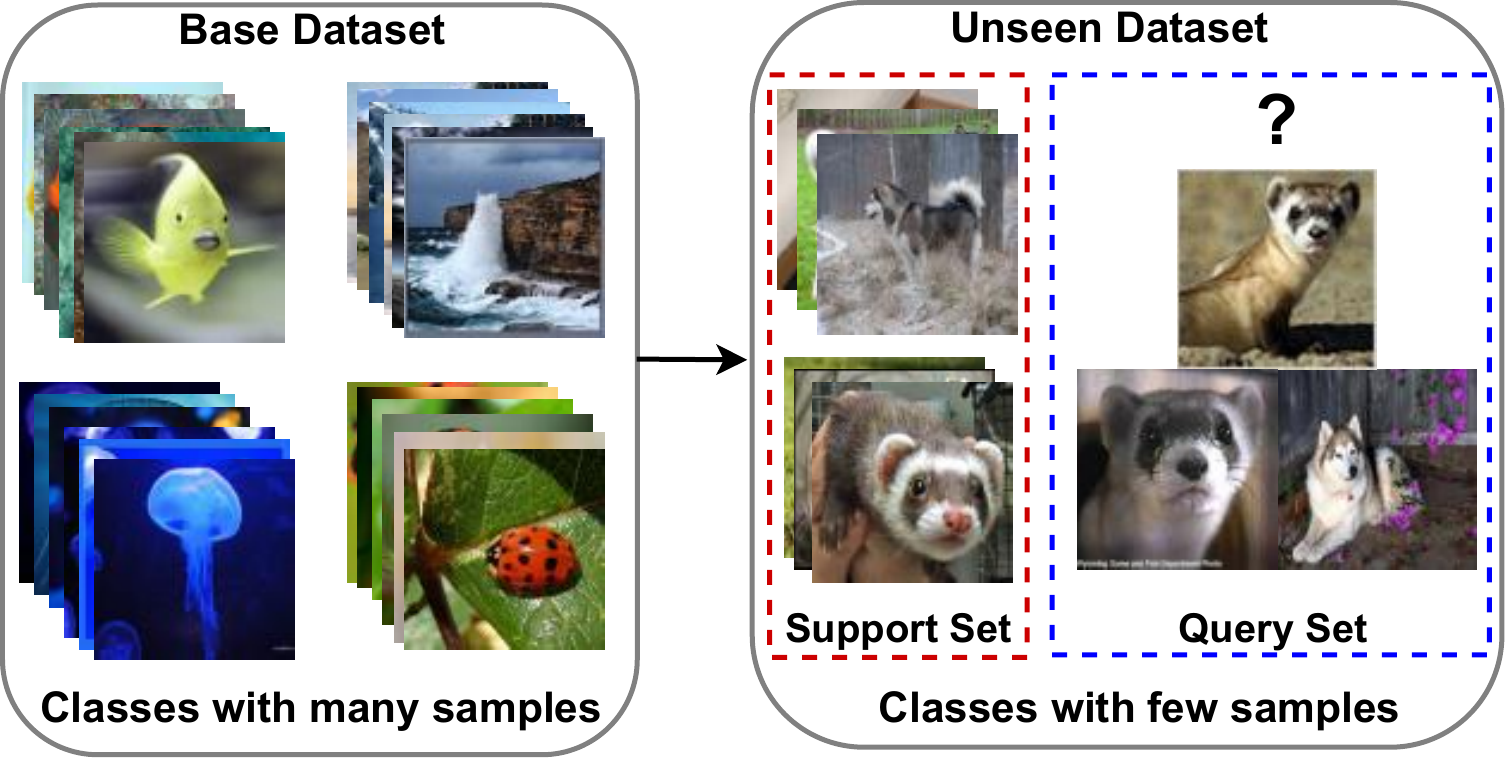}
\caption{The general pipeline of few-shot learning (FSL).}
\label{fig_fsl_pipeline}
\end{figure}

Rapid developments have been obtained in deep learning, with a particular example being image classification on large-scale image datasets, like ImageNet \cite{russakovsky2015imagenet}. Although their success on various benchmark datasets, a large number of annotated training images are typically required, which severely limits the scalability to new classes due to the labor-expensive annotation cost. Differently, humans are able to recognize objects in unseen classes with few examples. For example, children can easily recognize real-world marten by only taking a glance it from a picture. Motivated by this observation, few-shot learning (FSL) is proposed to classify unseen samples into a set of new visual categories (query set), given only few labelled examples in each category (support set) \cite{fei2006one,lake2015human}, as shown in Fig. \ref{fig_fsl_pipeline}. The FSL has many important applications in practice, such as recognizing a few labeled samples due to data scarcity (e.g., rare species), or high labeling cost (e.g., pixel-level labeling).

Typically, a support set only has 1-5 labeled examples per unseen class in the FSL, while each class usually has thousands of images in the general classification, such as ImageNet. Another difference between the FSL and general classification is that the former has disjoint classes in training and testing sets and the latter has the same classes.

\begin{figure}[ht!]
\centering
\includegraphics[width=0.46\textwidth]{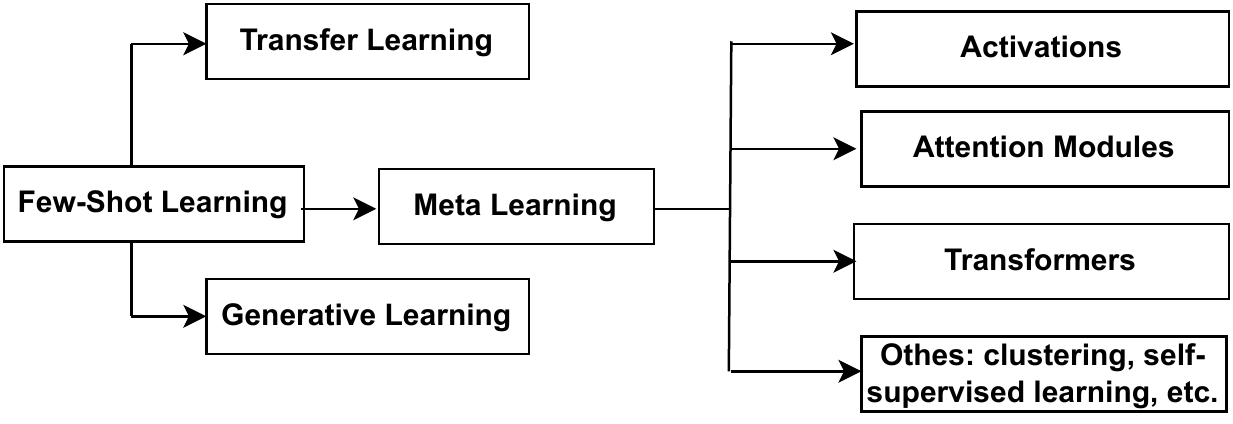}
\caption{The family of FSL with local mechanisms.}
\label{fig_fsl_family}
\end{figure}

To solve the FSL problem, methods can be mainly split into transfer learning, generation-based, and meta learning, as show in Fig. \ref{fig_fsl_family}. Transfer learning based approaches transfer the useful knowledge in the base dataset to the target dataset. Generation-based methods generate similar support images in few-shot tasks, which greatly address the issue of few images in novel classes. These methods are out of the scope of this survey. 

\begin{figure}[ht!]
\centering
\includegraphics[width=0.46\textwidth]{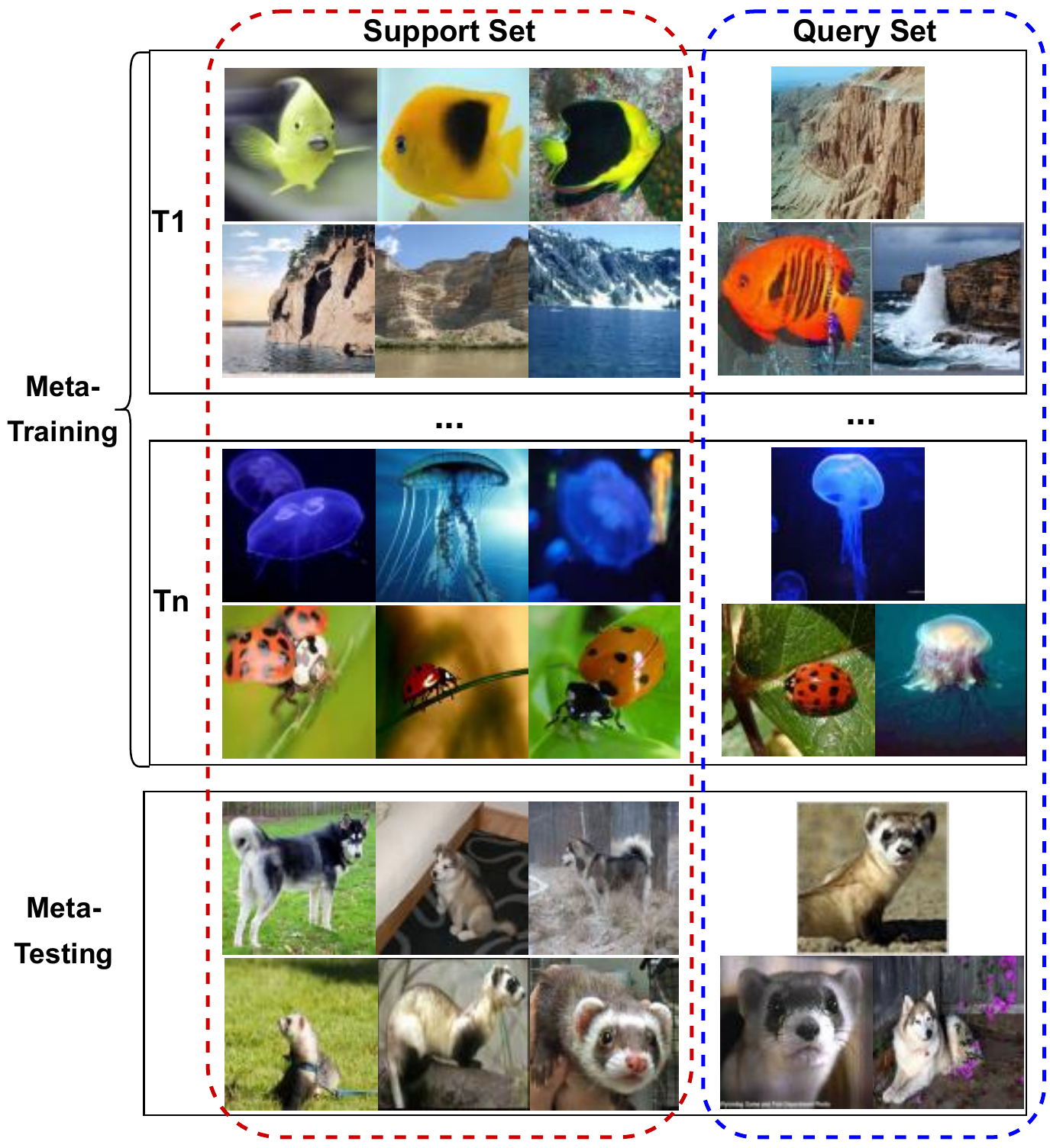}
\caption{The general pipeline of the meta-based FSL where T\textsubscript{n} represents the n\textsuperscript{th} task.}
\label{fig_fsl_meta_pipeline}
\end{figure}

Meta learning approaches learn a good parameter initialization for adaptation to new tasks. These learned weights can be quickly adapted to unseen classes with few labeled images using gradient-based optimization. They usually consist of an embedding function, a metric framework, and 
an optimization algorithm. Given an unlabeled query image and few labeled samples, the embedding function generates features for all images. Then, the metric framework calculates distances between query features and sample features using a distance metric to predict the classification. The general pipeline of the FSL is shown in Fig. \ref{fig_fsl_meta_pipeline}. In the N-way K-shot setting, during meta-training, each task is created by randomly selecting N classes where each class contain K images. The support set is used as training and the query set is the testing set. Repeat this process, it can learn find good parameter initializations. Then, in the meta-testing, the new task with unseen classes is presented to fine-tune the model, learning to recognize unseen classes. 

Most existing meta-based works depend on image-level pooled representations or fully connected layers for classification, while neglecting discriminative local features to transfer between seen and unseen classes \cite{snell2017prototypical,sung2018learning,gidaris2018dynamic,ye2020few}. It is somewhat idealistic to transfer image-level representations from seen classes to unseen categories. On the other hand, local features are more consistent among the seen and unseen categories. As shown in Fig. \ref{fig_fsl_family}, current meta learning approaches can be divided into following categories: 1) Activations: They are outputs of feature maps in CNNs. Since each activation corresponds to an image part, it contains local features. Therefore, some approaches are designed based on activations to transfer between seen and unseen classes; 2) Attention modules: With task information, attention mechanisms are explored to learn task-relevant information to generalize different tasks; 3) Transformers: Due to their strong capability in building relations among different parts, Transformers have been widely used in the FSL field; 4) Others: Other mechanisms, including clustering and self-supervised learning, are also explored to benefit the FSL. 

\subsection{Activations}

Recently, some works take advantage of local representations from activations of feature maps to boost the performance of FSL. 

In \cite{lifchitz2019dense}, local activations are the first time to be investigated in FSL where the dense classification is applied to each spatial location independently over feature maps, extracting additional knowledge from limited training data. Besides, neural implants are introduced to attach new neurons to learn new task-specific representations. A deep nearest neighbor neural network (DN4) \cite{li2019revisiting} learns deep local descriptors from feature maps, based on which an image-to-class similarity is conducted through a k-nearest neighbor selection between a given query image and each of the class in the support set, boosting the performance.

It can lead to the ambiguity problem if directly computing distances of dominate objects between the query image and support images because objects can appear at different locations. Semantic alignment metric learning (SAML) \cite{hao2019collect} is proposed to overcome this issue by aligning semantically relevant objects in a “collect-and-select” way. To be specific, a relation matrix is generated to collect the distances of each part pairs in feature maps between a query image and the mean tensor of support images. Then, an attention module is used to pay more attention 
to semantically relevant pairs, refining the relation matrix. Finally, a multi-layer perceptron maps the refined matrix to corresponding similarity scores.

Although prior works achieve promising performance, image-level representations in the same classes tend to have large differences because of background clutters and large intra-class appearance changes. On the other hand, local representations can provide transferable features across different categories. To extract local features and minimize the effects of distracted regions, DeepEMD \cite{zhang2020deepemd} takes advantages of Earth Mover’s Distance (EMD) which computes a structural distance on activations of feature maps to generate image relevances. Besides, to automatically determine the importance score for different image regions, a cross-reference module is designed.

Previous prototypical representations cannot model spatial relations between image parts. Besides, they are affected by orientations. A new network is proposed in \cite{wu2020attentive}. A capsule network models relative spatial relationships between features, which is guided by a new triplet loss to learn discriminative embeddings. Besides, attentive prototypes aggregate samples in the support set based on the importance score, which are defined by the reconstruction error for a query image.

It is important to understand the current task from a global perspective and locate different parts of the sample in the support set for different queries. Cross non-local networks (CNL) \cite{zhao2021looking} can capture task-specific and context-aware features dynamically by strengthening features at a specific position from all positions and the current task. Besides, mutual information between original and refined representations is maximized to reduce the issue of losing discriminative features. Moreover, a task-specific scaling is utilized to scale multi-scale and task-specific features in CNL. 

Most existing methods only focus on image-level or spatial-level feature alignments, while ignoring the channel discrepancy which is larger than information discrepancy among spatial locations. A dynamic alignment is proposed in \cite{zhu2021few} to effectively emphasize both query regions and channels based on different local support information. To be specific, dynamic meta-filter is utilized to align query features with position-specific and channel-specific knowledge. Moreover, a neural ordinary differential equation allows an accurate alignment control.

Prior methods characterize appearances at a grid of spatial locations. However, they fail to incorporate fine spatial details and tend to be over-fitted to pose or position issues. To address these issues, feature map reconstruction networks (FRN) \cite{wertheimer2021few} reconstruct a query feature map from the support images of a given class.

Recent methods meta-learn an embedding for each image, based on which the distance between images is calculated to conform the semantic space. However, they tend to be overfitted on irrelevant features, limiting their transferability to unseen classes. To address this issue, relational embedding networks (RENet) are proposed in \cite{kang2021relational}, which consist of self-correlational representations and cross-correlational attentions. The former extracts semantic structural patterns by calculating each activation of the feature map to its neighborhood. The latter generates semantic correspondence relations between two images by computing cross-correlation between two images. 

Recent methods usually adopt a unidirectional pipeline where the nearest support features are found for each query feature and matches for different local features are aggregated. A novel mutual centralized learning (MCL) is proposed in \cite{liu2022learning} to learn mutual affiliations between the query and support dense features in a bidirectional paradigm. 

Previous methods usually use pooled global features to classify low-data. Some recent works usually adopt deep descriptors, while failing to consider that some are useless because of the limited receptive field. Mutual Nearest Neighbor Neural Network (DMN4) is proposed in \cite{liu2022dmn4} to explicitly select query descriptors that are related with each task and discard less useful ones. This is achieved by the first attempt to combine mutual nearest neighbor with naive-bayes nearest neighbor.

\textbf{Pros and cons}. These methods use activations from feature maps in CNNs as local features, based on which different mechanisms are designed to transfer details from seen to unseen classes and calculate the distance between the query image and support images. However, activations are dense which contain information with different discriminative abilities. Some mechanisms are expected to remove useless activations and enhance discriminative ones. Besides, some activations may contain a part of semantic regions, leading to the misalignment issue. Moreover, current approaches fail to consider the relations among different activations, missing important contextual features.

\begin{table*}[ht!]
\centering
\caption{Representative FSL models with local mechanisms.}
\begin{tabular}{|c|c|c|c|c||c|c|c|c|c|}
\hline
\multicolumn{1}{|c|}{\multirow{1}{*}{Category}} & \multirow{1}{*}{Methods} & \multirow{1}{*}{Venue} & \multirow{1}{*}{Highlights} \\
\hline
\multirow{13}{*}{\rotatebox[origin=c]{90}{\makecell{Activations}}}& \makecell{ \cite{lifchitz2019dense}}&CVPR19  &\makecell{Dense classification on feature maps, neural implants}\\
\cline{2-4}
& \makecell{DN4 \cite{li2019revisiting}}&CVPR19 &\makecell{Deep local descriptors, k-nearest neighbor selection}\\
\cline{2-4}
& \makecell{SAML \cite{hao2019collect}}&ICCV19 &\makecell{Semantically relevant part alignment}\\
\cline{2-4}
& \makecell{DeepEMD \cite{zhang2020deepemd}}&CVPR20 &\makecell{Earth mover’s distance, cross-reference module}\\
\cline{2-4}
& \makecell{\cite{wu2020attentive}}&ECCV20 &\makecell{Capsule network, triplet loss, attentive prototypes}\\
\cline{2-4}
& \makecell{CNL \cite{zhao2021looking}}&AAAI21 &\makecell{Task-specific and context-aware features, mutual information, task-specific scaling}\\
\cline{2-4}
& \makecell{\cite{zhu2021few}}&CVPR21 &\makecell{Dynamic meta-filters, neural ordinary differential equation}\\
\cline{2-4}
& \makecell{FRN \cite{wertheimer2021few}}&CVPR21 &\makecell{Feature map reconstruction}\\
\cline{2-4}
& \makecell{RENet \cite{kang2021relational}}&ICCV21 &\makecell{Self-correlational representation, cross-correlational attention}\\
\cline{2-4}
& \makecell{DMN4 \cite{liu2022dmn4}}&AAAI22 &\makecell{Mutual nearest neighbor, naive-bayes nearest neighbor}\\
\cline{2-4}
& \makecell{MCL \cite{liu2022learning} }&CVPR22 &\makecell{Mutual affiliations between query and support features}\\
\hline
\multirow{11}{*}{\rotatebox[origin=c]{90}{\makecell{Attention Modules}}} & \makecell{PARN \cite{wu2019parn}}&ICCV19 &\makecell{Deformable feature extractor, dual correlation attention}\\
\cline{2-4}
& \makecell{CAN \cite{hou2019cross}}&NeurIPS19 &\makecell{Cross attention, transductive inference}\\
\cline{2-4}
& \makecell{CoAE \cite{hsieh2019one}}&NeurIPS19 &\makecell{Co-attention, co-excitation, detection}\\
\cline{2-4}
&\makecell{ATL-Net \cite{dong2020learning}}&IJCAI20 &\makecell{Episodic attentions, adaptive threshold-based selection}\\
\cline{2-4}
& \makecell{IPN \cite{ma2020few}}&IJCAI20 &\makecell{Coarse-to-fine cognition, local and global relations}\\
\cline{2-4}
& \makecell{MattML \cite{zhu2020multi}}&IJCAI20 &\makecell{Base and task learners, a gradient-based meta-learning}\\
\cline{2-4}
& \makecell{GLoFA \cite{lu2021tailoring}}&AAAI21 &\makecell{Global and local masks, a mask combiner}\\
\cline{2-4}
& \makecell{TPMN \cite{wu2021task}}&ICCV21 &\makecell{Meta part filters, adaptive importance generator}\\
\cline{2-4}
& \makecell{\cite{huang2022enhancing}}&ICASSP22 &\makecell{Local-agnostic training, local-level similarity and knowledge transfer}\\
\cline{2-4}
& \makecell{CAD \cite{chikontwe2022cad}}&CVPR22 &\makecell{Cross-relational adaptation}\\
\cline{2-4}
& \makecell{TDM \cite{lee2022task}}&CVPR22 &\makecell{Support attention module, query attention module}\\
\cline{2-4}
& \makecell{SetFeat \cite{afrasiyabi2022matching}}&CVPR22 &\makecell{Set-to-set matching}\\
\hline
\multirow{12}{*}{\rotatebox[origin=c]{90}{\makecell{Transformers}}} & \makecell{CrossTransformers \cite{doersch2020crosstransformers}}&NeurIPS20 &\makecell{Self-supervised learning, part-based comparisons, spatial alignments}\\
\cline{2-4}
& \makecell{HCTransformers \cite{he2022attribute}}&CVPR22 &\makecell{Attribute surrogates learning, spectral tokens pooling}\\
\cline{2-4}
& \makecell{TraNFS \cite{liang2022few}}&CVPR22 &\makecell{Dynamic noise
rejection Transformer, median or similarity weighted aggregation}\\
\cline{2-4}
& \makecell{SSFormers \cite{chen2021sparse}}&Arxiv22 &\makecell{Sparse spatial Transformer, image patch matching}\\
\cline{2-4}
& \makecell{MG-ViT \cite{chen2022mask}}&Arxiv22 &\makecell{Task-relevant mask, residual connections, active learning based sample selection}\\
\cline{2-4}
& \makecell{SUN \cite{dong2022self}}&ECCV22 &\makecell{Location-specific supervision, background path filtration, spatial-consistent augmentation}\\
\cline{2-4}
& \makecell{TransVLAD \cite{li2022transvlad}}&ECCV22 &\makecell{Self-supervised learning, NeXtVLAD, supervision bias, simple-characteristic bias}\\
\cline{2-4}
& \makecell{tSF \cite{lai2022tsf}}&ECCV22 &\makecell{Dataset-attention module}\\
\cline{2-4}
& \makecell{FewTURE \cite{hiller2022rethinking}}&NeurIPS22 &\makecell{Masked image modeling, meta fine-tuning Transformers, inner loop token importance reweighting}\\
\cline{2-4}
& \makecell{STANet \cite{lai2023spatialformer}}&AAAI23 &\makecell{SpatailFormer semantic attention, SpatialFormer target attention}\\
\cline{2-4}
& \makecell{SMKD \cite{lin2023supervised}}&CVPR23 &\makecell{Intra-class knowledge distillation, masked patch tokens reconstruction}\\
\hline
\multirow{3}{*}{\rotatebox[origin=c]{90}{\makecell{Others}}} & \makecell{PDANet \cite{chen2021few}}&IJCAI21 &\makecell{Self-supervised learning, part discovery network, part augmentation network}\\
\cline{2-4}
& \makecell{COSOC \cite{luo2021rectifying}}&NeurIPS21 &\makecell{Clustering-based object seeker, shared object concentrator}\\
\cline{2-4}
& \makecell{\cite{wu2023bi}}&AAAI23 &\makecell{Mutual support-query and query-support reconstruction}\\
\hline
\end{tabular}
\label{tab_fsl}
\end{table*}

\subsection{Attention Modules}

Attention modules are used to localize different task-specific object parts based on task information, meeting the requirements of different tasks.

There exist two issues in the FSL problem: 1) Similar semantic objects may appear at two different spatial positions; 2) Even if coarse objects are close in spatial positions, fine-grained parts do not. A position-aware relation network (PARN) \cite{wu2019parn} is proposed to alleviate these issues. Particularly, a deformable feature extractor is introduced to extract fewer low-responses or unrelated semantic representations, alleviating the issue 1). Besides, a dual correlation attention module is designed to compare objects or fine-grained features in different positions, addressing the issue 2).

The learned features in exiting works may be not discriminative enough because they are extracted from support and query images separately. This can be explained by the following two reasons: 1) An image may contain multiple objects. It is very likely that only objects from seen classes are located, while target objects from unseen classes are neglected; 2) The learned features from few labeled support images may be not representative for the true class distribution. To address this issue, cross attention networks (CAN) \cite{hou2019cross} are proposed. To be specific, a cross attention module learns the semantic relations between support and query features, adaptively localizing discriminative parts. Besides, a transductive inference algorithm alleviates the issue of limited data by iteratively generating pseudo-labeled query samples to augment the support set.

A co-attention and co-excitation (CoAE) is proposed in \cite{hsieh2019one} for object detection, which can detect all instances that belong to the same class in a target image. First, the co-attention operation is used to explore correlated evidences encoded in query-target pairs. Then, the co-excitation highlights discriminative features shared by both the query and the target images with a similar architecture as squeeze-and-excitation networks \cite{hu2018squeeze}.

It is observed that different local patches play different roles in varying tasks. For example, distinguishing between cats and birds is significantly different from differentiating between cats and cars where the head is more important in the former task. Therefore, local patches for a query image should be adaptively adjusted according to different tasks. An adaptive task-aware local representations network (ATL-Net) \cite{dong2020learning} can achieve this target. Specifically, a novel episodic attention mechanism learns discriminative patches to extract task-aware local features for FSL. Besides, an adaptive threshold-based selection strategy is designed to select discriminative patches for different tasks.

Most approaches only focus on either global or local representations, suffering from poor generalization abilities. An inverted pyramid network (IPN) \cite{ma2020few} intimates the human's coarse-to-fine cognition paradigm. It consists of a global stage and a local stage. The former learns support-query relations and generates query-to-class similarities based on the contextual memory. The latter further complements the coarse relations, providing precise query-to-class similarities.

Few works focus on applying FSL on fine-grained classification where local representations are important. A multi-attention meta-learning (MattML) method \cite{zhu2020multi} utilizes attention mechanisms of the base and task learners to extract discriminative image parts. The base learner consists of two convolutional block attention modules (CBAM) and a classifier. CBAM can learn diverse discriminative parts. The weights of the classifier are initialized by the task learner, enabling the classifier to have a task-related sensitive initialization. A gradient-based meta-learning updates the parameters of two CBAMs and the classifier, allowing the base learner to emphasize discriminative parts.  

Different feature dimensions should be emphasized for different tasks. Many recent methods focus on extracting task-specific features. However, they ignore the differences of various classes and generally apply a global transformation to the task. A global and local feature adaptor (GLoFA) \cite{lu2021tailoring} tailors task-specific representations with global and local feature masks. The global mask captures sketchy patterns, and local masks focus on detailed characteristics. Then, a mask combiner incorporates global and local masks with the target task context.  

Prior methods divide images into local patches using grids on feature maps. Consequently, some patches may only contain a small part of semantic regions of the object, causing misalignments when matching the local patches. Besides, a common set of local regions are used in different tasks, limiting the performance in diverse tasks with large distribution differences. A task-aware part mining network (TPMN) \cite{wu2021task} is proposed. More specifically, a group of meta part filters are proposed to automatically generate task-related local parts based on the task embedding in a meta-learning way. Also, an adaptive importance generator suppresses less useful parts and emphasizes important ones. 

Existing works often have the discriminative location bias, which causes models to over-emphasize image parts for the base classes, while neglecting their discriminative powers for unseen categories. To tackle this issue, \cite{huang2022enhancing} proposes several local level methods. First, a local agnostic training approach alleviates the discriminative location bias between base and unseen classes. Second, a local level similarity method calculates the comparison between local level features. Third, a local level knowledge transfer synthesizes different knowledge transfers from the base class for different location features.

Some existing methods use spatial representations to learn pixel-level correspondence, while only achieving slight improvements. CAD \cite{chikontwe2022cad} leverages self-attention to co-attend both query and support features. In particular, a shared module can learn co-adapted important embeddings for query and support images in two perspectives: 1) query to support; 2) support to query. Besides, it re-weights spatial attention maps with the corresponding prototypes to attend to important relevant parts. 

It is important to localize discriminative details for fine-grained few-shot classification. Task Discrepancy Maximization (TDM) \cite{lee2022task} can emphasize channels which contain discriminative information of the class, localizing class-wise discriminative regions. Specifically, it consists of a support attention module and a query attention module. The former generates support weights per class that represent high responses on discriminative channels. However, the prediction is biased towards the support set. The latter outputs query weights per instance that highlight object-relevant channels for a given query image. By combining these two weights, task-adaptive feature maps are generated to focus on important details.

To capture set-based image representations from base classes that better transfer to novel classes, SetFeat \cite{afrasiyabi2022matching} adopts self-attention mechanisms at different stages of convolutional networks. Consequently, set-to-set matching metrics predict the class of a query image based on the support set.

\textbf{Pros and cons}. Different mechanisms are designed to learn task-aware local representations for different tasks. While they achieve decent performance, task-specific information changes according to novel classes. The model needs to re-trained to meet the requirement of new classes. Consequently, the training process for new unseen classes is complex. Besides, attention modules generate mutually similar regions between the query set and the support set, which are regarded as the components of target objects. However, the attention map may be inaccurate where the mutually similar background is located to distract the model.

\subsection{Transformers}

Transformers can be used to learn relations among different patch tokens, providing rich contextual information. 

It is observed in CrossTransformers \cite{doersch2020crosstransformers} that the supervision collapse usually occurs in the FSL where some information is unnecessary for the training task, but plays an important role in transferring to new tasks. To overcome this issue, self-supervised learning is used to learn transferable general-purpose features. Besides, a novel Transformer network is proposed for part-based comparisons and spatial alignments, improving the generalization ability to unseen classes. 

Since there is no sufficient data in FSL, the ViT tends to suffer from the overfitting issue, leading to a performance degradation. To improve data efficiency, hierarchically cascaded transformers (HCTransformers)  \cite{he2022attribute} are proposed which consists of attribute surrogates learning and spectral tokens pooling. The former can well supervise the learning of both the class and patch tokens, and the latter learns relationships between image parts in both the spatial layout and semantic space.

Noisy data in the few-shot setting has a destructive negative effect. A novel Transformer model for Noisy Few-Shot Learning (TraNFS) \cite{liang2022few} can leverage the attention mechanism from the Transformer to dynamically reject noisy data. Besides, A median or similarity weighted aggregation is employed to replace the mean aggregator.

Most existing methods use global features or dense features for the FSL task, which suffer from losing either local or contextual information. Besides, they process each class separately, failing to leverage task-specific features. To address these issues, Sparse Spatial transFormers (SSFormers) \cite{chen2021sparse} are proposed, which can learn task-relevant features and suppress task-irrelevant ones. Specifically, each input image is divided into several image patches with varying sizes to extract local features, preserving contextual features, followed by a sparse spatial Transformer layer which matches spatial correspondences between the query image and the support set, focusing on task-relevant image patches and suppressing task-irrelevant ones. Last, an image patch matching module calculates distances between dense local features, generating category predictions.

It is noticed in \cite{dong2022self} that ViTs slowly learn the relations among input tokens because of a lack of inductive biases, leading to a performance drop. To address this issue, Self-promoted sUpervisioN (SUN) is proposed. Specifically, it uses ViTs to generate location-specific guidances, which tell the ViTs about the similarity of patch tokens, accelerating token relation learning and boosting the object grounding and recognition ability. Moreover, the location-specific supervision is boosted by background path filtration and spatial-consistent augmentation where the former alleviates the issue where background patch tokens are mis-classified as semantic categories and the latter provides additional diversity for data augmentation. 

Since the ViTs are data-intensive, they are less explored in existing fine-tuning based FSL approaches which have few labeled support images, limiting the generalization ability to unseen classes and deteriorating the performance. A novel mask-guided vision Transformer (MG-ViT) is proposed in \cite{chen2022mask} to apply a mask on image patches, emphasizing task-relevant discriminative patches and filtering out task-irrelevant ones. Besides, a residual connection is used to preserve global representations of visible image patches, maintaining the structure of the ViT. Moreover, an active learning based sample selection approach selects representative few-shot images, further improving the generalization ability.

In TransVLAD \cite{li2022transvlad}, self-supervised learning is firstly explored to deal with the overfitting issue of Transformers. Besides, NeXtVLAD aggregates local features to boost the FSL result. Moreover, the supervision bias and simple-characteristic bias are introduced to alleviate the few-shot biases. 

Most existing feature embedding methods are designed specifically for learning tasks (e.g., classification, detection, and segmentation), which have limited utility. To address this issue, transformer-based Semantic Filter (tSF) is proposed in \cite{lai2022tsf} to use dataset-attention module to transfer the knowledge from the base set to the novel set and filter semantic features for the target class.

Image-level annotations only describe a small subset of an image's content, leading to performance drops under scenarios where classes differ significantly between training and test time. Besides, supervision collapse can occur where certain image patterns are overemphasized during traning while become useless during testing. To address these problems, FewTURE is proposed in \cite{hiller2022rethinking}. In particular, self-supervised learning with masked image modeling is employed as pretext task, addressing the lack of fine-grained annotations. Besides, meta fine-tuning of Vision Transformers is combined with inner loop token importance reweighting, learning generalized features and avoiding supervision collapse.

CNN based approaches generate inaccurate attention maps based on local features and mutually similar background which may distract the model. A SpatialFormer model, called STANet, is proposed in \cite{lai2023spatialformer} to output accurate attention regions based on global features. Specifically, the SpatailFormer semantic attention emphasizes the mutual semantic similar regions. The SpatialFormer target attention locates foreground object parts of novel classes which are similar with base classes.

A novel Supervised Masked Knowledge Distillation model (SMKD) is proposed in \cite{lin2023supervised} which integrates label information into self-distillation. Intra-class knowledge distillation is applied on both class and patch tokens, leveraging multi-scale information. Masked patch tokens reconstruction is conducted across intra-class images, increasing the generalization ability of Transformers.

\textbf{Pros and cons}. Since it is required to infer properties of translation invariances, locality and hierarchical structure from the data, Transformers need more training data compared with CNNs, preventing them from being directly used in scenarios of scarce data like few-shot learning. Fine-tuning heavy parameters of Transformer models with just a few examples leads to the poor generalization issue. Therefore, it is worth exploring how to improve the generalization ability of data-hungry Transformers with few labeled training examples.

\subsection{Others}

Different mechanisms are also used to benefit the FSL, like self-supervised learning,  clustering, and reconstruction.

Meta-learning can be used to find inductive biases on similar tasks, generalizing to unseen classes. Unsupervised learning can be used to learn prior knowledge from unlabeled images. However, due to without labels, it tends to learn distracted images, like background. Part discovery and augmentation network (PDANet) \cite{chen2021few} is proposed to address this issue. To be specific, the part discovery network can learn discriminative part features from random crops in unlabeled images with self-supervised learning. The part augmentation network augments few support examples with relevant part features, alleviating the over-fitting issue. Self-supervised learning can learn less biased features towards base classes and alleviate the over-fitting issue, leading to a better generalization ability towards novel categories. However, the optimization objective of self-supervision and supervision is conflicting. It is necessary to balance them during training, simultaneously leveraging the strengths of self-supervised learning and supervised learning.

It is argued that image background is one harmful source knowledge for the FSL. This is because there exist correlations between background and image categories in training, possibly failing to generalize well to novel classes. COSOC is proposed in \cite{luo2021rectifying} to extract foreground of images during both training and evaluation. Specifically, it consists of a clustering-based object seeker (COS) and a shared object concentrator (SOC). The former seeks shared local patterns with the contrastive learning and clustering algorithm in training. The latter applies iterative feature matching in the support set to find potential foreground regions in evaluation.

Previous works use a support set to reconstruct the query image. However, it is revealed that the unidirectional methods perform poorly in tackling inter-class and intra-class variations. To address this issue, a bi-reconstruction mechanism is introduced in  \cite{wu2023bi} to simultaneously increase inter-class variations and decrease intra-class variations by mutual support-query and query-support reconstruction which can learn subtle and discriminative features.

\textbf{Pros and cons}. Other modules include clustering, reconstruction, and self-supervised learning can extract unsupervised features. The unsupervised feature learning methods are complementary with current methods. However, since unsupervised features are not guided under a proper signal, the complmentarity between supervised features and unsupervised features are not explored thoroughly, possibly degrading the performance.

\subsection{Discussions}
The representative FSL models with local mechanisms are summarized in Table \ref{tab_fsl}. 

A good FSL model should be able to overcome the overfitting issue and learn discriminative features from few labeled images. To overcome the overfitting issue, it is necessary to design a simple network which tends to have a weak ability to represent images. Therefore, it is worth exploring how to balance the overfitting issue of models and representational ability of features. Given few labeled images in the FSL, extracting transferable and general local representations can achieve the target.

Besides, although there are possibly only few labeled samples in a novel class, numerous unlabeled images contain rich representations, especially local representations, which can be explored to benefit the FSL task.

\section{Zero-Shot Learning}\label{sec_zsl}

\begin{figure}[ht!]
\centering
\includegraphics[width=0.46\textwidth]{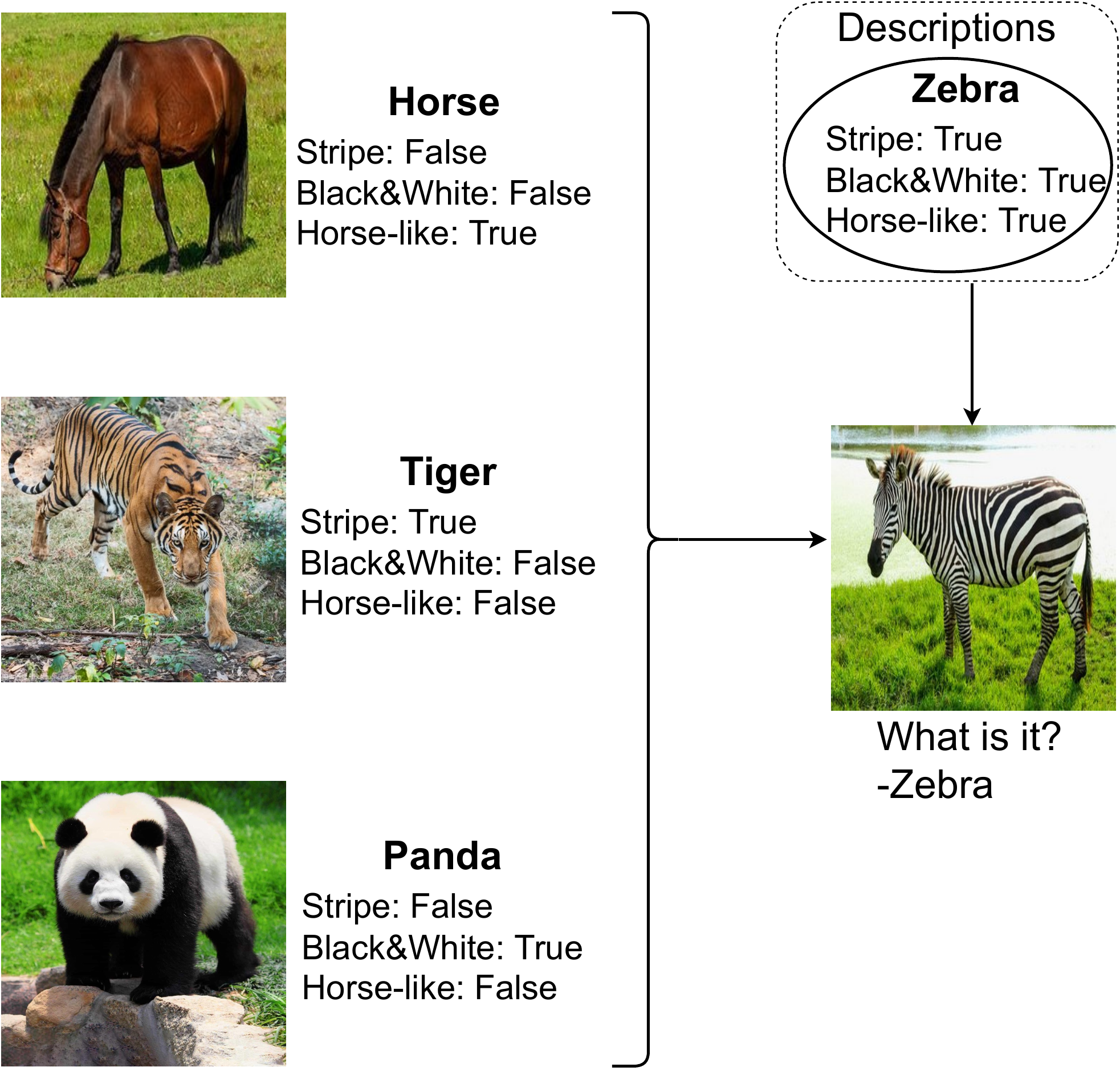}
\caption{The general pipeline of zero-shot learning (ZSL).}
\label{fig_zsl_pipeline}
\end{figure}

``What does a zebra look like? It is horse-like with black \& white stripes". Zebras are not native to China, yet knowledgeable travelers can describe them in relation to Chinese native  animals, like horses, tigers, and pandas, as shown in Fig. \ref{fig_zsl_pipeline}. This can inspire people who have never seen zebras before imagine and identify them. On the other hand, although deep learning has obtained significant success in computer vision, its performance relies heavily on the number of annotated data. However, in the real world, it is sometimes necessary to recognize novel classes without annotated images. In order to mimic the impressive ability of humans to identify unseen objects from pure language descriptions, the zero-shot learning (ZSL) is proposed. Due to data unavailability of novel categories during training, the ZSL is achieved by transferring knowledge from seen classes with category descriptions to bridge the gap between seen and unseen classes, including visual attributes \cite{farhadi2009describing,ferrari2007learning}, word embeddings \cite{lei2015predicting,li2017zero}, and sentences \cite{reed2016learning}.

\begin{figure}[ht!]
\centering
\includegraphics[width=0.34\textwidth]{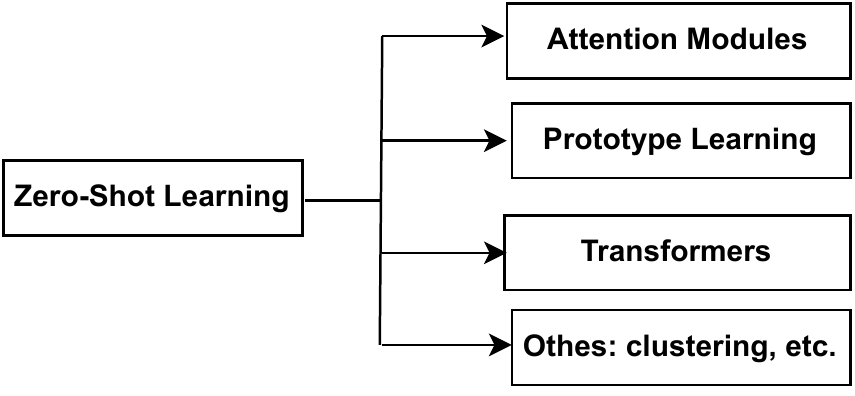}
\caption{The family of ZSL with local mechanisms.}
\label{fig_zsl_family}
\end{figure}

Existing works can be divided into generative methods and embedding approaches. Generative methods generate synthetic data via the category descriptions of unseen classes, transforming the ZSL into a supervised learning problem without bias toward seen or unseen classes. Embedding approaches map the input image into a shared vector space where the prediction is achieved by searching the nearest class. Some works try to align global image representations with semantic vectors in a joint embedding space \cite{frome2013devise,annadani2018preserving}. However, since there exist large domain gaps between seen classes and unseen categories, unseen images tend to be misclassified as seen classes. All these methods rely on global features of the whole image. However, there exist two drawbacks for these approaches: 1) it is hard to distinguish between seen and unseen images in the global feature space; 2) because semantic vectors of some categories are very similar, it is difficult to match similar images with confused semantic vectors. It is observed that local feature can better transfer knowledge from seen to unseen classes. Besides, some parts can better distinguish local appearance differences for the same attribute. Motivated by these two observations, recent works explore local features in ZSL. As shown in Fig. \ref{fig_zsl_family}, current ZSL methods with local mechanisms can be divided into the following categories: 1) Attention modules: they can automatically locate discriminative object parts under the supervision of signals, like attribute information, gaze estimations; 2) Prototypes: several methods are designed to learn transferable prototypes, including attribute-level and category-level prototypes. They can transfer from seen classes to unseen classes based on the learned prototypes; 3) Transformers: The popular Transformers are also explored in this field; 4) Others: other local feature learning methods are also used, including part clustering.

\subsection{Attention Modules}

Without relying on part annotations or detection, several works explore attention mechanisms to automatically discover discriminative parts in images, enabling models to match semantic representations accurately.

It is observed that humans tend to focus on discriminative parts based on class semantics. Besides, humans achieve the semantic alignment by locating the most related parts progressively, and filtering out the uninformative ones. Motivated by this observation, novel stacked semantics-guided attentions (S\textsuperscript{2}GA) \cite{ji2018stacked} are proposed for fine-grained ZSL. Guided by class semantic descriptions, different regions are assigned with different weights progressively. Both global and local features are incorporated to represent an image.

It is noticed that few existing works explore the discriminative power of local image parts, which have a better discrimination ability than attributes to boost the semantic transfer between seen and unseen classes. An attentive region embedding network (AREN) is proposed in \cite{xie2019attentive} to mine semantic regions. Specifically, an attentive region embedding stream is proposed to learn different attention maps under the guidance of high-level semantic attributes, discovering multiple part regions. Moreover, an adaptive thresholding module is designed to suppress responses on redundant regions. Besides, an attentive compressed second-order embedding is proposed to guarantee stable semantic transfer. 

It is claimed in VSE \cite{zhu2019generalized} that the semantic gap exists in prior works because visual features is not semantic. To bridge the gap, a new statistical model is proposed to embed a visual instance into a low-dimensional probability matrix. Besides, additional constraints are utilized to encourage attention maps to be compact within the same part and divergent among different parts, learning multiple parts. 

Prior works usually treat attributes equally, ignoring their different discriminative abilities. A latent feature guided attribute
attention (LFGAA) network \cite{liu2019attribute} performs object-level attribute attentions for semantic disambiguation by distracting semantic activations that cause ambiguity.

Unlike existing approaches which mainly focus on learning a proper mapping function between visual and semantic embeddings, a semantic-guided multi-attention localization model is proposed in \cite{zhu2019semantic} to automatically locate the most important objects. This is achieved by a multi-attention loss which applies geometric constraints over attention maps to learn compact and diverse attention distributions. Meanwhile, under the guidance of embedding softmax loss and class-center triplet loss, global and local representations are simultaneously learned to provide an enhanced visual features.

Traditional models perform global embedding between visual features and class semantic descriptions, which would miss local visual features. A dense attribute-based attention network (DAZLE) is proposed in \cite{huynh2020fine} to learn attentions for each attribute that locate the most discriminative parts related with each attribute for learning attribute-based features, followed by an attribute embedding method to align each attribute-based features with its attribute semantic vectors. Besides, a self-calibration loss to tackle prediction bias towards seen classes by adjusting the probability of unseen classes.

Most existing works fail to capture the appearance relationships among different image parts. A Region Graph Embedding Network (RGEN) is proposed in \cite{xie2020region} to address this issue which consists of the constrained part attention (CPA) branch and the parts relation reasoning (PRR) branch. The former generates the located object parts on each input image, and the latter models the region-based relations with graph convolutional networks \cite{kipf2016semi}. Besides, a novel balance loss is proposed to address the severe domain bias problem in GZSL.

One family of approaches for ZSL is to generate unseen images with auxiliary semantic information. However, these approaches suffer from two drawbacks: 1) They are easily affected by distracted background. 2) No mechanisms are designed to weight the importance of different parts. A novel divide-and-conquer
method, named multi-patch generative adversarial nets (MPGAN), is proposed\cite{chen2020rethinking}. It can generate a set of predefined local patches with multiple generative models. Besides, a voting mechanism is designed to measure the importance of each patch based on its discriminative ability.

Prior works mainly use one-off object localizer with a non-transparent region localization process, lacking interpretability and flexibility. A novel Semantic-guided
Reinforced Region Embedding (SR2E) network is proposed in \cite{ge2021semantic}. A reinforced region module is proposed where the reinforcement learning is firstly used to teach the localizer to search related regions under the guidance of semantic-driven reward. Besides, a semantic alignment module is developed to perform semantic-visual alignment to preserve the semantic relationship.

Motivated by the observation that human would automatically gaze at parts with certain semantic clues when recognizing unseen images, a goal-oriented gaze estimation method (GEM-ZSL) is proposed in \cite{liu2021goal} to predict the human gaze locations that are transformed to attribute attentions for recognizing unseen classes.

Most attention-based approaches use unidirectional attention, which only discover limited semantic alignments between visual and attribute features. A mutually semantic distillation network (MSDN) \cite{Chen2022MSDN} is proposed to progressively distill the inherent semantic features between visual and attribute representations. More specifically, it consists of an attribute$\rightarrow$visual attention sub-net and a visual$\rightarrow$attribute attention sub-net to extract attribute-based visual features and visual-based attribute features. Besides, a semantic distillation loss is introduced to mutually train these two sub-nets.

\textbf{Pros and cons}. Attention mechanisms are the mostly widely used in the ZSL. They are usually used to highlight discriminative visual regions under some semantic signals, like attributes, capturing subtle but important differences between various categories. The semantic-visual interactions can be utilized to transfer between seen and unseen classes. However, it should be pointed out although different signals are used to guide the learning of attentions, it is inevitable that some object parts are missed, especially under some challenging scenarios, like occlusions. Some approaches try to address this issue by proposing diversity losses to learn diverse attentions. However, some image details may be still missed because no modules are designed to mine comprehensive parts.

\subsection{Prototypes}

Prototypes can correlate each attribute or category with semantic visual attentions, learning attribute or category related prototypes.  

Although attention modules can locate some discriminative parts, parts and attributes are biased towards seen classes due to the relations among attributes which can maximize the likelihood of training data, but fail to generalize to unseen classes. An attribute prototype network (APN) is proposed in \cite{xu2020attribute} which jointly regresses and decorrelates attributes from intermediate layers to learn local features with semantic visual attributes. The APN learns prototypes to describe visual patterns related with each attribute. However, learned prototypes are shared for all images, limiting their effectiveness for images with large variances. A dual progressive prototype network (DPPN) is proposed in \cite{wang2021dual} where attributes prototypes are dynamically adjusted according to different images and category prototypes progressively enlarge category margins, preserving attribute-region correspondence and enhancing category discriminability, respectively. Consequently, it has an excellent attribute localization and transferability ability between seen and unseen classes.

It is argued that attribute localization is beneficial for zero-shot and few-shot image classification tasks. To this end, attribute prototype network (APN) \cite{xu2022attribute} is proposed to jointly learn global and local representations with class-level attributes. It learns local features with attribute prototype network which can simultaneously regressing and decorrelating attributes in intermediate features. Furthermore, a zoom-in module is introduced to localize and crop informative regions, learning discriminative features.

\textbf{Pros and cons}. Prototypes is used to learn attribute and category related prototypes, based on which to transfer the knowledge from seen to unseen classes. The shared attribute prototypes are used to localize attribute-related regions. The prototypes are generated by different input images which can alleviate effects of noisy images. However, they ignore mutual interactions between semantic and visual domains, failing to alleviate the semantic ambiguity problem which can appear due to the large visual variances for each attribute, such as attribute noses from dolphins and birds. Besides, prototypes lack a mechanism to take comprehensive image details into consideration. Furthermore, relations among different prototypes are failed to be built.

\begin{table*}[ht!]
\centering
\caption{Representative ZSL models with local mechanisms.}
\begin{tabular}{|c|c|c|c|c||c|c|c|c|c|}
\hline
\multicolumn{1}{|c|}{\multirow{1}{*}{Category}} & \multirow{1}{*}{Methods} & \multirow{1}{*}{Venue} & \multirow{1}{*}{Highlights} \\
\hline
\multirow{13}{*}{\rotatebox[origin=c]{90}{\makecell{Attention Modules}}}& \makecell{S\textsuperscript{2}GA \cite{ji2018stacked}}&NeurIPS18 &\makecell{Stacked semantics-guided attention}\\
\cline{2-4}
& \makecell{AREN \cite{xie2019attentive}}&CVPR19  &\makecell{Attentive region embedding, adaptive thresholding,\\ attentive compressed second-order embedding}\\
\cline{2-4}
& \makecell{VSE \cite{zhu2019generalized}}&CVPR19 &\makecell{A statistical model}\\
\cline{2-4}
& \makecell{LFGAA \cite{liu2019attribute}}&ICCV19  &\makecell{Semantic disambiguation}\\
\cline{2-4}
& \makecell{\cite{zhu2019semantic}}&NeurIPS19 &\makecell{A multi-attention loss, embedding softmax loss, class-center triplet loss}\\
\cline{2-4}
& \makecell{DAZLE \cite{huynh2020fine}}&CVPR20 &\makecell{Attribute-based features, an attribute embedding, self-calibration loss}\\
\cline{2-4}
& \makecell{RGEN \cite{xie2020region}}&ECCV20 &\makecell{Constrained part attention, parts relation reasoning, balance loss}\\
\cline{2-4}
& \makecell{MPGAN \cite{chen2020rethinking}}&MM20 &\makecell{Multiple generative models, a voting mechanism}\\
\cline{2-4}
& \makecell{SR2E \cite{ge2021semantic}}&AAAI21 &\makecell{A reinforced region module, a semantic alignment module}\\
\cline{2-4}
& \makecell{GEM-ZSL \cite{liu2021goal}}&CVPR21 &\makecell{Predict human gaze locations}\\
\cline{2-4}
& \makecell{MSDN \cite{Chen2022MSDN}}&CVPR22 &\makecell{Attribute-based visual features, visual-based \\attribute features, mutual learning}\\
\hline
\multirow{1}{*}{\rotatebox[origin=c]{90}{\makecell{Prototypes}}}
& \makecell{APN \cite{xu2020attribute}}&NeurIPS20 &\makecell{Attribute prototype, \\attribute decorrelation}\\
\cline{2-4}
& \makecell{DPPN \cite{wang2021dual}}&NeurIPS21 &\makecell{Progressive attributes \\and category prototypes}\\
\cline{2-4}
& \makecell{APN \cite{xu2022attribute}}&IJCV22 &\makecell{Attribute prototype network\\ a zoom-in module}\\
\hline
\multirow{10}{*}{\rotatebox[origin=c]{90}{\makecell{Transformers}}}& \makecell{TransZero \cite{Chen2022TransZero}}&AAAI22 &\makecell{A feature augmentation encoder, a visual-semantic decoder}\\
\cline{2-4}
& \makecell{TransZero++ \cite{chen2022transzero++}}&TPAMI22&\makecell{Attribute$\rightarrow$visual Transformer, visual$\rightarrow$attribute Transformer, \\semantic collaborative loss}\\
\cline{2-4}
& \makecell{DPDN \cite{ge2022dual}}&MM22&\makecell{Attribute-guided part discovery, category-guided part discovery}\\
\cline{2-4}
& \makecell{I2Dformer \cite{naeem2022idformer}}&NeurIPS22&\makecell{Local paired features from noisy documents,\\fine-grained cross-modal interactions}\\
\cline{2-4}
& \makecell{DUET \cite{chen2023duet}}&AAAI23&\makecell{Pre-trained language models, cross-modal semantic \\grounding network, attribute-level contrastive learning}\\
\cline{2-4}
& \makecell{PSVMA \cite{liu2023progressive}}&CVPR23&\makecell{Instance-motivated semantic Transformer-based encoder, \\semantic-motivated instance Transformer-based decoder, a debiasing loss}\\
\hline
\multirow{1}{*}{\rotatebox[origin=c]{90}{\makecell{Others}}}& \makecell{VGSE \cite{xu2022vgse}}&CVPR22 &\makecell{Image region clusters, \\class relation discovery}\\
\cline{2-4}
& \makecell{MUST \cite{li2023masked}}&ICLR23&\makecell{Self-training target, masked image modeling, \\global-local feature alignment}\\
\hline
\end{tabular}
\label{tab_zsl}
\end{table*}

\subsection{Transformers}

It is observed that several limitations exist in previous methods: 1) Entangled part features are used, limiting the transferability of visual features from seen to unseen classes; 2) Region representations are learned, while the importance of discriminative attribute localization is neglected. To address these challenges, an attribute-guided Transformer, called TransZero \cite{Chen2022TransZero}, is proposed which consists of a feature augmentation encoder and a visual-semantic decoder. This is the first work to extend the Transformer to the ZSL. An extension work, TransZero++ \cite{chen2022transzero++}, is proposed. Several aspects are improved compared to TransZero: 1) Except the attribute$\rightarrow$visual Transformer in TransZero, the visual$\rightarrow$attribute Transformer is proposed to improve the performance of attribute localization. These two representations achieve desirable visual-semantic interactions; 2) Feature-level and prediction-level semantic collaborative losses are introduced to discover rich semantic information between visual and attribute features.

Recent methods over-rely on attribute information and neglect the category features, resulting in poor generalization. A novel Dual Part Discovery Network (DPDN) is proposed in \cite{ge2022dual} where both attribute and category information are taken into consideration. In specific, the attribute-guided part discovery module relates regions with specific attribute information. The category-guided part discovery module mines important details that are complementary with attribute-guided parts.

Popular ZSL methods usually depend on human annotated attributes as the side information, which are labor-expensive to annotate and hard to scale to large-scale datasets. To this end, Image to Document Transformer (I2Dformer) \cite{naeem2022idformer} is proposed to align global representations of image and document pairs and local representations of image parts and document words in a shared embedding space from noisy documents. Besides, a cross-modal attention module learns fine-grained interactions between image patches and document words.    

Attributes are the most widely used semantic information to describe classes. However, they fail to capture the subtle visual differences between images due to the shortage of fine-grained annotations and issues of attribute imbalance and co-occurrence. CrossmoDal semantic groUnding for contrastivE zero-shoT learning (DUET) \cite{chen2023duet} is proposed to address these issues. Specifically, a cross-modal semantic grounding network is developed to transfer inherent semantic information from the pre-trained language models (PLMs) to the ViT. It is the first attempt to integrate PLMs in the ZSL. An attribute-level contrastive learning is proposed to address the issues of attribute imbalance and co-occurrence. 

Prior works localize regions with the shared attributes. If different visual appearances correspond to the same attributes, the shared attributes tend to have semantic ambiguity, leading to inaccurate semantic-visual interactions. To address this issue, Progressive Semantic-Visual Mutual Adaption (PSVMA) \cite{liu2023progressive} is proposed to progressively build the correspondences between attribute prototypes and visual features. An instance-motivated semantic Transformer-based encoder learns instance-centric prototypes adapting to different images, recasting the unmatched semantic-visual pair into the matched one. A semantic-motivated instance Transformer-based decoder builds accurate cross-domain interactions for learning transferable and unambiguous visual features. A debiasing loss is proposed to balance predictions towards unseen classes. 

\textbf{Pros and cons}. Transfomers can take comprehensive object parts into consideration and build global-range relations among these object parts. However, some mechanisms are expected to solve the hard-split issues where semantically similar parts may appear at different token positions, limiting the representational ability of subtle visual features. Besides, the semantic ambiguity problem can also appear. It is challenging to construct cross-domain interactions for ambiguous visual-semantic pairs.

\subsection{Others}

Human annotated attributes are important to transfer between seen and unseen classes. However, obtaining attributes involves a labor-intensive  process. Some works rely on embeddings from class names or encyclopedia articles. They tend to suffer from a sub-optimal performance because class relations may not be detected. Visually-grounded semantic embedding (VGSE) \cite{xu2022vgse} can discover important semantic embeddings without relying on human knowledge. First, a set of images are divided into clusters of local image regions based on visual similarity. Second, word embeddings and a novel class relation discovery module are used to relate learned clusters with unseen classes.

The zero-shot performance of CLIP-like models are limited in real-world scenarios. To address this issue, Masked Unsupervised Self-Training (MUST) \cite{li2023masked} is proposed to use two complementary training signals: pseudo-labels and original images. In particular, a self-training target is introduced to learn task-specific global prediction. Masked image modeling is designed to learn local pixel-level information. Global-local feature alignment is used to associate the knowledge from the above two supervisions. 

\textbf{Pros and cons}. Other modules are also explored to learn transferable parts, like part clustering and self-supervised learning, which can be used to transfer from seen to unseen classes. Clustering can automatically group images into clusters of object parts. Self-supervised learning can learn task-agnostic features. However, since part clusters are formed without a strong supervision signal, some image clusters may be not discriminative enough to specific tasks, possibly learning useless or even noisy clusters and missing some important object parts. Self-supervised features require an additional stage to fine-tune models on labeled data, adapting features to specific tasks.

\subsection{Discussions}
The representative zero-shot learning models with local mechanisms are summarized in Table \ref{tab_zsl}. 

Most existing works use human-labeled attributes as shared auxiliary information between seen and unseen classes to transfer knowledge from seen classes to unseen categories. Despite their promising performance, the attributes are difficult to annotate and expensive to scale. Unsupervised alternatives worth exploring. Online text documents, like Wikipedia, can provide rich auxiliary information for ZSL. Besides, large language models which are trained on large-scale text can also generate high-quality auxiliary class information. However, the granularity between class descriptions and visual cues should be properly aligned.

\section{Multimodal Learning}\label{sec_mml}

\begin{figure}[ht!]
\centering
\includegraphics[width=0.36\textwidth]{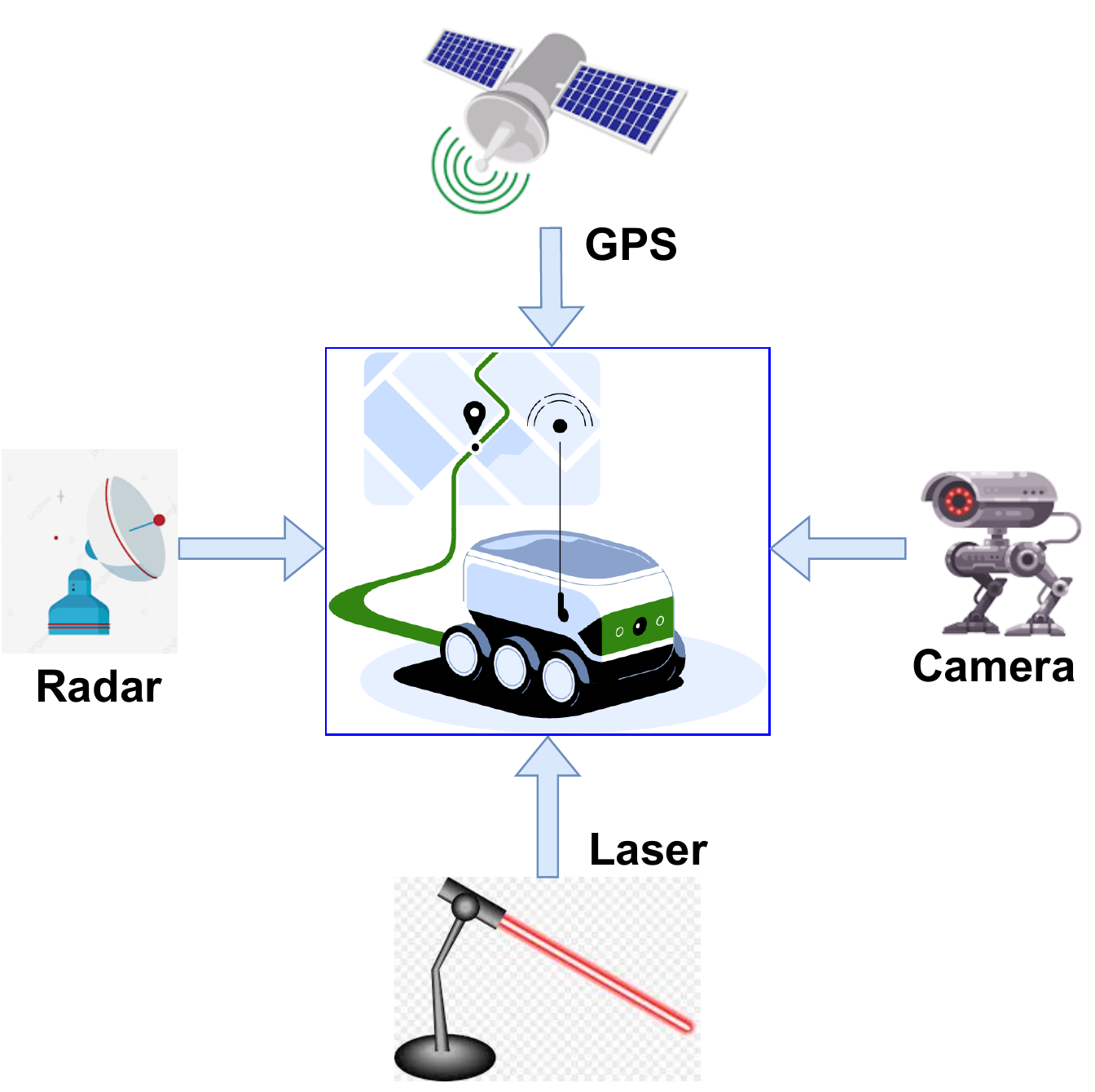}
\caption{Multimodal sensors used in navigation robots to know the real-world environment.}
\label{fig_mml_pipeline}
\end{figure}

Multimodal learning (MML) is imitating human perception abilities to leverage multiple modalities, engaging machines properly with the world (e.g., seeing, talking, reading). For example, as shown in Fig. \ref{fig_mml_pipeline}, navigation robots make precise decisions under the real-world environment through multimodal sensors, like camera, radar, GPS, and laser. Thanks to the development of the Internet and widespread intelligent devices in recent decades, the quantity of multimodal data is increasing. Meanwhile, MML can extract comprehensive information from multimodal data \cite{baltruvsaitis2018multimodal}. Thus, the research on MML is becoming increasingly popular. For example, visual question answering \cite{antol2015vqa} outputs a natural language answer given an image/video and a natural language question about the image/video; image-text retrieval \cite{lin2014microsoft} retrieves images given descriptions or finds sentences from query images; Natural language for visual reasoning for real \cite{suhr2018corpus} determines if a caption is correct with regard to a pair of images. 

\begin{figure}[ht!]
\centering
\includegraphics[width=0.40\textwidth]{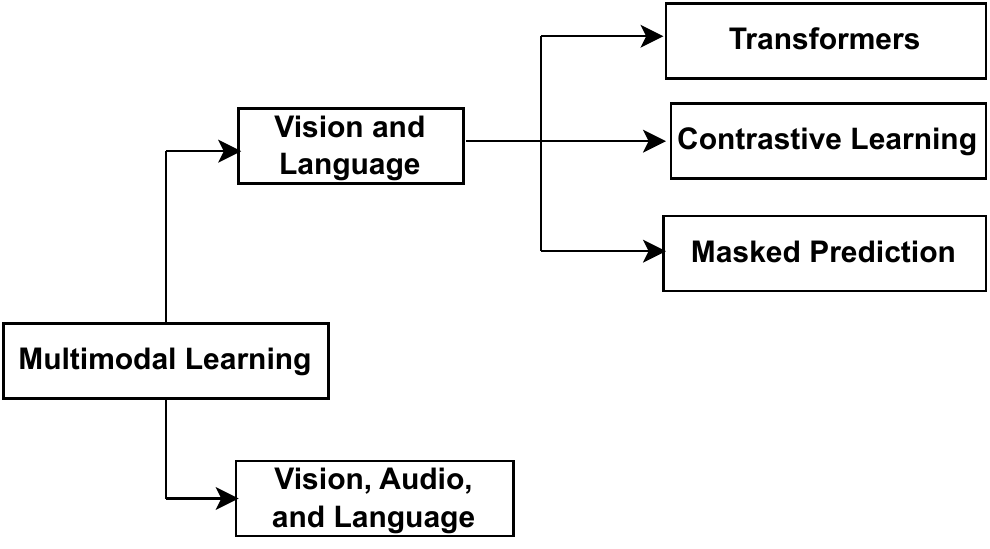}
\caption{The family of multimodal learning (MML) with local mechanisms.}
\label{fig_mml_family}
\end{figure}

In this survey, we mainly focus on research on vision, audio, and language, as shown in Fig. \ref{fig_mml_family}. 

\subsection{Vision and Language}

Vision and language models first use a text encoder and an image encoder to learn text and image representations, and then extract the vision-language correlations with certain pre-training objectives. Two types of networks are generally used as encoders to learn image and text features, i.e., CNN models and Transformer models. Since Transformers is gaining increasingly popular, we mainly focus on Transformer encoders in this survey. As another core, different vision-language objectives have been designed to extract vision-language correlations. They can be broadly categorized into two groups: contrastive learning and masked prediction.

\subsubsection{Transformers}

Transformer and its variants \cite{kenton2019bert,vaswani2017attention} have been widely used to learn language features. In the past several years, Vision Transformers (ViT) \cite{dosovitskiy2021image} have also been employed in various vision tasks. Due to their powerful capabilities, Transformers with minor modifications can be used to learn representations from both vision and language modalities. 

ActBERT \cite{zhu2020actbert} is proposed to jointly learn video-text representations from unlabeled data with self-supervised learning. Firstly, global actions, local region objects, and text descriptions are integrated for fine-grained visual and text relation learning. Secondly, a tangled Transformer block encourages better interactions between three source data, discovering global-local correspondences.

The vision- and language-specific task is first pre-trained in a task-specific way, and fine-tuned for the target task, which lacks generic vision and language pre-training and suffers from overfitting when the target data is limited. Visual-Linguistic BERT (VL-BERT)\cite{su2020vl} uses either a word from a sentence or a region-of-interest from an image as the input of Transformers to learn multimodal representations, which is a good fit for vision-language downstream tasks.

There are three drawbacks in current bounding box-based region features. First, contextual information is neglected, which plays an important role in relation understanding and reasoning. Secondly, region representations are limited to the pre-defined categories. Thirdly, region features based on bounding boxes are low-quality, noisy, and rely on large-scale annotations. Without relying on slow bounding boxes, Seeing Out of tHe bOx (SOHO) \cite{huang2021seeing} directly learns image and language embeddings, and semantic alignments on image-text pairs. In particular, a visual dictionary denotes high-level abstractions of the similar content, better aligning visual features and language tokens. Masked visual modeling randomly masks image features before inputting them to the Transformer.

Many approaches have a heavy computation in computing visual embeddings. Vision-and-Language Transformer (ViLT) \cite{kim2021vilt} can extract lightweight and fast visual embeddings. Without relying on region or grid features, linear projection is applied on image patches to have a similar performance. Further, whole word masking and image augmentations can improve the performance.

Since self-attention has quadratic time and memory complexities with regard to the input sequence length, it is expensive to apply Transformers into long documents or high-resolution images. Long-Short Transformer (Transformer-LS) \cite{zhu2021long} can efficiently model long language and vision sequences with linear complexity. Firstly, a dynamic project based attention builds long-range connections and a local window attention models fine-grained correlations. Secondly, a dual normalization approach can address the scale mismatch between two attention modules.

Previous video captioning methods use off-the-shelf feature extractors to learn video features, however, there exists a gap in both data distribution and task formulation between off-the-shelf extractors and downstream video captioning. On the other hand, training feature extractors is computationally expensive. An end-to-end fully Transformer (SWINBERT) \cite{lin2022swinbert} is proposed. A video Transformer is used to learn spatial-temporal features that can be applied to variable video lengths. To reduce redundancy in video frames, a sparse attention mask is proposed to focus on frame patches that contain spatio-temporal movements.

Although fully Transformer-based models have better performance than previous region-based approaches, they suffer from performance degradation on downstream tasks. A Multimodal End-to-end TransformER framework (METER) \cite{dou2022empirical} investigates several design principles: vision encoders, text encoders, multimodal fusion module, architecture design, and pre-training objectives. It is observed that cross-attention and encoder-only are beneficial for multimodal fusion. Besides, ViT is more important than language Transformer and masked image modeling cannot improve downstream task results.

Existing image captioning methods depend on a object detector to provide region features, which have two shortcomings: 1) Large computational complexity brought by RPN, ROI pooling, and NMS; 2) Labor-intensive box annotations. To build detector-free models, a VIsion
Transformer based image CAPtioning model (ViTCAP) \cite{fang2022injecting} uses grid features. Besides, the concept token network generates the semantic information and then injects them into the end-to-end caption.

Previous vision-language pre-training models can be divided into one-stream and two-stream networks. The former has a high computational complexity due to calculation of all possible query-candidate pairs and a additional object detector. The latter neglects interactions between vision and language. A novel COllaborative Two-Stream vision-language pretraining model (COTS) \cite{lu2022cots} achieves three-levels of cross-modal interactions: Instance-level interaction via cross-modal momentum contrastive learning; token-level interaction with masked vision-language modeling on both tokenized images and texts; task-level interaction between text-to-image and image-to-text retrieval tasks. Besides, to reduce the negative effects of large-scale noisy data, an adaptive momentum filter module adaptively filters out noisy image-text pairs with the momentum mechanism in the instance-level alignment.

Previous video-text pre-training methods can be divided into dual-encoder and joint-encoder methods. The former uses two  encoders to contrast video-level and text-level features, neglecting local relations between these two modalities. The latter concatenates video and text representations and learns fine-grained correlations, leading to high computational cost. BridgeFormer \cite{ge2022bridging} can learn fine-grained interactions between videos and texts with high efficiency by a novel multiple choice questions. Specifically, questions are first constructed by erasing noun and verb phrases, then answered by resorting video features. In this way, the video encoder can learn regional features and temporal dynamics, and build relations between local video-text features.

Recent works cannot process all modalities at once or need multiple well designed tasks. DAVINCI is the first self-supervised vision-language model, being capable of different downstream tasks across modalities (vision/language/vision+language), types (understanding/generation), and settings (zero-shot, fine-tuning, and linear evaluation) \cite{diao2023write}. DAVINCI first divides either the word sequence or image token sequence, and then use a cross-modal Transformer to combine the textual and visual features into a joint space. Prefix language modeling and prefix image modeling conduct language modeling with the image supervision and image modeling with natural language supervision simultaneously.

In visual grounding, previous methods first extract features on pre-defined proposals or anchors, and then combine them with text embeddings to locate the target in the text query. However, the predefined objects are inflexible to fully capture the visual context and attribute information in the text. TransVG \cite{deng2021transvg} leverages Transformers to build multimodal correspondence. Besides, it reformulates the visual grounding as a direct coordinates regression problem, without relying on region proposals or anchor boxes. However, TransVG adopts the shared Transformer encoder layers, achieving sub-optimal results. In VLTVG \cite{yang2022improving}, a Transformer-based framework is proposed to build useful features and conduct iterative cross-modal reasoning for locate target objects accurately. In specific, a visual-linguistic verification module highlights visual regions related with the text descriptions and suppresses unrelated ones. A language-guided feature encoder aggregates informative visual contexts to boost the discriminative ability. Further, a multi-stage cross-modal decoder iteratively refines visual and text information for accurate target localization.

In visual grounding, a single-step grounding method selects correct image regions without continuous optimization. In multimodal dynamic graph Transformer (M-DGT) \cite{chen2022multi}, the visual grounding is reformulated as a progressively optimized visual semantic alignment process. Specifically, a dynamic graph is built where regions and their semantic relations are regarded as nodes and edges, respectively. It can discover a matching region-text as a graph structure transformation. Consequently, it progressively gets tighter matching regions for phrases for each target.

In the video question answering, prior cross-modal pretraining methods usually suffer from two drawbacks: 1) Video encoders are two simple; 2) The formulation of video question answering is inferior. Video Graph Transformer (VGT) \cite{xiao2022video} is proposed. Specifically, a dynamic graph Transformer learns information about visual objects, their relations, and dynamics for spatio-temporal
reasoning. Further, video and text Transformers are used to encode video and text, respectively, and then vision-text interactions are performed by cross-modal interaction modules.

Most existing works either test high-level image understanding of images or address region-level image understanding. Fusion-In-the-Backbone-based transformER (FIBER) \cite{dou2022coarsetofine} can address these tasks simultaneously. Specifically, a cross-attention is inserted image and text backbones to achieve multimodal fusion. Besides, a coarse-grained pretraining is conducted on image-text data, followed by a fined-grained pretraining which is based on the image-text-box dataset.

\textbf{Pros and cons}. Currently, Transformer models have already dominated in the vision and language field, which makes it possible to learn vision and language with a single Transformer model. Naturally, a question is raised: Whether the Transformer model is the optimal architecture in this field? It is noted that diffusion model \cite{saharia2022photorealistic} for image generation is gaining popular recently. It may be worth investigating whether the diffusion model can be employed in this field.

\subsubsection{Contrastive Learning}

Contrastive learning learns vision and language representations by pulling paired samples close and pushing away unpaired ones.  

There are several drawbacks in existing methods: 1) Image features and word embeddings reside in their own spaces, making it challenging to build their relations; 2) Object detectors are used to extract region-level features, which are annotation and computation expensive; 3) Masked language modeling may overfit to the inherently noisy datasets collected from the web. ALign BEfore Fuse (ALBEF) \cite{li2021align} is proposed. Specifically, an image-text contrastive loss aligns image features and text features before fusing them through cross-modal attention. Besides, momentum distillation is proposed to learn under noisy data.

Cross-modal alignment maximizes the mutual information between image and text pairs. However, it fails to consider that similar content within the same modality are close.  Triple contrastive learning (TCL) \cite{yang2022vision} leverages both cross-modal intra-modal self-supervision signals. Besides, instead of only using global information for multimodal contrastive learning, we maximize local mutual information between local regions and global information, leveraging localized and structural information from image and text.

Multimodal Information Injection Plug-in (MI2P) \cite{liang2022expanding} fine-tunes pre-trained unimodal models for multimodal classification. Besides, it learns cross-modal interactions at different layers between visual and textual features.

Although contrastive vision-language pre-training (CLIP) \cite{radford2021learning} has achieved decent results on image classification, it has sub-optimal results on object detection. RegionCLIP \cite{zhong2022regionclip} can address this issue by extending CLIP to learn part-level visual representations. Specifically, candidate regions are generated from an input image with object proposals or dense sliding windows. Meanwhile, region descriptions are synthesized from text corpus. Then, a pretrained CLIP model is used to align region text with image parts, generating pseudo region-text pairs. Lastly, except the image-text pairs, pseudo region-text pairs provide additional information to better pretrain vision-language model by contrastive learning.

Learning fine-grained information is beneficial to many tasks. A grounded language-image pretraining (GLIP) model \cite{li2022grounded} is proposed to extract object-level, language-aware, and semantic-rich visual features by unifying object detection and phrase grounding. The phrase grounding can identify the detailed correspondence between phrases in a sentence and objects in an image, leveraging region-word pairs. Further, cross-modality deep fusion combines information from two modalities in early layers, learning language-aware representations. GLIPv2 \cite{zhang2022glipv} combines localization pre-training and vision-language pre-training via three pretraining tasks: vision language reformulation of detection via phrase grounding, region-word contrastive learning, and masked language modeling.


\textbf{Pros and cons}. 
Contrastive learning-based vision and language models have demonstrated great capabilities in learning powerful features for various vision-language tasks. However, image features and text representations usually exhibit different levels of granularity. Language contains supervisions at multiple levels of granularity, e.g., on objects, scenes, context, and their relations. Similarly, visual concepts provide hierarchical semantics, e.g, pixels, parts, objects, and scenes. There lacks a mechanism to explicitly align visual and semantic concepts at the same level.

\subsubsection{Masked Prediction}

Masked prediction can be applied into vision and language tasks. It masks information in single modal or both modalities, then reconstructs the masked information.  

Prior works either focus on single-modal tasks or multi-modal tasks, which fail to adapt to each other. To address this issue, UNIMO \cite{li2021unimo} employs masked image modeling for image-only data, and masked language modeling and sequence-to-sequence generation for text-only data. Then, both unimodal and multimodal data is used to find similar samples from the unimodal data as positive pairs for cross-modal contrastive learning.

For contrastive methods, they usually have good performance for cross-modal tasks, while tending to suffer from inferior results on multimodal tasks. For Transformer models, they have either earlier fusion or shared self-attention across modalities, leading to inferior performances on unimodal vision-only or language-only tasks. Foundational language and vision alignment model (FLAVA) \cite{singh2022flava} is a holistic universal model to deal with vision, language, cross-, and multimodal vision and language tasks. Masked image modeling and masked language modeling are used on an image encoder and a text encoder, respectively. Further, masked multimodal modeling and image-text matching are used for paired image-text data.

Contrastive learning is dominant in the vision-language area. While effective, it introduces sample bias due to data augmentation, and is limited to image-text pairs, failing to leverage large-scale unpaired data. Multimodal Masked Autoencoder (M3AE) \cite{geng2022multimodal} consists of an encoder to project language tokens and image patches into a joint feature space, and a decoder to reconstruct the original pixels and text and predict masked tokens, without relying on paired image-text data.

Few attempts have been made to generate both text and image data with a single model, resulting in high computational consumption and deployment limitations in resource constraint environments. Masked Generative VL Transformer (MAGVLT) \cite{kim2023magvlt} is proposed which can explore a unified masked generative vision-language model. It combines image-to-text, text-to-image, and joint image-and-text mask prediction tasks. Besides, step-unrolled mask prediction and selective prediction on the mixture of two image-text pairs are designed to improve the generalization ability.

Due to the complexity of vision and language, large-scale training is important which is especially time-consuming. To this end, Fast Language-Image Pre-training (FLIP) \cite{li2023scaling} conducts contrastive learning on image and text pairs where image patches are randomly masked out with a large mask ratio and only the visible patches are encoded. Masked image reconstruction is not performed. 

\textbf{Pros and cons}. It is worth investigating how to design proper reconstruction targets in both modalities, complementing each other in both targets and improving mutual benefits. Besides, the masking strategy in the vision and language modalities should also be explored (e.g., mask ratios) to maximize the potentials of learned self-supervised features.

\begin{table*}
\centering
\caption{Representative MML methods with local mechanisms.}
\begin{tabular}{|c|c|c|c|c|c|c|c|}
\hline
\multicolumn{2}{|c|}{\multirow{1}{*}{Category}} & \multirow{1}{*}{Methods} & \multirow{1}{*}{Venue} & \multirow{1}{*}{Highlights} \\
\hline
\multirow{45}{*}{\rotatebox[origin=c]{90}{\makecell{Vision and Language}}}&\multirow{24}{*}{\rotatebox[origin=c]{90}{\makecell{Transformers}}} & \makecell{ActBERT \cite{zhu2020actbert}}&CVPR20 &\makecell{Global actions, local objects, and text descriptions, tangled Transformer block}\\
\cline{3-5}
&& \makecell{VL-BERT \cite{su2020vl}}&ICLR20 &\makecell{Multimodal features, Transformers}\\
\cline{3-5}
& &\makecell{SOHO \cite{huang2021seeing}}&CVPR21 &\makecell{Visual dictionary, masked visual modeling}\\
\cline{3-5}
& &\makecell{ViLT \cite{kim2021vilt} }&ICML21 &\makecell{Image patches, whole word masking, image augmentations}\\
\cline{3-5}
&&\makecell{Transformer-LS \cite{zhu2021long}}&NeurIPS21 &\makecell{Dynamic project attention, local window attention, dual normalization}\\
\cline{3-5}
&& \makecell{SWINBERT \cite{lin2022swinbert} }&CVPR22 &\makecell{A video Transformer, sparse attention mask}\\
\cline{3-5}
&& \makecell{METER \cite{dou2022empirical}}&CVPR22 &\makecell{Vision encoders, text encoders,\\ multimodal fusion module, architecture design, pre-training objectives}\\
\cline{3-5}
&& \makecell{ViTCAP \cite{fang2022injecting}}&CVPR22 &\makecell{Detector-free, grid features, concept token network}\\
\cline{3-5}
&&\makecell{COTS \cite{lu2022cots}}&CVPR22 &\makecell{Instance-, token-, and task-level interactions, adaptive momentum filter}\\
\cline{3-5}
&&BridgeFormer \cite{ge2022bridging}& \makecell{CVPR22}&\makecell{Fine-grained video-text interactions, multiple choice questions}\\
\cline{3-5}
&&DAVINCI \cite{diao2023write}& \makecell{ICLR23}&\makecell{Multimodal understanding/generation, cross-modal Transformer, \\prefix language modeling, prefix image modeling}\\
\cline{3-5} 
&& \makecell{TransVG \cite{deng2021transvg}}&ICCV21 &\makecell{Multimodal Transformer, direct coordinates regression}\\
\cline{3-5} 
&& \makecell{VLTVG \cite{yang2022improving}}&CVPR22 &\makecell{ Visual-linguistic verification, language-guided feature encoder, \\multi-stage cross-modal decoder}\\
\cline{3-5} 
&& \makecell{M-DGT \cite{chen2022multi}}&CVPR22 &\makecell{ Progressive visual semantic alignment optimization, \\ dynamic graph structure transformation}\\
\cline{3-5} 
&& \makecell{VGT \cite{xiao2022video}}& ECCV22 &\makecell{Dynamic graph Transformer, cross-modal interactions}\\
\cline{3-5} 
&& \makecell{FIBER \cite{dou2022coarsetofine}}& NeurIPS22 &\makecell{Cross-attention multimodal fusion, coarse-grained and fine-grained pretraining}\\
\cline{2-5}
&\multirow{2}{*}{\rotatebox[origin=c]{90}{\makecell{Contrastive Learning}}}& \makecell{ALBEF \cite{li2021align}}&NeurIPS21 &\makecell{Image-text contrastive loss, cross-modal attention, \\momentum distillation}\\
\cline{3-5}
&& \makecell{TCL \cite{yang2022vision}}&CVPR22 &\makecell{Cross-modal alignment,
intra-modal contrastive, \\local mutual information}\\
\cline{3-5}
&& \makecell{MI2P \cite{liang2022expanding}}&CVPR22 &\makecell{Multimodal information injection, cross-modal interactions}\\
\cline{3-5}
&& \makecell{RegionCLIP \cite{zhong2022regionclip}}&CVPR22 &\makecell{Pseudo region-text pairs}\\
\cline{3-5}
&& \makecell{GLIP \cite{li2022grounded}}&CVPR22 &\makecell{Phrase grounding, cross-modality deep fusion}\\
\cline{3-5}
&& \makecell{GLIPv2 \cite{zhang2022glipv}}&NeurIPS22 &\makecell{Phrase grounding, region-word \\contrastive learning, masked language modeling}\\ 
\cline{2-5}
&\multirow{6}{*}{\rotatebox[origin=c]{90}{\makecell{Masked Prediction}}}&UNIMO \cite{li2021unimo}& \makecell{ACL21}&\makecell{Masked image modeling, masked language modeling, \\sequence-to-sequence generation, cross-modal contrastive learning}\\
\cline{3-5}
&&M3AE \cite{geng2022multimodal}& \makecell{Arxiv22}&\makecell{Unpaired image-text data, masked token prediction}\\
\cline{3-5}
&& \makecell{FLAVA \cite{singh2022flava}}&CVPR22 &\makecell{Masked image modeling, masked language modeling, \\masked multimodal modeling, image-text matching}\\
\cline{3-5}
&& \makecell{MAGVLT \cite{kim2023magvlt}}&CVPR23 &\makecell{Image-to-text, text-to-image, and joint image-and-text mask prediction, \\step-unrolled mask prediction, \\selective prediction on the mixture of two image-text pairs}\\
\cline{3-5}
&& \makecell{FLIP \cite{li2023scaling}}&CVPR23 &\makecell{Contrastive learning on masked image content and text pairs, \\ no image reconstruction}\\
\hline
\multirow{8}{*}{\rotatebox[origin=c]{90}{\makecell{Vision, Audio, \\and Language}}}
&&MMV \cite{alayrac2020self}& NeurIPS20 &\makecell{Modality embedding graphs, multimodal contrastive loss, deflation method}\\
\cline{3-5}
&& \makecell{MCN \cite{chen2021multimodal}}& ICCV21 &\makecell{Contrastive loss, clustering loss,  reconstruction loss}\\
\cline{3-5}
&& \makecell{LAViT \cite{yun2021pano}}& ICCV21 &\makecell{Spherical spatial information, a multimodal Transformer encoder}\\
\cline{3-5}
&&VATT \cite{akbari2021vatt}& NeurIPS21 &\makecell{Modality-specific patch embeddings, \\modality-agnostic Transformers, DropToken}\\
\cline{3-5} 
&& \makecell{\cite{shvetsova2022everything}}&CVPR22 &\makecell{Multimodal fusion Transformer, combinatorial contrastive loss}\\
\cline{3-5} 
&& \makecell{\cite{li2022learning}}&CVPR22 &\makecell{Spatial grounding, temporal grounding, multimodal fusion}\\

\hline
\end{tabular}
\label{tab_mml}
\end{table*}

\subsection{Vision, Audio, and Language}

Most existing works focus on the joint modeling of vision and language, but ignore the rich information in audio. Although the semantic information in audio may have large overlaps with language, audio can provide extra information, like emotion. Besides, pre-trained models make the model being capable of audio-related downstream tasks. Several works shed light on this field. 

To learn multimodal representations in a self-supervised way without relying on human annotations in videos, MultiModal Versatile (MMV) network \cite{alayrac2020self} is proposed. Particularly, modality embedding graphs explore the best manner to preserve fine-grained features of visual and audio modalities and coarse-grained information of visual and text modalities. To train without manual annotations, a multimodal contrastive loss is used. Besides, a deflation method is proposed to process either statistic images or videos. 

To learn embeddings where samples with different modalities but similar content are close to each other, Multimodal Clustering Network (MCN) is proposed \cite{chen2021multimodal} where features from different modalities are projected into a joint space. Specifically, local semantic relationships are calculated by a contrastive loss which pulls features close across modalities. Global semantic relationships are calculated by a clustering loss which performs multimodal clustering across modalities to bring together semantically similar samples with different modalities. Further, multimodal feature reconstructions can extract features that are ignored by contrastive/clustering learning.

To tackle audio-visual question answering on panoramic videos, Language Audio-Visual Transformer (LAViT) \cite{yun2021pano} is proposed. Specifically, spherical spatial information is learned from a set of visual objects and audio events. Besides, a multimodal Transformer encoder is used to combine three different modalities. 

In order to train Transformers under large-scale unlabeled data, Video Audio-Text Transformer (VATT) \cite{akbari2021vatt} can extract video/audio/text representations which are beneficial to various downstream tasks. Modality-specific patch embeddings are extracted on different modalities, followed by modality-agnostic Transformers which have shared weights among different modalities. Besides, DropToken is proposed to randomly drop some video and audio tokens, allowing for high-resolution inputs. 

It is becoming increasingly popular to learn multimodal information from videos, because semantically meaningful features are extracted without human annotations. A multimodal fusion Transformer is proposed in \cite{shvetsova2022everything} to learn a joint multimodal embedding space on multiple modalities, including video, audio, and text where input tokens from different modalities are input of a fusion Transformer. Besides, a combinatorial contrastive loss adopts contrastive loss between all possible modality combinations, learning model-agnositic information. 

Audio-visual question answering is introduced in \cite{li2022learning}, which targets at answering questions based on visual information, sounds, and their relations. Firstly, a spatial grounding module is proposed to associate specific visual locations with the input sounds. Secondly, a temporal grounding module localizes important audio and visual segments using question features as queries. Finally, a multimodal fusion combines audio, visual, and question representations for predicting the answer.

\textbf{Pros and cons}. The aforementioned works are cutting-edge works which focus on combining different modalities effectively. However, joint modeling across vision, language, and audio is still an open problem left for further investigations. More fine-grained correlations mining in these three modalities is worth exploring to further boost the performance.

\subsection{Discussions}
Representative MML methods with local mechanisms are summarized in Table \ref{tab_mml}.

Currently, heavy models are used to achieve promising performance in the MML. However, model compression and acceleration remain to be explored where heavy models can be converted into light ones, meeting the requirements of fast inferences on resource-constrained devices. Multiple modalities can provide complementary information with each other. However, it should also be noted there exist redundant local information in different modalities. Efforts could be made to improve the efficiency of MML models by focusing on pruning redundant local representations or knowledge distillation on local features. 

Besides, current models tend to learn with a single language or audio (i.e., English). Consequently, models tend to have culture and region biases. Pre-training with multiple source of languages or audios can not only alleviate the bias issue, but also broaden application scenarios.

\section{Outlook}\label{sec_outlook}

In previous sections, we review how local mechanisms have been understood and integrated in various computer vision tasks and approaches. There are still space for further improvement and better ways to incorporate local mechanisms in various computer vision fields and approches. Inspired by the biological properties of the brain, below we have outlook about some potential future directions on how we can make better use of local mechanisms in computer vision research in several aspects: 1) Diversity; 2) Selectivity; 3) Knowledgeableness; 4) Sparsity; 5) Learning; 6) Contexts.

\begin{enumerate}

\item \textbf{Diversity}. It is found in \cite{yao2021transcriptomic} that the primary motor cortex of mice contain over 56 neuronal cell types to process real-world multimodal data. Similarly, the abundance of neuronal species in the cortex is revealed in \cite{alon2021expansion}. Inspired by these observations, it is highly expected to apply diversity in local mechanisms to benefit computer vision. First, it can improve the robustness of models. If some parts are invisible due to occlusions, pose or viewpoint variations, models can rely on the remaining visible parts to extract discriminative information, boosting the generalization ability. Second, it can allow models to learn from less data. Since diverse local information is extracted from the given input, this makes models suitable for various scenarios with few examples because of the richness of learned features. Third, it can alleviate the redundancy of models. In some cases, multiple local modules are repeated to cover comprehensive local features, however, there may exist redundancy between these local modules. Differently, diverse modules are designed to achieve this target with less model complexity. Currently, although many diverse local mechanisms are designed in different fields \cite{zheng2017learning,wang2022cqa,li2018diversity,chen2022principle} through encouraging the diversity between each two attentions, superior human prior knowledge can be explored to boost the diversity among multiple local mechanisms, simulating the powerful representational capability of brain neurons.

\item \textbf{Selectivity}. Attention selectivity plays an important role in conscious perception that allows us to filter out irrelevant sensory information in favor of the relevant. If attention is diverted, salient stimuli fails to perceive visual awareness. Monkeys are observed in \cite{roelfsema2011attention} to have the ability to control spatial attention of single neurons in vision. It is noticed in \cite{zhang2012neural} that the initial visual processing stage contains an attention mechanism to localize important information. It is investigated in \cite{saalmann2007neural} that a top-down feedback is required to select relevant sensory neurons that represent related features and locations in the environment. Inspired by them, it is highly expected to select discriminative local information with proper feedback guidance from the network. This is because there exist abundant information in digital images or videos, especially with real-world background clutters, however, some information may be meaningless or even noisy, deteriorating the performance. Therefore, it is desired to keep discriminative information and discard useless ones. This can have several benefits. First, it can improve the efficiency of models. Since useless information is discarded, this can reduce the complexity of models and boost the efficiency. Second, it can be flexibly adapted to different environments. In the real world, models could be deployed into different environments. With a proper guidance from the network, models can easily locate discriminative local features, adapting to different complex real-world scenarios. There are some work that explore the selectivity in different areas \cite{wang2017residual,anderson2018bottom,zhang2018top}. Selection involves not only actively processing target information, but also actively inhibiting or suppressing distracting information. One interesting question in selectivity is how to design top-down attention methods to accomplish selection under varying complex scenarios, keeping meaningless or even distracting information out of the focus and emphasizing important information.


\item \textbf{Knowledgeableness}. It is explored in \cite{josselyn2020memory} how new information is integrated into existing knowledge in brains. Understanding how humans store, process, and use information can inspire artificial intelligence researchers to develop more intelligent algorithms and architectures. Local features can well generalize to different test domains. It is therefore expected to memorize the necessary information and reactivate the memorized information by memory retrieval when local representations are used to transfer between different domains. To achieve this target, it is required to meet the following requirements: First, extracted local features are general which are applicable for different scenarios. Second, local features can be memorized for possible retrieval from different application scenarios. Third, local features from new domains can associate features in memory to benefit the representations. 

\item \textbf{Sparsity}. The receptive fields of simple cells in primary visual cortex are characterized by being spatially localized, oriented, and selective to structure at different scales. It is investigated in \cite{olshausen1996emergence} that a sparseness coding strategy is sufficient to meet these characteristics. The primary visual cortex adopts a sparse code to represent natural scenes which is computationally efficient for both early vision and higher visual processing \cite{vinje2000sparse}. The sparsity can have several benefits. First, it can reduce the computational cost, especially for complex models with heavy parameters and high memory cost. Second, it can reduce the over-fitting risk and boost the transferable ability across different domains. It is possible that models may over-represent objects with too much fine-grained details. This could deteriorate the model's capability of recognizing objects in unseen domains. Sparsification can remove redundant and useless information, making models robust to various challenging scenarios. The sparsity has been used in different areas \cite{mei2021image,zhu2021long,wang2021dsa}. It is worth noting that sparseness degree is a subject for future research. First, it can change under different layers. The low-level neurons represent detailed knowledge, like corners or textures. The information in high-level neurons stands for meaningful features of the real-world, such as complex object components. Intuitively, due to the difference of information richness, the sparseness degree in low-level neurons may be higher than high-level neurons, making it necessary to learn different sparseness degrees for different layers. Second, it can vary at different scenarios. For images with high redundant or even distracting information, the sparseness is expected to be high. Otherwise, the sparseness degree is low. It is desirable to design automatic processes, assigning different sparseness degrees under different scenarios. Third, it can differ if models are deployed at different application scenarios. Sometimes, models are deployed in an environment with general running speed and decent accuracy. Differently, models are expected to have the highest throughput. This can be achieved by controlling the sparseness degree of models.

\item \textbf{Learning}. It is pointed out in \cite{tenenbaum2011grow} that humans have the ability to learn generalized abstract knowledge from a small amount of data. For example, typical 2-year-olds can learn how to use new words from seeing only a few examples, such as horses. They can grasp the inherent meaning, not just the sound, allowing them to have the ability to use the words properly in different situations. Generalization from rare data is important in learning languages for humans. The human perception system is observed to have a unique visual perceptual mechanism, named cognitive penetrability \cite{maier2019no}, which employs the linguistic prior knowledge to tune visual processing to category stimulus features, benefiting the recognizing of novel objects with few samples. Motivated by this observation, it is desired to extract generalizable local features from limited data and transfer across different domains and tasks under proper guidance. Deep understanding why the human brain works the way it does, in terms of what information the mind knows about the real world and abstract knowledge how it generalize to different environments, can benefit the designing of extraction and transferability of discriminative local features in computer vision. 

\item \textbf{Contexts}. Humans can well perceive the real world by using contexts which are important to memorize the past, understand the present, and anticipate the future \cite{maren2013contextual}. Contexts can provide information not only about how semantic concepts (scenes, objects, object parts, or pixels) are related to each other, but also the relative positions or co-occurrence between different semantic concepts. For example, humans can identify the object in front of the computer monitor in the office environment is the keyboard, even if the object is partially occluded or heavily blurred. Context information can be well explored in our brain vision system. An object which cannot be identified in isolation can be recognized if relevant context appears. Relevant parts can be inferred which are useful for interpreting other parts. The context has been explored in various computer vision tasks \cite{zhang2020putting,liu2022nommer,ding2022davit}. However, it still faces some challenges to reason context information about objects and relations when being applied to computer vision. Current methods only learn from a limited number of images with a certain context, while humans observe objects in different contexts in the real world. Therefore, it is expected to develop contextual data augmentation methods, simulating various context scenarios. Humans can not only learn context and objects separately, but also establish the relations between objects and context. However, it still faces challenges to model the relations between object and context in computer vision. If the object is discriminative and it has a weak relation with its surrounding context, the context information is difficult to be captured. In comparison, if the object has a strong relation with its context, the context can be effectively learned. Besides, if without a proper guidance, the context may lead to the dataset bias issue where the relation between object and context is misleading. These variations make the computer vision system difficult to model context systematically and meaningfully. Many effective mechanisms can be designed to integrate highly semantic prior knowledge like humans into the process of learning contexts.

\end{enumerate}

\section{Conclusions}\label{sec_conclusion}

Local mechanisms play an important role in various computer vision tasks and approaches which have attracted significant research attention in the past decades. This survey covers the application of local mechanisms in different fields, including Vision Transformers, self-supervised learning, fine-grained visual recognition, person re-identification, few-shot learning, zero-shot learning, and multimodal learning. In conclusion, methods with local mechanisms have achieved great success and surpassed the performance of methods without local mechanisms. Besides, the advantages and disadvantages of local mechanisms in these fields are analyzed comprehensively. Some potential research directions have been discussed.

\ifCLASSOPTIONcaptionsoff
  \newpage
\fi

\bibliographystyle{IEEEtran}
\small
\bibliography{0bare_jrnl_new_sample4}

\newpage

 




\vfill

\end{document}